\documentclass[preprint,3p,fleqn]{elsarticle}
\usepackage{lmodern}
\usepackage{xcolor}
\usepackage{graphicx}
\usepackage{caption}
\usepackage{subcaption}
\usepackage{float}
\usepackage{booktabs}
\usepackage{multirow}
\usepackage{adjustbox}
\usepackage{enumitem}
\usepackage{algorithm}
\usepackage{algpseudocode}
\usepackage{indentfirst}
\usepackage{tabularx}
\usepackage[colorlinks=true, citecolor=blue, linkcolor=blue, urlcolor=blue]{hyperref}
\algrenewcommand{\algorithmiccomment}[1]{\hfill$\triangleright$ #1}

\biboptions{sort&compress}
\usepackage{amssymb}
\usepackage{amsmath}
\usepackage{textcomp}
\usepackage{hyperref}
\usepackage{cleveref}
\crefname{equation}{Eq.}{Eqs.}

\journal{arXiv}

\begin{document}

\begin{frontmatter}


\title{Interpretable identification of constitutive parameter field for three-dimensional heterogeneous and graded hyperelastic materials from full-field displacements\tnoteref{label1}}

\author[iisc_label]{Kanhaiya Lal Chaurasiya}            
\author[iisc_label]{Vaskar Halder\fnref{equal}}  
\author[iisc_label]{Prajyot Patil\fnref{equal}} 
\author[iisc_label]{Gaurav Navnath Shinde}              
\author[iisc_label]{Debashish Das}              
\author[iisc_label]{Akshay Joshi\corref{cor1}}  
\cortext[cor1]{Corresponding author}
\ead{akshayjoshi@iisc.ac.in}

\fntext[equal]{These authors contributed equally.}

\affiliation[iisc_label]{organization={Department of Mechanical Engineering, Indian Institute of Science}, % Department and Organization 
            city={Bengaluru},
            country={India}}
            
%% Abstract
\begin{abstract}
The inverse problem of identifying spatially varying material properties of heterogeneous solids using experimentally available displacement fields and boundary reaction forces alone remains challenging. Being able to characterize heterogeneous and graded solids would be advantageous in the design and manufacturing of functionally graded materials and composites. Additionally, an efficient technique for identifying heterogeneous material properties from in vivo measurements would be useful in biomedical contexts to characterize heterogeneous biological tissues. Although several techniques developed over the past two decades have addressed inverse identification for homogeneous materials, including linear elastic, hyperelastic, viscoelastic, and elastoplastic solids, comparatively fewer studies have considered heterogeneous and graded materials. Furthermore, several recent approaches to characterizing heterogeneous and graded materials employ parameter-heavy, deep learning-based frameworks trained to enforce force equilibrium at specified collocation points within the material. In this work, we propose an interpretable and time-efficient framework that predicts the three-dimensional relative spatial variation of a constitutive parameter field for multiple parameters of a heterogeneous hyperelastic material, using only displacement-field data. To achieve this, we subdivide the geometry into finite elements and formulate an objective function that depends on the constitutive parameters at each finite element, rather than on the parameters of a neural network that outputs the constitutive parameters. The objective function enforces force equilibrium at the integration point of each element, along with additional regularization terms to ensure realistic and physically consistent constitutive parameters. In this work, we validate our framework on 3D synthetic and 2D experimental benchmarks involving heterogeneous, graded, and anisotropic hyperelastic materials. By leveraging the linear nature of the inverse problem for hyperelastic materials and directly optimizing for the constitutive parameters at the finite-element level, we achieve acceptable prediction accuracy with significantly lower computational time compared to recently reported deep-learning-based frameworks. 
\end{abstract}

%% Keywords
\begin{keyword}
Heterogeneity \sep hyperelastic material \sep constrained optimization \sep elastography  \sep digital image correlation
\end{keyword}

\end{frontmatter}


\section{Introduction}
\label{sec_introduction}

Functionally graded materials (FGMs) are a class of heterogeneous materials in which composition, microstructure, and mechanical properties vary continuously across the specimen rather than changing abruptly between distinct material regions \citep{Birman2007,Li2020}. Such spatial grading of mechanical properties is often encountered in engineered composites, architected materials, and biological tissues, where it can arise either by design or through the underlying material architecture \citep{Li2020,Bayar2026}. However, their constitutive characterization is challenging because the material property varies with position, especially in case of anisotropic FGMs such as fiber-reinforced composites and biological tissues containing collagen fiber networks. Oriented micro- and mesoscale fibers impart directional stiffness to anisotropic hyperelastic materials at the macroscopic scale \citep{Holzapfel2019}. A method to accurately, reliably and rapidly characterize the spatially varying constitutive model of FGMs would enable their use in various engineering applications. For example, accurately designed fiber-reinforced composites with desired spatial and directional variations could be used to make lighter, stronger, and more fracture-resistant airplane wings \citep{Brooks2019}. Similarly, well-designed anisotropic elastomers can be used for direction-specific morphing in soft robotics \citep{Connolly2015,SydneyGladman2016}. In biological materials, methods for modeling graded hyperelastic materials can better aid modern elastography techniques. Such techniques are frequently used to characterize pathological changes in tissue mechanics, as malignant tissues are commonly associated with increased stiffness and exhibit different anisotropies compared to softer healthy tissues \citep{Skerl2016, Nia2020, Micalet2021, Koorman2022}.\par 

The increasing availability of full-field deformation measurements through techniques such as Digital Image Correlation (DIC) \citep{Genovese2021, Chen2022a, apunaman2026, Abbasi2026}, Digital Volume Correlation (DVC) \citep{Bay1999,Wang2024,Grega2025} and Magnetic Resonance Imaging (MRI) \citep{Avril2008a,Estrada2020} has motivated a wide range of inverse approaches to obtain constitutive models for homogeneous materials. These include the Finite Element Model Updating (FEMU) \citep{Kavanagh1971,Avril2008,Marwala2010,Pottier2011} approach, the Virtual Fields Method (VFM) \citep{Grdiac2006, Promma2009}, Equilibrium-Gap based Methods \citep{Ikehata1990, Claire2004, Claire2007,Geymonat2002,Geymonat2003,Florentin2010}, and the EUCLID framework \citep{Flaschel2021,Marino2023,Flaschel2022,Thakolkaran2022,Joshi2022}. Some of these frameworks have been adapted to identify the spatial variation of Young's Modulus (Poisson's ratio usually assumed to be 0.5) in linear elastic materials using surface displacement and boundary force data \citep{Toussaint2006, Avril2008a, Haustrate2025, Tan2017}. More recent works \citep{Chaurasiya2026, Shi2025} extend the Bayesian-EUCLID \citep{Joshi2022} framework and the NN-EUCLID framework to model heterogeneous hyperelastic solids with clear segment boundaries between the heterogeneities. These works use residual forces to highlight inter-segment boundaries and model each segment separately, and thus are unable to model graded materials (such as FGMs) which have no clear distinct boundaries between materials. 

All of the aforementioned frameworks-- FEMU, VFM and Equilibrium Gap-based methods such as EUCLID, rely on enforcing the weak-form of momentum balance on the underlying geometry to obtain an appropriate constitutive model for the material of interest. An alternative approach has been to represent the spatial distribution of material properties (usually Young's modulus) using neural networks (NNs) or neural operators, which are then trained to minimize the residual of the strong form of momentum balance at every collocation point \citep{Zhang2020,Mallampati2021,Chen2021,Tripura2023}. ElastNet and El-UNet \citep{Chen2023,Kamali2023, Kamali2024}, introduced deep-learning-based inverse frameworks that map spatial coordinates to a heterogeneous elastic-modulus and Poisson's ratio field. Another work similar to ElastNet \citep{Liu2025}, provided a framework for 3D elastography for linear elastic solids. PIE-PINN model \citep{Srikitrungruang2025, Srikitrungruang2026} employs a probabilistic formulation combining a B-spline-guided displacement approximation, neural correction, and adaptive residual weighting to recover spatially varying Young's modulus and Poisson's ratio from noisy and low-resolution displacement measurements. Similarly, AR-PINN \citep{Zhu2026} extends the technique to the large deformation regime, reconstructs an unknown heterogeneous field from sparse Green–Lagrange strain observations, and uses residual-adaptive collocation to resolve sharp interfaces. Alternatively, Kim et al. \citep{Kim2026} devised a supervised Fourier Neural Operator (FNO) \citep{Li2020a} architecture to map ultrasound-derived axial displacement fields directly to spatial elastic modulus distributions. Once trained, the FNO-based model provides rapid inference and exhibits robustness to measurement noise. These studies demonstrate the potential of deep learning techniques to reconstruct material-property fields directly from deformation measurements. However, it remained predominantly two-dimensional and restricted to isotropic linear-elasticity: ElastNet, El-UNet, PIE-PINN, and FNO-based models consider small-strain, linear-elastic, incompressible behavior under plane stress/strain conditions, whereas the AR-PINN considers a scalar Neo-Hookean form, with the bulk parameter held fixed in a two-dimensional isotropic setting. \par

    \begin{figure}[H]
        \centering
        \includegraphics[width=\linewidth]{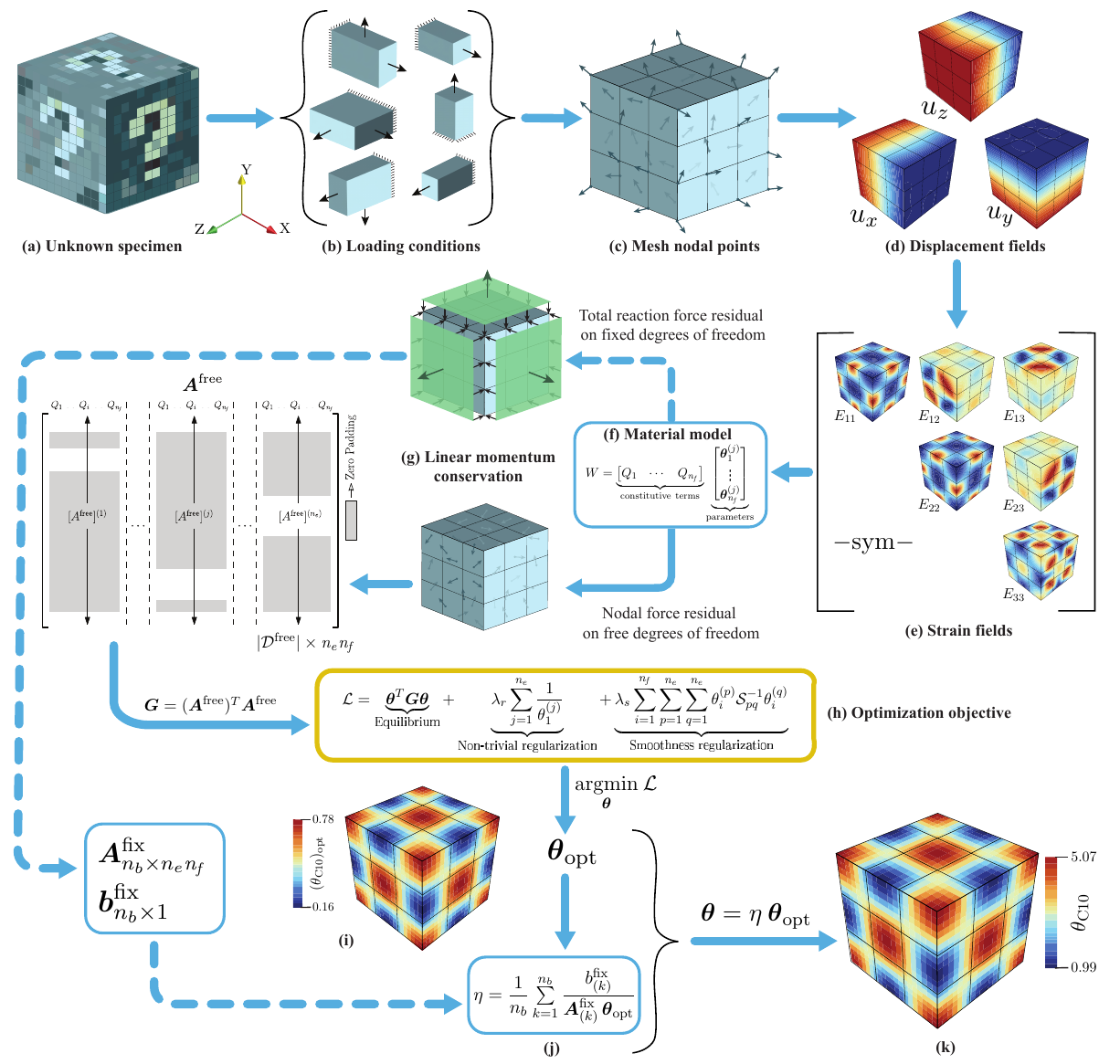}
        \caption{\textbf{Detailed schematic of the proposed inverse identification framework.} \textbf{a}, Unknown three-dimensional graded hyperelastic specimen. \textbf{b}, Combination of biaxial and uniaxial loading conditions applied to the specimen. \textbf{c}, Finite-element representation of the measured nodal displacement field. \textbf{d} and \textbf{e}, Full-field displacements and corresponding Green--Lagrange strain components, respectively. \textbf{f}, \textit{A priori} prescribed constitutive model with element-wise material parameters. \textbf{g}, Balance of linear momentum, yielding the equilibrium matrices associated with the free degrees of freedom, \(\boldsymbol{A}^{\mathrm{free}}\), and the fixed degrees of freedom, \(\boldsymbol{A}^{\mathrm{fix}}\). \textbf{h}, Constrained optimization problem used to identify the element-wise constitutive parameters. \textbf{i}, Relative elastography \(\boldsymbol{\theta}_{\mathrm{opt}}\) obtained from the displacement-only equilibrium equations, identifying the spatial material-property contrast up to a global multiplicative constant. \textbf{j}, Evaluation of the scaling factor \(\eta\) from the measured boundary reaction forces. \textbf{k}, Recovered absolute material parameter field obtained as \(\boldsymbol{\theta}=\eta\,\boldsymbol{\theta}_{\mathrm{opt}}\). Solid blue arrows indicate the displacement-only pathway leading to relative elastography, whereas the dashed blue pathway incorporates the boundary reaction-force information required to recover the absolute material parameter scale.}
        \label{fig_graphical_abstract}
    \end{figure}
    
A notable recent work by Taç et al. \citep{Ta2026} proposed a fully data-driven inverse framework based on hyper-network Neural Ordinary Differential Equations (NODEs) \citep{Tac2022}. In this framework, a hyper-network maps each material point to the parameters of a local NODE constitutive model for hyperelastic material. Consequently, it can represent a continuously varying material response field without explicitly requiring either spatial segmentation or a closed-form strain energy density function. The NODE architecture they proposed automatically satisfies polyconvexity, objectivity, and stress-free reference configuration \citep{Tac2022, Fuhg2022, Linden2023, Ta2023}. The framework was demonstrated for complex heterogeneous hyperelastic geometries, an anisotropic inclusion, noisy measurements, and experimental validation using DIC data. Thus, the NODE-based framework substantially broadens the scope for heterogeneous and graded material characterization using full-field measurements. An important aspect of this formulation is its lack of interpretability due to the absence of any presumed constitutive model field. Therefore, the NODE-framework provides an implicit constitutive law at every point in the geometry, rather than a conventional set of material parameters at the mesh element or integration-point level. Although the absence of a prescribed constitutive model allows more complex material behavior to be characterized, it complicates visualizing and assessing the spatial variation of the predicted constitutive model field. This is because the NODE-framework requires pointwise queries of uniaxial stretch behavior at each material point to assess the material behavior at the respective point in the geometry. \par
    
A limitation common to all the aforementioned deep-learning-based elastography and NODE-based constitutive modeling frameworks is that they are computationally intensive to train and cannot be extended to 3D contexts in an efficient manner. This is because they use a large number of parameters to make their deep learning models more expressive. Furthermore, to the best of our knowledge, no framework combines interpretability and 3D constitutive model mapping for hyperelastic heterogeneous/graded materials. To address this gap, we propose a physics-informed, data-driven framework for the inverse identification of constitutive parameter fields in heterogeneous/graded hyperelastic materials. Fig.~\ref{fig_graphical_abstract} illustrates the schematic of the proposed framework. Similar to Equilibrium-Gap-based approaches such as EUCLID \citep{Flaschel2021}, our method formulates the inverse problem as a constrained optimization problem, with the main objective defined by the residual of the balance of linear momentum at the internal nodes. Further, we add smoothing and regularization terms to the objective to promote spatial continuity and non-zero values in the identified parameter fields. A salient feature of our method is that it solves for the constitutive parameters at each element separately, rather than imposing the same constitutive model for all elements in the geometry \citep{Flaschel2021}. However, unlike the EUCLID framework, we do not select the constitutive model as a part of the inverse problem, but choose a constitutive model \textit{a priori}. The proposed framework is physics-informed in the sense that the governing equilibrium equation directly enters the optimization objective, rather than relying on traditional data-supervised loss functions. Another salient feature of our framework is that in the absence of boundary-force information, we identify the constitutive parameter fields (such as the shear and bulk moduli, etc.) up to a global multiplicative constant \citep{Roux2020}. In situations where boundary-force data are additionally available, we recover the absolute parameter fields.\par

We demonstrate the identification of spatially varying constitutive parameters for both isotropic and anisotropic hyperelastic solids using synthetically generated, noisy, full-field displacement data for 3D geometries. We also validate our proposed approach using experimental 2D-DIC measurements on biaxial tension tests of thin plate-like heterogeneous hyperelastic specimens. The predicted material property fields are interpretable and are utilized to perform forward finite element (FE) simulations of the experiments for assessing the accuracy of the prediction. The 3D numerical studies demonstrate robustness to spatially varying heteroscedastic Gaussian noise in the displacement measurements. The optimization phase of our framework requires approximately 17--30 minutes for the 3D isotropic cases and 73 minutes for the 3D anisotropic case, with shorter times (around 10 minutes) for the 2D experimental and 3D local elastography examples. These computational times compare favorably with neural-network and neural-operator based approaches for elastography and constitutive identification in heterogeneous materials. The proposed framework therefore provides a computationally efficient means for characterizing heterogeneous and graded hyperelastic materials, with potential applications ranging from heterogeneous biological tissues to fiber-reinforced composites, while enabling the identified material fields to be used directly in predictive FE simulations and material design. \par

\section{Methods}
\label{sec_methods}

\subsection{Equilibrium Gap formulation for inverse identification}
\label{subsecEqGap}
Consider a hyperelastic solid occupying the reference configuration \(\Omega\subset\mathbb{R}^{3}\), discretized using \(n_e\) eight-node brick elements (C3D8). For a given load step \(t\), the measured nodal displacements are denoted by \(\boldsymbol{u}^{a,t}\), where \(a=1,\ldots,n_n\). Within an element, the displacement field and the corresponding deformation gradient \(\boldsymbol{F}\) are interpolated as shown in \cref{eq_FE_kinematics}
\begin{equation}
    \boldsymbol{u}^{t}(\boldsymbol{\mathrm{X}}) =
    \sum_{a=1}^{n_n} \mathbb{N}^{a}(\boldsymbol{\mathrm{X}})\boldsymbol{u}^{a,t} 
    \ ;  \qquad 
    \boldsymbol{F}^{t}(\boldsymbol{\mathrm{X}}) =  \boldsymbol{I} +  
    \sum_{a=1}^{n_n}\boldsymbol{u}^{a,t}\otimes\nabla_{\boldsymbol{\mathrm{X}}}\mathbb{N}^{a}(\boldsymbol{\mathrm{X}})
    \label{eq_FE_kinematics}
\end{equation}

where \(\mathbb{N}^{a}\) denotes the trilinear interpolation function associated with node \(a\). For a prescribed constitutive model, the strain-energy density in element \(j\) is expressed using the constitutive feature vector \(\boldsymbol{Q}(\boldsymbol{F})\) as shown in \cref{eq_W_elementwise}.
\begin{equation}
    W^{(j)}(\boldsymbol{F};\boldsymbol{\theta}^{(j)})
    =
    \boldsymbol{Q}^{T}\boldsymbol{\theta}^{(j)},
    \qquad
    \boldsymbol{\theta}^{(j)}
    =
    \left[
    \theta_{1}^{(j)},\ldots,\theta_{i}^{(j)},\ldots,
    \theta_{\mathit{n_f}}^{(j)}
    \right]^{T}
    \label{eq_W_elementwise}
\end{equation}
where \(\mathit{n_f}\) denotes the number of terms associated with the selected strain-energy function/constitutive model. Here, the subscript \(i\) identifies the corresponding constitutive feature \(Q_i\), while the superscript in parentheses \((j)\) denotes the finite element to which the parameter belongs.\par

Using the right Cauchy-Green deformation tensor \(\boldsymbol{C}=\boldsymbol{F}^{T}\boldsymbol{F}\), the principal invariants of \(\boldsymbol{C}\) and the anisotropic invariant associated with a prescribed fiber direction \(\boldsymbol{a}\) are defined as shown in \cref{eq_invariants}.
\begin{equation}
    I_1(\boldsymbol{C}) = \mathrm{tr}(\boldsymbol{C}), \quad 
    I_2(\boldsymbol{C}) = \frac{1}{2}[\mathrm{tr}^{2}(\boldsymbol{C}) - \mathrm{tr}(\boldsymbol{C}^2)], \quad
    I_3(\boldsymbol{C}) = \mathrm{det}(\boldsymbol{C}), \quad I_{a}(\boldsymbol{C},\boldsymbol{a}) = \boldsymbol{a} \cdot\boldsymbol{Ca}
    \label{eq_invariants}    
\end{equation}
where the unit vector describing the fiber orientation in three dimensions is written as shown in \cref{eq_fiber_direction}.
\begin{equation}
    \boldsymbol{a}
    =
    \begin{pmatrix}
    \cos\alpha\,\sin\beta \\
    \sin\alpha\,\sin\beta \\
    \cos\beta
    \end{pmatrix}
    \label{eq_fiber_direction}
\end{equation}
with \(\alpha\) denoting the azimuthal angle in the \(X\)-\(Y\) plane and \(\beta\) the polar angle measured from the \(Z\)-axis. The volumetric change is described by \(J=\mathrm{det}(\boldsymbol{F})=I_3^{1/2}\). The corresponding isochoric invariants are obtained from the unimodular right Cauchy--Green tensor \(\tilde{\boldsymbol{C}}=J^{-2/3}\boldsymbol{C}\) as \(\tilde{I}_1=J^{-2/3}I_1\), \(\tilde{I}_2=J^{-4/3}I_2\), and \(\tilde{I}_a=J^{-2/3}I_a\). \par

For a given constitutive model, we construct a  feature vector with the corresponding strain-energy terms as shown in \cref{eq_constitutive_features}.
\begin{align}
    &\boldsymbol{Q}(\boldsymbol{F})
    =\left[
    Q_1, \ldots,
    Q_i, \ldots,
    Q_{\mathit{n_f}}
    \right]^T, \nonumber\\
    &\mathrm{where,}~Q_i
    \in
    \begin{cases}
        \left(\tilde{I}_{1}-3\right)^{p}\left(\tilde{I}_{2}-3\right)^{\,q-p},
        & q \in \{1,\dots,N_{\mathrm{iso}}\},\;
          p \in \{0,\dots,q\} \\[2pt]
        \left(J-1\right)^{2r},
        & r \in \{1,\dots,N_{\mathrm{vol}}\} \\[2pt]
        \left(\tilde{I}_{a}-1\right)^{s},
        & s \in \{2,\dots,N_{\mathrm{aniso}}\}
    \end{cases}
\label{eq_constitutive_features}
\end{align}

The three branches in Eq.~\eqref{eq_constitutive_features} correspond to the isochoric, volumetric, and anisotropic feature terms, respectively,  while \(\mathit{n_f}\) denotes the number of constitutive terms in the selected model. For example, \(\mathit{n_f}=2\) for the compressible Neo-Hookean model considered in this study, while the Mooney-Rivlin and modified HGO* models contain three (\(\mathit{n_f}=3\)) constitutive terms. Eqs.~\eqref{eq_W_elementwise} and \eqref{eq_constitutive_features} inherently ensure that the predicted constitutive models satisfy objectivity, stress-free configuration and material symmetry constraints \citep{Joshi2022}.

Subsequently, the first Piola-Kirchhoff stress tensor in element \(j\) is written as
\begin{equation}
    \boldsymbol{P}^{(j)}
    =
    \frac{\partial W^{(j)}}{\partial\boldsymbol{F}}
    =
    \frac{\partial\boldsymbol{Q}^{T}}
         {\partial\boldsymbol{F}}
    \boldsymbol{\theta}^{(j)}
    \label{eq_PK1_elementwise}
\end{equation}
Importantly, the stress response can be nonlinear with respect to the deformation; however, it remains linear with respect to the unknown constitutive parameters. The complete unknown constitutive parameter field \(\boldsymbol{\theta}\) is obtained by collecting the parameters of all \(n_e\) elements,
\begin{equation}
    \boldsymbol{\theta}
    =
    \left[
    \theta_{1}^{(1)},\ldots,\theta_{\mathit{n_f}}^{(1)},
    \ldots,
    \theta_{1}^{(j)},\ldots,\theta_{\mathit{n_f}}^{(j)},
    \ldots,
    \theta_{1}^{(n_e)},\ldots,\theta_{\mathit{n_f}}^{(n_e)}
    \right]^{T},
    \qquad
    \boldsymbol{\theta}\in\mathbb{R}^{n_e\mathit{n_f}\times 1}
    \label{eq_theta_elementwise}
\end{equation}

Assuming quasi-static loading and negligible body forces, the balance of linear momentum is imposed in weak form over the entire geometry $\Omega$ as
\begin{equation}
    \int_{\Omega}
    \boldsymbol{P}:\nabla\boldsymbol{v}~\mathrm{d}V
    -
    \int_{\partial\Omega}
    \hat{\boldsymbol{t}}\cdot\boldsymbol{v}~\mathrm{d}S
    =   0
    \label{eq_weak_equilibrium}
\end{equation}
where \(\boldsymbol{v}\) is an admissible test function and \(\hat{\boldsymbol{t}}\) denotes the traction acting on the boundary $\partial\Omega$. Let
\begin{equation*}
    \mathcal{D} = \{(a,i) : a=1,\dots,n_n \, ; \, i = 1,2,3 \}
\end{equation*}
denote the set of all nodal degrees of freedom. The set is divided into two mutually exclusive and exhaustive subsets -- the free degrees of freedom \(\mathcal{D}^{\mathrm{free}}\) and the displacement-constrained degrees of freedom \(\mathcal{D}^{\mathrm{fix}}\). Using the same finite element interpolation as the displacement field, the test function is written as
\begin{equation}
    v_i(\boldsymbol{\mathrm{X}}) = \sum_{a=1}^{n_n} \mathbb{N}^a(\boldsymbol{\mathrm{X}})v_i^a
    \label{eq_test_function}
\end{equation}
Substitution of Eq.~\eqref{eq_test_function} into Eq.~\eqref{eq_weak_equilibrium} gives
\begin{equation}
    \sum_{a=1}^{n_{n}} \left[\int_{\Omega} P_{i\mathbb{J}} \mathbb{N}_{,\mathbb{J}}^{a}(\boldsymbol{\mathrm{X}}) v_{i}^{a} \mathrm{~d} V - \int_{\partial \Omega} \hat{t_{i}} \mathbb{N}^{a}(\boldsymbol{\mathrm{X}}) v_{i}^{a} \mathrm{~d} S\right] = 0
    \label{eq_discrete_weak_form}
\end{equation}

    \begin{figure}[H]
        \centering
        \includegraphics[width=0.9\linewidth]{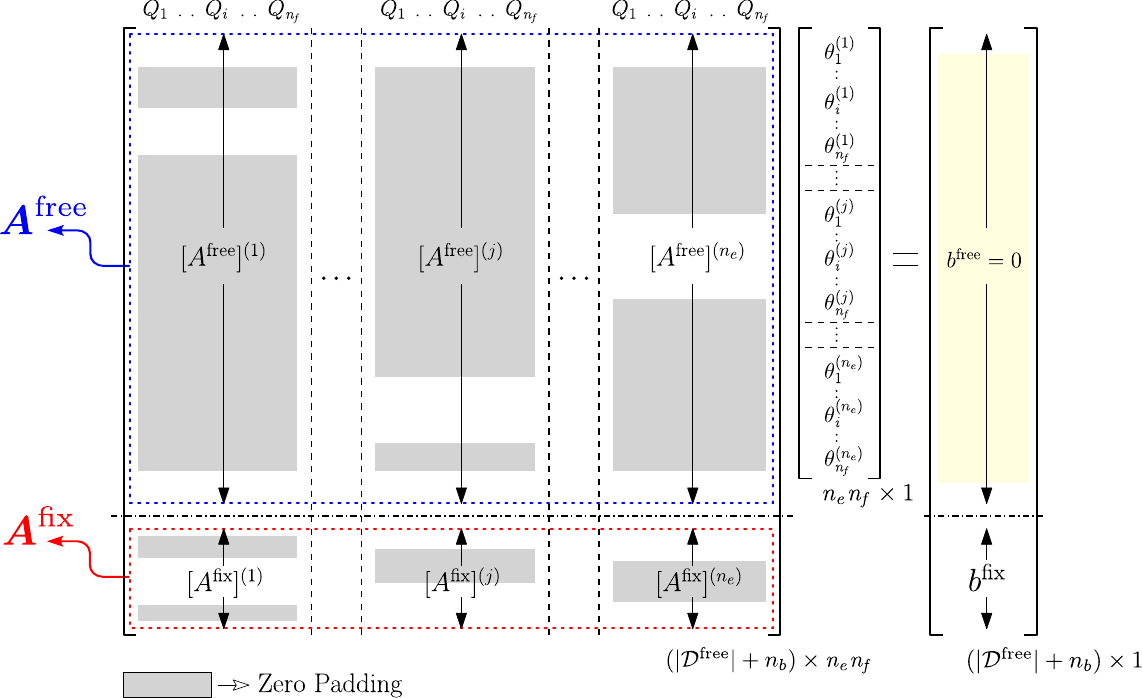}
        \caption{\textbf{Graphical illustration of the global discrete equilibrium system.} The left hand side shows free-degree-of-freedom matrix \( \boldsymbol{A}^{\mathrm{free}} \) and the displacement-controlled boundary matrix \( \boldsymbol{A}^{\mathrm{fix}} \). Each element-wise block \( \boldsymbol{A}^{(j)} \) contains \( \mathit{n_f} \) columns associated with the constitutive terms \( \boldsymbol{Q} \) and is zero-padded (gray regions) before column-wise assembly over all \( n_e \) finite elements. The parameter vector \( \boldsymbol{\theta}\in\mathbb{R}^{n_e \mathit{n_f}\times 1} \) stacks the constitutive coefficients of all elements, while \( \boldsymbol{b}^{\mathrm{free}}=\boldsymbol{0} \) enforces equilibrium at the free degrees of freedom and \( \boldsymbol{b}^{\mathrm{fix}} \) contains the boundary-aggregated reaction forces data at $n_b$ boundaries.}
        \label{fig_A_theta_b_FGM}
    \end{figure}
    
\noindent\textbf{Definition 1 (Equilibrium at free degrees of freedom)}

Continuing from \cref{eq_discrete_weak_form}, we see that:
\begin{equation}
    \sum_{a=1}^{n_{n}} v_i^{a}\left[\int_{\Omega} P_{i\mathbb{J}} \mathbb{N}_{,\mathbb{J}}^{a}(\boldsymbol{\mathrm{X}})  \mathrm{~d} V - \int_{\partial \Omega} \hat{t_{i}} \mathbb{N}^{a}(\boldsymbol{\mathrm{X}}) \mathrm{~d} S\right] = 0
    \label{eq_discrete_weak_form2}
\end{equation}

For, \cref{eq_discrete_weak_form2} to hold for arbitrary $v_i^{a}$, it is necessary and sufficient that 
\begin{equation}
    \int_{\Omega} P_{i\mathbb{J}} \mathbb{N}_{,\mathbb{J}}^{a}(\boldsymbol{\mathrm{X}})  \mathrm{~d} V - \int_{\partial \Omega} \hat{t_{i}} \mathbb{N}^{a}(\boldsymbol{\mathrm{X}}) \mathrm{~d} S = 0 \quad \forall~(a,i)\in \mathcal{D}
    \label{eq_discrete_weak_form3}
\end{equation}

When we consider only free degrees of freedom, i.e., $(a,i) \in \mathcal{D}^{\text{free}}$, the surface integral in \cref{eq_discrete_weak_form3} vanishes, i.e., $\int_{\partial \Omega} \hat{t_{i}} \mathbb{N}^{a}(\boldsymbol{\mathrm{X}}) \mathrm{~d} S =0 )$  as the shape function $\mathbb{N}^{a}$ vanishes on $\partial\Omega$. Thus, we will get \cref{eq_discrete_weak_form4} as
\begin{equation}
    \int_{\Omega} P_{i\mathbb{J}} \mathbb{N}_{,\mathbb{J}}^{a}(\boldsymbol{\mathrm{X}})  \mathrm{~d} V  = 0 \quad \forall~(a,i)\in \mathcal{D}^{\text{free}}
    \label{eq_discrete_weak_form4}
\end{equation}

Next, writing the volume integral in \cref{eq_discrete_weak_form4} as a summation of the integral over all the $n_e$ elements, and using \cref{eq_PK1_elementwise}, we get \cref{eq_free_elementwise}.
\begin{equation}
    \sum_{j=1}^{n_e}
    \left[
    \int_{\Omega^{(j)}}
    P_{i\mathbb{J}}^{(j)}
    \mathbb{N}^{a}_{,\mathbb{J}}~
    \mathrm{d}V
    \right]
    = \sum_{j=1}^{n_e}
    \underbrace{\left[
    \int_{\Omega^{(j)}}
    \frac{\partial\boldsymbol{Q}^T}{\partial F_{i\mathbb{J}}}
    \mathbb{N}^{a}_{,\mathbb{J}}~
    \mathrm{d}V
    \right]}_{\left[\boldsymbol{A}^{\text{free}}\right]^{(j)}}
    \boldsymbol{\theta}^{(j)} = 0   ~ \quad
    \forall~(a,i)\in\mathcal{D}^{\mathrm{free}}
    \label{eq_free_elementwise}
\end{equation}
where $\Omega^{(j)}$ denotes the volume associated with element $j$. The integral associated with element \(j\) provides the coefficients multiplying the constitutive parameters of element $j$, i.e,
\(\boldsymbol{\theta}^{(j)} = [\theta_{1}^{(j)},\ldots,\theta_{\mathit{n_f}}^{(j)}]^{T}\). Assembly over all elements and all free degrees of freedom therefore gives
\begin{equation}
    [\boldsymbol{A}^{\mathrm{free}}]^{(1)}
    \boldsymbol{\theta}^{(1)}
    +    \cdots    +
    [\boldsymbol{A}^{\mathrm{free}}]^{(j)}
    \boldsymbol{\theta}^{(j)}
    +    \cdots    +
    [\boldsymbol{A}^{\mathrm{free}}]^{(n_e)}
    \boldsymbol{\theta}^{(n_e)}
    =    \boldsymbol{0}
    \label{eq_Afree_elementblocks}
\end{equation}
\cref{eq_Afree_elementblocks} can be written compactly as
\begin{equation}
    \boldsymbol{A}^{\mathrm{free}}
    \boldsymbol{\theta}
    =
    \boldsymbol{b}^{\mathrm{free}}
    =
    \boldsymbol{0}; \quad \boldsymbol{b}^{\text{free}}\in\mathbb{R}^{\left|\mathcal{D}^{\text{free}}\right|\times1}
    \label{eq_Afree}
\end{equation}
where,
\begin{align*}
    &\boldsymbol{A}^{\mathrm{free}}
    =
    \left[
    [\boldsymbol{A}^{\mathrm{free}}]^{(1)}
    ;\ldots;
    [\boldsymbol{A}^{\mathrm{free}}]^{(j)}
    ;\ldots;
    [\boldsymbol{A}^{\mathrm{free}}]^{(n_e)}
    \right]; \quad \boldsymbol{A}^{\mathrm{free}} \in \mathbb{R}^{\left|\mathcal{D}^{\text{free}}\right|\times n_e \mathit{n_f}}, \\
    &\boldsymbol{\theta} = [\boldsymbol{\theta}^{(1)};\ldots;\boldsymbol{\theta}^{(j)};\ldots;\boldsymbol{\theta}^{(n_e)}]
    \label{eq_Afree1}
\end{align*}

Each elemental block \([\boldsymbol{A}^{\mathrm{free}}]^{(j)}\) contains \(\mathit{n_f}\) columns corresponding to the \(\mathit{n_f}\) constitutive parameters of element \(j\) in Eq.~\eqref{eq_constitutive_features}. An element contributes only to the equilibrium equations of its constituent nodes. This would imply that all rows in \([\boldsymbol{A}^{\mathrm{free}}]^{(j)}\) corresponding to degrees of freedom not associated with element $j$ would be zero. This gives $\boldsymbol{A}^{\text{free}}$ the sparse, zero-padded block structure illustrated in Fig.~\ref{fig_A_theta_b_FGM}.\par

For multiple loading states, the equilibrium information from each load step \(t\) would require vertically stacking the corresponding matrices \(\boldsymbol{A}_{t}^{\mathrm{free}}\), which can become memory intensive. Instead, for each load step \(t\), we compute 
\begin{equation}
    \boldsymbol{G}_{t}    =
    \left(\boldsymbol{A}_{t}^{\mathrm{free}}\right)^{T}
    \boldsymbol{A}_{t}^{\mathrm{free}}
    \label{eq_G_step}
\end{equation}
and accumulate the contribution from all \(n_t\) loading states from \cref{eq_G_step} as
\begin{equation}
    \boldsymbol{G}    =
    \sum_{t=1}^{n_t}\boldsymbol{G}_{t}
    =    \sum_{t=1}^{n_t}
    \left(\boldsymbol{A}_{t}^{\mathrm{free}}\right)^{T}
    \boldsymbol{A}_{t}^{\mathrm{free}}, \qquad \boldsymbol{G}\in\mathbb{R}^{n_e\mathit{n_f} \times n_e\mathit{n_f}}
    \label{eq_G_sum}
\end{equation}

This is equivalent to first stacking the individual \(\boldsymbol{A}_{t}^{\mathrm{free}}\) matrices and subsequently evaluating \((\boldsymbol{A}^{\mathrm{free}})^T\boldsymbol{A}^{\mathrm{free}}\), while avoiding the need to store the considerably larger stacked matrix. The resulting \(\boldsymbol{G}\), shown in \cref{eq_G_sum}, encapsulates the combined equilibrium and displacement field information from all load steps in a compact form and is used in the subsequent optimization formulation. \\ \par

\noindent\textbf{Definition 2 (Optimization objective)}

To obtain a constitutive parameter solution ($\boldsymbol{\theta}$) to \cref{eq_Afree}, we seek to minimize $\lVert \boldsymbol{A}^{\mathrm{free}}
    \boldsymbol{\theta} \rVert_2^2 \left(\equiv  \boldsymbol{\theta}^{T} \boldsymbol{G} \boldsymbol{\theta} \right)$, with a few additional constraints on $\boldsymbol{\theta}$. The first constraint is that $\boldsymbol{\theta}$ is non-trivial, i.e, not $\boldsymbol{0}$. More specifically, we ensure that components of $\boldsymbol{\theta}$ proportional to the shear modulus at each element, i.e, $\theta_1^{(j)}$ ($j^{\text{th}}$ element's $C_{\mathrm{10}}$ parameter) is required to be strictly non-zero. The second constraint we impose is the requirement for all components of $\boldsymbol{\theta}$ to be non-negative. The non-negativity of each entry of $\boldsymbol{\theta}$ would ensure that the strain energy density at each element would be a polyconvex function of the local deformation gradient, provided the energy library (\cref{eq_constitutive_features}) consists of polyconvex features\footnote{The $\left(\tilde{I}_2-3\right)$ term that appears in Ishihara models is not polyconvex \citep{Hartmann2003} and should ideally be replaced with $\left(\tilde{I}_2^{3/2}-3^{3/2}\right)$. However, we retain the traditional Ishihara term to maintain compatibility with traditional models.}. To achieve this, we optimize for $\alpha_i^{(j)}$ instead of $\theta_i^{(j)}$, where $\theta_i^{(j)} = \left(\alpha_i^{(j)}\right)^2$. For ease of notation we henceforth write $\boldsymbol{\theta} = \boldsymbol{\alpha}^2$.
    The third constraint we impose on $\boldsymbol{\theta}$ is the smoothness of each of its component in space. That is, we specify a distance-based correlation to each component of $\boldsymbol{\theta}$ depending on the distance between the respective elements. The objective function that incorporates all these constraints is shown in \cref{eq_Lossfunction}.
    \begin{equation}
        \mathcal{L} =
        \underbrace{\boldsymbol{\theta}^{T} \boldsymbol{G} \boldsymbol{\theta}}_{\text{Equilibrium}}
        +
        \underbrace{\lambda_{r} \sum\limits_{j=1}^{\mathit{n_e}} \dfrac{1}{\theta_1^{(j)}}}_{\text{Non-trivial regularization}}
        +
        \underbrace{\lambda_{s}\sum\limits_{i=1}^{\mathit{n_f}}\sum\limits_{p=1}^{\mathit{n_e}}\sum\limits_{q=1}^{\mathit{n_e}}\theta^{(p)}_{i} 
        \mathcal{S}^{-1}_{pq} \theta^{(q)}_{i}}_{\text{Smoothness regularization}}
        \label{eq_Lossfunction}
    \end{equation}

Writing the objective in terms of $\boldsymbol{\alpha}$ gives us \cref{eq_Lossfunction_alpha}.
\begin{equation}
    \mathcal{L}(\boldsymbol{\alpha}) =
        \left(\boldsymbol{\alpha}^2\right)^{T} \boldsymbol{G} \boldsymbol{\alpha}^2
        +
        \lambda_{r} \sum\limits_{j=1}^{\mathit{n_e}} \dfrac{1}{\left(\alpha_1^{(j)}\right)^2}
        +
        \lambda_{s}\sum\limits_{i=1}^{\mathit{n_f}}\sum\limits_{p=1}^{\mathit{n_e}}\sum\limits_{q=1}^{\mathit{n_e}} \left(\alpha^{(p)}_{i}\right)^2 
        \mathcal{S}^{-1}_{pq} \left(\alpha^{(q)}_{i}\right)^2
    \label{eq_Lossfunction_alpha}
\end{equation}
In the above equations, $\lambda_r$ and $\lambda_s$ are the weights assigned to the ``non-trivial'' and ``smoothness'' regularization terms, respectively. In this work, we have chosen $\lambda_r = 10^{-5}$ and $\lambda_s = 10^{-4}$ for all benchmarks. The spatial covariance matrix $\boldsymbol{\mathcal{S}}$ induces a Gaussian prior correlation between field values at different elements. It is defined as $\mathcal{S}_{ij} = \exp\left(-\dfrac{\lVert \boldsymbol{x}_i-\boldsymbol{x}_j\rVert_2^2}{2 \ell^2}\right)$, where $\boldsymbol{x}_i$ denotes the position vector associated with the centroid of the $i^{\text{th}}$ element, and $\ell$ is a length-scale parameter associated with the problem. In supplementary section S3.6, Fig.~S8 and S9 report the effect of the regularization length scale \(\ell\) on the predicted parameter field. We chose $\ell = 1.25~\ell_{\text{elem}}$ for all synthetic case-studies discussed in \cref{case_3D_checker_NH,case_3D_tanhsph_MR,case_3D_tanhsph_NH_HGO,case_3D_tanhsph_MR_local}, and $\ell = 3.0~\ell_{\text{elem}}$ for the experimental benchmarks discussed in \cref{case_exp}, where $\ell_{\text{elem}}$ denotes the cube-root of the average element volume for 3D synthetic cases, and square-root of average element surface area for 2D experimental cases (\cref{case_exp}). In supplementary section S1, we provide a more elaborate derivation of the objective \cref{eq_Lossfunction} as the posterior distribution for $\boldsymbol{\theta}$, given the experimental data (encapsulated by $\boldsymbol{G}$). Further, in supplementary section S3.3, we perform an ablation study of each term in \cref{eq_Lossfunction}.\par

We thus obtain the predicted constitutive parameter field $\boldsymbol{\theta}_{\text{opt}} = \boldsymbol{\alpha}^2_{\text{opt}}$, where \(\boldsymbol{\alpha}_{\text{opt}}= \text{argmin}_{\boldsymbol{\alpha}} \ \mathcal{L}(\boldsymbol{\alpha})\). The constitutive parameter field $\boldsymbol{\theta} = \boldsymbol{\theta}_{\text{opt}}$ would be correct up to a multiplicative constant. This is because we are yet to consider the boundary force information \cite{Roux2020} to determine the multiplicative factor $\eta$ (discussed next). In practical situations where boundary force information is not available -- such as embedded subdomains in X-ray or MRI scans, the $\boldsymbol{\theta}_{\text{opt}}$ provides a relative estimate of the parameter fields.
\\

\noindent\textbf{Definition 3 (Force balance at displacement-controlled boundaries)}

Consider a displacement-controlled boundary \(\partial\Omega_k(\subset \partial\Omega)\), and $(a,i)\in \mathcal{D}^{\mathrm{fix}}_k(\subset\mathcal{D}^{\mathrm{fix}})$, where \(\mathcal{D}^{\mathrm{fix}}_k\) denotes the degrees of freedom associated with the \(k^{\mathrm{th}}\) loaded boundary. Further, \(n_b\) is the total number of boundaries for which force information is available. Eq.~\eqref{eq_discrete_weak_form3} then gives the force balance at the $k^{\text{th}}$ boundary, $a^{\text{th}}$ node, and $i^{\text{th}}$ direction as
\begin{equation}
    \sum_{j=1}^{n_e}
    \left[
    \int_{\Omega_k^{(j)}}
    \frac{\partial\boldsymbol{Q}^{T}}
         {\partial F_{i\mathbb{J}}}
    \mathbb{N}^{a}_{,\mathbb{J}}~
    \mathrm{d}V
    \right]
    \boldsymbol{\theta}^{(j)}
    =  R_i^{k,a} \quad \mathrm{where,}~
                          i \in \{1, 2, 3\}; \quad
                          k \in \{1,\dots,n_b\};\quad a \in \mathcal{N}_k
    \label{eq_fix_elementwise}
\end{equation}
where, $\mathcal{N}_k$ denotes the set of nodes on the surface $\partial\Omega_k$ and
\(R_i^{k,a} ( = \int_{\partial \Omega_k} \hat{t_{i}} \mathbb{N}^{a}(\boldsymbol{X}) \mathrm{~d} S ) \) denotes the resultant reaction force acting on node $a$ at boundary \(k\) along direction \(i\). In most experiments, we only measure the sum of normal reaction forces on the boundaries, rather than reaction forces at individual nodes. If we assume $\boldsymbol{n}^{(k)}$ to be the normal vector for boundary $k$, we need to consider total the node-wise components along $\boldsymbol{n}^{(k)}$. Thus, we need to multiply $n^{(k)}_i$ on both sides of \cref{eq_fix_elementwise} and sum along all the nodes in set $\mathcal{N}_k$ as follows:
\begin{equation}
    \sum_{j=1}^{n_e}\underbrace{\sum_{a\in\mathcal{N}_k}\sum_{i=1}^{3}
    \left[
    \int_{\Omega_k^{(j)}}
    \frac{\partial\boldsymbol{Q}^{T}}
         {\partial F_{i\mathbb{J}}}n_i^{(k)}
    \mathbb{N}^{a}_{,\mathbb{J}}~
    \mathrm{d}V
    \right]}_{A^{\text{fix}(j)}_{k}}
    \boldsymbol{\theta}^{(j)}
    =  \sum_{a\in\mathcal{N}_k}\sum_{i=1}^{3}R_i^{k,a}n_i^{(k)} =  R^{(k)}
    \label{eq_fix_elementwise_nodessummed}
\end{equation}
In \cref{eq_fix_elementwise_nodessummed} $R_i^k$ denotes the $i^{(th)}$ component of reaction force measured at boundary $k$.
\cref{eq_fix_elementwise_nodessummed} can be further expressed as 
\begin{equation}
    [\boldsymbol{A}^{\mathrm{fix}}]^{(1)}
    \boldsymbol{\theta}^{(1)}
    +    \cdots    +
    [\boldsymbol{A}^{\mathrm{fix}}]^{(j)}
    \boldsymbol{\theta}^{(j)}
    +    \cdots    +
    [\boldsymbol{A}^{\mathrm{fix}}]^{(n_e)}
    \boldsymbol{\theta}^{(n_e)}
    =    \boldsymbol{b}^{\mathrm{fix}}
    \label{eq_Afix_elementblocks}
\end{equation}
\cref{eq_Afix_elementblocks} can be written compactly as
\begin{equation}
    \boldsymbol{A}^{\mathrm{fix}}
    \boldsymbol{\theta}
    = \boldsymbol{b}^{\mathrm{fix}}
    \label{eq_Afix}
\end{equation}
where,
\begin{align*}
    &\boldsymbol{A}^{\mathrm{fix}}   =
    \left[[\boldsymbol{A}^{\mathrm{fix}}]^{(1)}
    ;\ldots;
    [\boldsymbol{A}^{\mathrm{fix}}]^{(j)}
    ;\ldots;
    [\boldsymbol{A}^{\mathrm{fix}}]^{(n_e)}
    \right]; \quad \boldsymbol{A}^{\mathrm{fix}} \in \mathbb{R}^{n_b \times n_e \mathit{n_f}},  \\
    &\boldsymbol{b}^{\mathrm{fix}} = \left[R^{(1)},\dots,R^{(k)},\dots,R^{(n_b)}\right]^T ; \quad \boldsymbol{b}^{\mathrm{fix}}\in \mathbb{R}^{n_b\times1}
\end{align*}

The resulting structure of \(\boldsymbol{A}^{\mathrm{free}}\), \(\boldsymbol{A}^{\mathrm{fix}}\), and the element-wise parameter vector \(\boldsymbol{\theta}\) is summarized schematically in Fig.~\ref{fig_A_theta_b_FGM}. The formulation follows the weak-equilibrium construction introduced in Hetero-EUCLID \citep{Chaurasiya2026}, but replaces the segment-wise constitutive description by element-wise parameterization, which forms the basis for recovering continuously varying constitutive fields in the present work. Further, unlike Hetero-EUCLID and previous EUCLID works \citep{Chaurasiya2026,Flaschel2021}, we do not impose a sparsity promoting term in the objective function (\cref{eq_Lossfunction}).

In the absence of force information, Eq.~\eqref{eq_Afree} determines the constitutive parameter field only up to a global multiplicative scaling factor using the constrained optimization described in the previous section. We further utilize \cref{eq_Afix} in recovering this scale and to determine the absolute spatial parameter fields as 
\begin{align}
    \boldsymbol{\theta}
    &= \eta \, \boldsymbol{\theta}_{\mathrm{opt}}
    \label{eq_theta_elastofactor} \\
    \eta
    &= \dfrac{1}{n_b}\sum_{k=1}^{n_b}
    \dfrac{b_{(k)}^{\mathrm{fix}}}
    {\boldsymbol{A}_{k}^{\mathrm{fix}}
    \boldsymbol{\theta}_{\mathrm{opt}}}
    \label{eq_elastofactor}
\end{align}
where \(\eta\), defined in Eq.~\eqref{eq_elastofactor}, is computed as the average scaling factor obtained from the force balance over all traction boundaries of the domain. Note that we do not need $\eta$ when performing local elastography, as done in \cref{case_3D_tanhsph_MR_local}.

\subsection{Incorporating 2D-DIC data}
\label{sec_2D_formulation}
In the experimental validation performed in \cref{case_exp}, we use 2D-DIC-based displacement data that provides us only two in-plane components of the surface displacement. In order to construct the 3D deformation gradient field (see \cref{subsecEqGap}) from this 2D information, following Abbasi et al. \citep{Abbasi2026} we assume the following form for the deformation gradient,
\begin{equation}
    \boldsymbol{F} = 
    \begin{bmatrix}
    F_{11} & F_{12} & 0 \\
    F_{21} & F_{22} & 0 \\
    0 & 0 & F_{33}
    \end{bmatrix} , \qquad \mathrm{where}~F_{33} = \frac{1}{F_{11}F_{22} - F_{12}F_{21}}
    \label{eq_defGrad}
\end{equation}

The incompressibility assumption ($J = \text{det}(\boldsymbol{F})=1$) is necessary to obtain the unknown $F_{33}$ component. For most hyperelastic materials incompressibility is a reasonable assumption to make as their bulk modulus is usually an order of magnitude higher compared to their shear modulus. The constitutive law applicable to strictly incompressible materials differ from the law discussed in \cref{eq_W_elementwise}. The element-wise strain energy for strictly incompressible materials is shown in \cref{eq_W_elementwise_incompressible}.
\begin{equation}
    W^{(j)}(\boldsymbol{F};\boldsymbol{\theta}^{(j)})
    =   \boldsymbol{Q}^{T}(I_1,I_2,I_a)\boldsymbol{\theta}^{(j)}
    -p(J-1)
    \label{eq_W_elementwise_incompressible}
\end{equation}

where \(p\) is the Lagrange multiplier introduced to enforce the incompressibility constraint \(J=1\). Under strict incompressibility, the volumetric feature terms are omitted from \(\boldsymbol{Q}\) (Eq.~\eqref{eq_constitutive_features}), since the volume change is constrained directly through \(p\). Using Eq.~\eqref{eq_W_elementwise_incompressible} along with the relation \(\dfrac{\partial J}{\partial F_{i\mathbb{J}}}=J F_{i\mathbb{J}}^{-T}\), the first Piola--Kirchhoff stress tensor in element \(j\) becomes
\begin{equation}
    {P}^{(j)}_{i\mathbb{J}}    =
    \frac{\partial W^{(j)}}{\partial F_{i\mathbb{J}}}
    =    \frac{\partial\boldsymbol{Q}^{T}}
         {\partial F_{i\mathbb{J}}} \boldsymbol{\theta}^{(j)} - pJF_{i\mathbb{J}}^{-T}
    \label{eq_PK1_elementwise_lagrange}
\end{equation}

For the thin specimens subjected to in-plane biaxial loading, the plane-stress condition \citep{Abbasi2026} is further imposed through setting \(P_{33}=0\), \(i=3\) and \(\mathbb{J}=3\) in Eq.~\eqref{eq_PK1_elementwise_lagrange} to yield
\begin{equation}
    p = F_{33}~\frac{\partial\boldsymbol{Q}^{T}}
         {\partial F_{33}} \boldsymbol{\theta}^{(j)}
    \label{eq_lagrange_multiplier}
\end{equation}
Substituting the expression for Lagrange multiplier from Eq.~\eqref{eq_lagrange_multiplier} in Eq.~\eqref{eq_PK1_elementwise_lagrange}, the first Piola-Kirchoff stress tensor under plane-stress condition can be expressed as
\begin{equation}
    {P}^{(j)}_{i\mathbb{J}}    = \left(
    \frac{\partial\boldsymbol{Q}^{T}} {\partial F_{i\mathbb{J}}}  - F_{33}~\frac{\partial\boldsymbol{Q}^{T}}
         {\partial F_{33}} F_{i\mathbb{J}}^{-T}\right) \boldsymbol{\theta}^{(j)}
    \label{eq_PK1_elementwise_lagrange_planestress}
\end{equation}

We then use the expression for $P_{i\mathbb{J}}$ from \cref{eq_PK1_elementwise_lagrange_planestress} in \cref{eq_discrete_weak_form} to obtain $\boldsymbol{A}^{\text{free}}$, $\boldsymbol{A}^{\text{fix}}$ and the objective function as per steps outlined in \cref{subsecEqGap}.

\subsection{Computational data generation}

We performed finite element (FE) simulations in Abaqus/CAE 2025 to generate synthetic full-field displacement data for specimens with prescribed spatially varying material properties. The three-dimensional cases were modeled using a cubic domain of dimensions \(50\times50\times50~\mathrm{mm}^3\), whereas the synthetic in-plane deformation case was represented by a thin square plate specimen of dimensions \(50\times50\times1~\mathrm{mm}^3\). Throughout, we denote the in-plane/domain length by \(L=50~\mathrm{mm}\). The specimens were subjected to quasi-static displacement-controlled tensile loading comprising equi-biaxial (BT) and uniaxial (UT) loading modes. The computational domain was discretized using eight-node, fully integrated three-dimensional brick elements (C3D8), and spatially varying element-wise material properties were assigned through an Abaqus User Material (UMAT) implementation (refer supplementary section S4) of the prescribed constitutive law. For the cubic specimens, six loading steps were considered, corresponding to \(\mathrm{BT}_{x-y}\), \(\mathrm{BT}_{x-z}\), \(\mathrm{BT}_{y-z}\), \(\mathrm{UT}_{x}\), \(\mathrm{UT}_{y}\), and \(\mathrm{UT}_{z}\) as listed in Table~\ref{table_loading_conditions}. The nodal displacements and the resultant normal reaction forces on the loaded boundaries were stored for subsequent inverse identification. The constitutive models and the prescribed spatial variation of their material parameters as a function of reference coordinates \(\theta(X,Y,Z)\), for the different test cases are summarized below.

\begin{enumerate}[label=\textbf{\Alph*)}]
    \item \textbf{Neo-Hookean model (NH2/NH)} \citep{Treloar1943}
    \begin{itemize}
        \item \textit{Three-dimensional smooth checkerboard heterogeneity} (refer \cref{case_3D_checker_NH})
                \begin{equation}
                     W_{\text{NH2}} = \theta_{\mathrm{C10}}(\tilde{I}_{1}-3)
                                       + \theta_{\mathrm{1/D1}} \left(J-1\right)^{2}
                \label{eq_NH2}
                \end{equation}
                where,
                \begin{equation}
                \begin{split}
                    \theta_{\mathrm{C10}}(X,Y,Z)  &= 3 +  2 \cos\left(\frac{2 \pi X}{L}\right)  \cos\left(\frac{2 \pi Y}{L}\right) \cos\left(\frac{2 \pi Z}{L}\right) \\
                    \theta_{\mathrm{1/D1}}(X,Y,Z) &= 30 + 20 \cos\left(\frac{2 \pi X}{L}\right)  \cos\left(\frac{2 \pi Y}{L}\right) \cos\left(\frac{2 \pi Z}{L}\right)
                \end{split}
                \label{eq_3D_checker_NH}
                \end{equation}
                
        \item \textit{In-plane incompressible smooth checkerboard heterogeneity} (refer \cref{case_exp} and Fig.~\ref{fig_2D_checker})
                \begin{equation}
                     W_{\text{NH}} = \theta_{\mathrm{C10}}(\tilde{I}_{1}-3) - p(J-1)                        
                \label{eq_NH}
                \end{equation}
                where,
                \begin{equation}
                \begin{split}
                    \theta_{\mathrm{C10}}(X,Y)  &= 3 +  2 \sin\left(\frac{3 \pi X}{L}\right)  \cos\left(\frac{3 \pi Y}{L}\right)
                \end{split}
                \label{eq_2D_checker_NH}
                \end{equation}        
    \end{itemize}
    
    \item \textbf{Mooney-Rivlin model (MR)} \citep{Mooney1940,Rivlin1948}
    \begin{itemize}
        \item \textit{Smoothly varying spherical inclusion} (refer \cref{case_3D_tanhsph_MR})
                \begin{equation}
                     W_{\text{MR}} = \theta_{\mathrm{C10}}(\tilde{I}_{1}-3)
                                       + \theta_{\mathrm{C01}}(\tilde{I}_{2}-3)
                                       + \theta_{\mathrm{1/D1}} \left(J-1\right)^{2}
                \label{eq_MR}
                \end{equation}
                where,
                \begin{equation}
                \begin{split}
                    \theta_{\mathrm{C10}}(X,Y,Z)  &= 4- 2\tanh\left(\frac{R-12}{4} \right) \\
                    \theta_{\mathrm{C01}}(X,Y,Z)  &= 2 - \tanh\left(\frac{R-12}{4} \right) \\
                    \theta_{\mathrm{1/D1}}(X,Y,Z) &= 40 - 20\tanh\left(\frac{R-12}{4} \right)
                    % \mathrm{where,} \ R  &= \left[(x-L/2)^2 + (y-L/2)^2 + (z-L/2)^2\right]^{1/2}
                \end{split}
                \label{eq_3D_tanhsph_MR}
                \end{equation}
                where the radial coordinate from the center of the specimen is given in \cref{eq_radial_coordinate} as
                \begin{equation}
                R =\left[(X-L/2)^2 + (Y-L/2)^2 + (Z-L/2)^2\right]^{1/2}
                \label{eq_radial_coordinate}
                \end{equation}
    \end{itemize}

    \item \textbf{Modified Holzapfel-Gasser-Ogden model (HGO*)} \cite{Holzapfel2000} 
    \begin{itemize}
        \item  \textit{Anisotropic graded spherical inclusion} (refer \cref{case_3D_tanhsph_NH_HGO})
                \begin{equation}
                     W_{\text{HGO*}} = \theta_{\mathrm{C10}}(\tilde{I}_{1}-3)
                                       + \theta_{k1}[e^{(\tilde{I}_{a}-1)^2}-1]/2
                                       + \theta_{\mathrm{1/D1}} \left(J-1\right)^{2}
                \label{eq_HGO}
                \end{equation}
                where,
                \begin{equation}
                \begin{split}
                    % \theta_{\mathrm{C10}}(X,Y,Z)  &= 3 + \tanh(R-18) \\
                    % \theta_{k1}(X,Y,Z)   &= 1 - \tanh(R-18) \\
                    % \theta_{\mathrm{1/D1}}(X,Y,Z) &= 30 + 10\tanh(R-18) \\
                    \theta_{\mathrm{C10}}(X,Y,Z)  &= 3 + \tanh\left(\frac{R-18}{1.0} \right)\\
                    \theta_{k1}(X,Y,Z)   &= 1 - \tanh\left(\frac{R-18}{1.0} \right) \\
                    \theta_{\mathrm{1/D1}}(X,Y,Z) &= 30 + 10\tanh\left(\frac{R-18}{1.0} \right) \\
                \end{split}
                \label{eq_3D_tanhsph_HGO}
                \end{equation}
    \end{itemize}
\end{enumerate}

\begin{table}[H]
    \centering
    \begin{tabular}{c l l l}
    \toprule
    \textbf{Load step} & \textbf{Loading mode} & \textbf{Prescribed displacement} & \textbf{Nominal stretch} \\
    \midrule
    1 & $\mathrm{BT}_{x-y}$ &
    $u_x = u_y=50~\mathrm{mm}$ &
    $\lambda_x=\lambda_y=2.0$ \\
    
    2 & $\mathrm{BT}_{x-z}$ &
    $u_x = u_z=60~\mathrm{mm}$ &
    $\lambda_x=\lambda_z=2.2$ \\
    
    3 & $\mathrm{BT}_{y-z}$ &
    $u_y = u_z=40~\mathrm{mm}$ &
    $\lambda_y=\lambda_z=1.8$ \\
    
    4 & $\mathrm{UT}_{x}$ &
    $u_x=40~\mathrm{mm}$ &
    $\lambda_x=1.8$ \\
    
    5 & $\mathrm{UT}_{y}$ &
    $u_y=50~\mathrm{mm}$ &
    $\lambda_y=2.0$ \\
    
    6 & $\mathrm{UT}_{z}$ &
    $u_z=60~\mathrm{mm}$ &
    $\lambda_z=2.2$ \\
    \bottomrule
    \end{tabular}
    \caption{Prescribed displacement-controlled loading for the 3D FE simulations. The initial specimen length is $L=50~\mathrm{mm}$, and the nominal stretch is defined as $\lambda_i = 1 + u_i/L$. Directions not listed under the prescribed loading are left unconstrained.}
    \label{table_loading_conditions}
\end{table}

\subsection{Experimentation and measurements}
\label{methods_exp}

\begin{figure}[H]
        \centering
        \includegraphics[width=\linewidth]{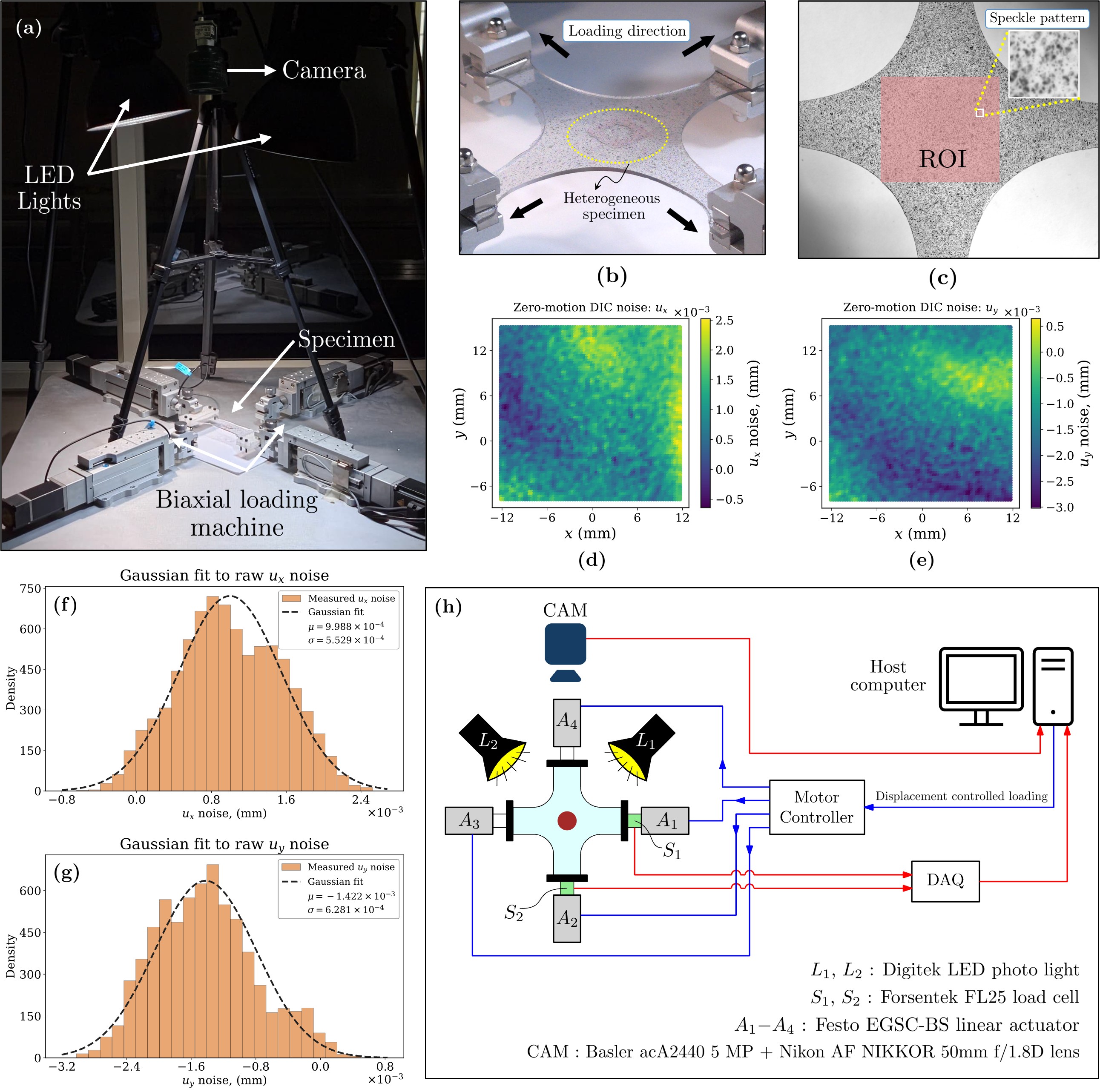}
        \caption{\textbf{Experimental setup and characterization of DIC measurement noise.} \textbf{a}, Biaxial testing setup comprising the loading system, specimen, camera, and LED illumination. \textbf{b}, Heterogeneous cruciform specimen mounted in the biaxial loading device, with the principal loading directions indicated. \textbf{c}, Region of interest (ROI) used for digital image correlation (DIC), with an enlarged view of the speckle pattern. \textbf{d} and \textbf{e}, Spatial distributions of the zero-motion DIC noise in the \(u_x\) and \(u_y\) displacement components, respectively. \textbf{f} and \textbf{g}, Corresponding noise distributions with Gaussian fits, illustrating the measured noise distribution in each in-plane direction. \textbf{h}, Schematic representation of the experimental system showing the camera, LED illumination, linear actuators, load cells, motor controller, data-acquisition (DAQ) unit, and host computer used for synchronized displacement-controlled biaxial testing and image acquisition.}
        \label{fig_exp_schematic}
    \end{figure}

For experimental validation, heterogeneous specimens were prepared using PDMS with different base-to-curing-agent ratios to introduce material contrast. A 15:1 ratio was used for the softer matrix, while 5:1 and 10:1 mixtures were used to prepare the stiffer regions. The mixtures were stirred, vacuum-degassed for approximately 30-45 minutes, and cast into laser-cut acrylic molds. We first cured the matrix material at \(100^{\circ}\mathrm{C}\) for 1 hour, following which the preformed cavities were filled with the inclusion material and cured again. For biaxial testing, a custom cruciform-type specimen geometry was designed with uniform thickness. The biaxial specimen consisted of four loading arms and a central region of interest (ROI) to accommodate different heterogeneous configurations, as highlighted in Fig.~\ref{fig_exp_schematic}b. A fine random speckle pattern was applied to the upper surface using black spray paint diluted with acetone. The speckle density of around 30–35\% of the surface was selected to provide sufficient grayscale variation for accurate DIC correlation. The experimental arrangement is shown in Fig.~\ref{fig_exp_schematic}a-c, with a detailed schematic of the instrumentation provided in Fig.~\ref{fig_exp_schematic}h. The specimen was mounted on a displacement-controlled biaxial testing machine equipped with four Festo EGSC-BS linear actuators and subjected to cycles of non-equibiaxial loading and unloading at an actuation speed of 0.1 mm/s. The corresponding reaction forces along the two orthogonal loading directions were recorded using Forsentek FL25 load cells connected to a data acquisition (DAQ) system. Images of the deforming specimen were captured using a Basler acA2440 (5 MP) camera fitted with a Nikon AF NIKKOR 50 mm f/1.8D lens, positioned normal to the specimen surface as shown in Fig.~\ref{fig_exp_schematic}a. Two digital LED photo lights were arranged around the specimen to maintain adequate illumination during image acquisition. For each cruciform specimen, companion dog-bone-shaped specimens were also prepared for every constituent material formulation, using the same PDMS batches and curing conditions as the corresponding material regions. The dog-bone geometry was selected based on ASTM D412-16(2021), Test Method A, \citep{sheets2012standard} for tensile testing of elastomers. Independent uniaxial tensile tests were performed on these specimens, and the measured first Piola--Kirchhoff stress--stretch responses were fitted using the incompressible Neo-Hookean model to determine \(\theta_{\mathrm{C10}}\) as reported in supplementary section S3.5. The fitted values are compared with the mean material parameters obtained from the inverse identification in supplementary Table~S1. \par

We acquired the deformed specimen images using a frame rate of 5 frames per second (\textit{fps}) during loading. The images were subsequently processed for subset-based tracking using VIC-2D software \citep{vic2dmanual7} to obtain the in-plane displacement measurements over the ROI, as illustrated in Fig.~\ref{fig_exp_schematic}c. The measured coordinates and surface displacement components \(u_x\) and \(u_y\) were exported to construct the FE mesh with C3D8 type elements, nodal displacement data, and other inputs required by our inverse identification framework. To improve the aspect ratio of the one-element-thick C3D8 discretization, the through-thickness coordinates were rescaled by \(s_z=\sqrt{\bar A}/d_{t}\), where \(\bar A\) is the mean in-plane element area and \(d_{t}\) is the specimen thickness. The measured in-plane reaction forces were also scaled by the same factor to maintain consistency, while the in-plane displacement field remained unchanged. Prior to loading, a zero-motion DIC analysis was also performed by recording images of the specimen under stationary, unloaded conditions to estimate the measurement noise. Since the specimen remained undeformed, any apparent displacement obtained from the correlation analysis provides an estimate of the DIC noise. The spatial distributions of the measured \(u_x\) and \(u_y\) noise are shown in Fig.~\ref{fig_exp_schematic}d and \ref{fig_exp_schematic}e representing spatially correlated noise, with the corresponding Gaussian fits presented in Fig.~\ref{fig_exp_schematic}f and \ref{fig_exp_schematic}g. The fitted mean values were \(9.988\times10^{-4}\) mm and \(-1.422\times10^{-3}\) mm, with standard deviations of \(5.529\times10^{-4}\) mm and \(6.281\times10^{-4}\) mm for the two displacement components, respectively. Based on these experimental measurements, a conservatively higher noise magnitude of \((\sigma_{\boldsymbol{u}}=1\times10^{-3})\) was used during synthetic data generation to examine the robustness of the proposed inverse framework under noise-perturbed conditions. \par

\section{Results and discussion}
\label{sec_results}

The schematic of the proposed framework is illustrated in Fig.~\ref{fig_graphical_abstract}. For a prescribed constitutive model, the objective is to identify the material parameters at each distinct element in the geometry. The specimen with an initially unknown material-property distribution is subjected to a displacement-controlled loading at its boundaries with a combination of biaxial loading in the \(x\)-\(y\), \(x\)-\(z\), and \(y\)-\(z\) planes, together with uniaxial loading along each principal direction, providing rich deformation-state data for inverse identification. The full-field displacement measurements are obtained for the entire geometry. In the pre-processing stage, the domain is discretized using eight-node brick elements (C3D8), and the full-field displacement data are mapped to the corresponding nodes. The matrix \(\boldsymbol{A}^{\mathrm{free}}\) is then constructed using the displacement-field and the derived deformation gradient field. The inverse problem is subsequently formulated as the minimization of the objective function (see Eq.~(\ref{eq_Lossfunction})), resulting in the relative constitutive parameter field \(\boldsymbol{\theta}_{\mathrm{opt}}\). In the absence of force information, this field is identifiable only up to a global multiplicative constant \citep{Roux2020}. When the measured boundary force is available, the force balance over the traction boundary is used to determine the scaling factor \(\eta\). The absolute material parameter field is then recovered as \(\boldsymbol{\theta}=\eta\,\boldsymbol{\theta}_{\mathrm{opt}}\).

The inverse problem is solved using both-- noiseless displacement data and displacement fields perturbed with spatially varying heteroscedastic Gaussian noise of magnitude, \(\sigma_{\boldsymbol{u}}=1\times10^{-3}\) (supplementary section S2). For all case studies presented here, we only consider the highest level of spatially correlated noise magnitude reported by Taç et al. \citep{Ta2026} and Krijnen et al. \citep{Krijnen2026}. We minimized the loss function defined in Eq.~\eqref{eq_Lossfunction} using the Adam (Adaptive Moment Estimation) optimizer \citep{Kingma2014}. All optimizations are performed on an NVIDIA RTX 5000 Ada Generation GPU with 32 GB of GPU memory.\par
    
\subsection{Identification of complex three-dimensional Neo-Hookean heterogeneity}
\label{case_3D_checker_NH}

\begin{figure}[H]
        \centering
        \includegraphics[width=\linewidth]{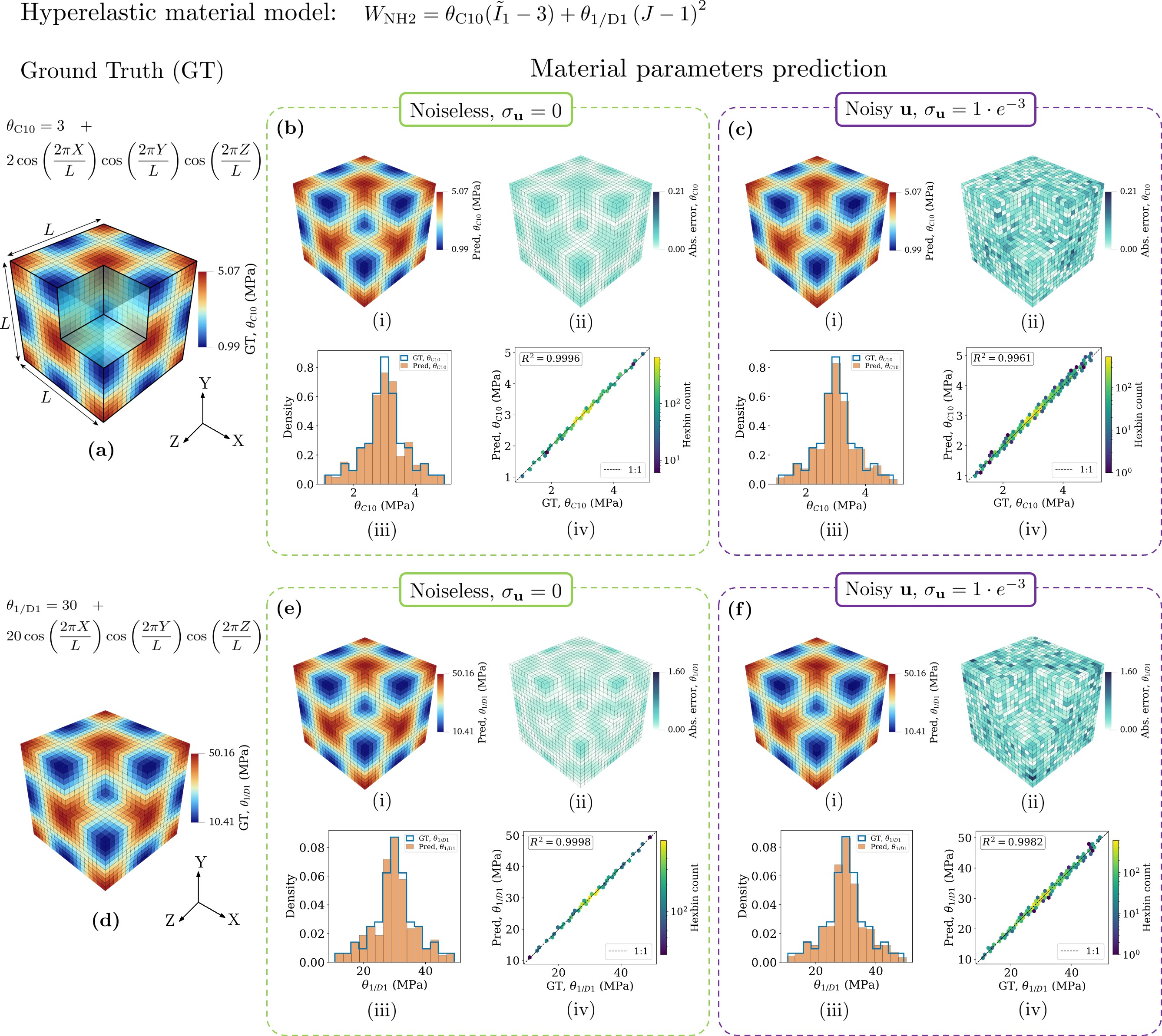}
        \caption{\textbf{Smoothly varying checkerboard heterogeneity for Neo-Hookean material.}  The ground truth (GT) spatial distributions of \(\boldsymbol{\theta}_{\mathrm{C10}}\) and \(\boldsymbol{\theta}_{\mathrm{1/D1}}\) are visualized in \textbf{a} and \textbf{d}, respectively. Figure panels \textbf{b} and \textbf{e} correspond to the inverse identification results for noiseless displacement data,  showing sub-figures (i) the predicted (Pred.) material parameter field, (ii) the absolute error, (iii) comparison of the GT and predicted parameter histogram distributions, and (iv) hexbin parity plots between the ground truth and predicted values. Figure panels \textbf{c} and \textbf{f} highlight the corresponding results for noise-perturbed displacement data, presented using the same quantities. The dashed diagonal line in the parity plots denotes the \(1{:}1\) reference, and the corresponding \(R^2\) values quantify agreement between the predicted and ground truth parameter fields.}
        \label{fig_3D_checker_NH}
    \end{figure}

We first evaluate the proposed framework on a complex three-dimensional material field with smooth, periodic checkerboard-like graded heterogeneity, assuming a compressible Neo-Hookean \citep{Treloar1943} constitutive model (Eq.~\eqref{eq_NH2}). The ground-truth distributions of the two constitutive parameters-- $\theta_{\mathrm{C10}}$  (proportional to shear modulus), and $\theta_{\mathrm{1/D1}}$ (proportional to bulk modulus) are visualized using octant cut-section in Fig.~\ref{fig_3D_checker_NH}a and \ref{fig_3D_checker_NH}d, while their prescribed spatial variations are given in Eq.~\eqref{eq_3D_checker_NH}. The cubic domain is discretized using 8,000 C3D8 elements and subjected to displacement-controlled loading at the boundaries as summarized in Table~\ref{table_loading_conditions}. \par

The Neo-Hookean strain-energy density given by Eq.~\eqref{eq_NH2} contains two independent constitutive terms, \((\tilde{I}_{1}-3)\) and \(\left(J-1\right)^{2}\), and the framework therefore simultaneously identifies the corresponding element-wise parameters \(\boldsymbol{\theta}_{\mathrm{C10}}\) and \(\boldsymbol{\theta}_{\mathrm{1/D1}}\). Fig.~\ref{fig_3D_checker_NH}b(i)-(iv) presents the recovered \(\boldsymbol{\theta}_{\mathrm{C10}}\) field for the noiseless case, including the predicted field, absolute error, histogram comparison, and hexbin parity plot, respectively. The corresponding results obtained from the noise-perturbed displacement field are shown in Fig.~\ref{fig_3D_checker_NH}c(i)-(iv). In both cases, the recovered \(\boldsymbol{\theta}_{\mathrm{C10}}\) field closely follows the ground truth (Fig.~\ref{fig_3D_checker_NH}a), with \(R^2\) values of 0.9996 and 0.9961 for the noiseless and noisy cases, respectively. The recovery of \(\boldsymbol{\theta}_{\mathrm{1/D1}}\) is summarized in Fig.~\ref{fig_3D_checker_NH}e and \ref{fig_3D_checker_NH}f for the noiseless and noisy cases, respectively. The predicted distributions show close agreement with the prescribed ground-truth field in Fig.~\ref{fig_3D_checker_NH}d, as also reflected by the histogram comparisons in Fig.~\ref{fig_3D_checker_NH}e(iii) and \ref{fig_3D_checker_NH}f(iii). The corresponding \(R^2\) values of 0.9998 and 0.9982 further confirm the accuracy of the recovered \(\boldsymbol{\theta}_{\mathrm{1/D1}}\) field. The  optimization is performed for 60,000 epochs, requiring 16.81 minutes and 17.20 minutes for the noiseless and noisy cases, respectively. \par

\subsection{Simultaneous recovery of multiple spatially varying Mooney-Rivlin parameter fields}
\label{case_3D_tanhsph_MR}

\begin{figure}[H]
        \centering
        \includegraphics[width=0.96\linewidth]{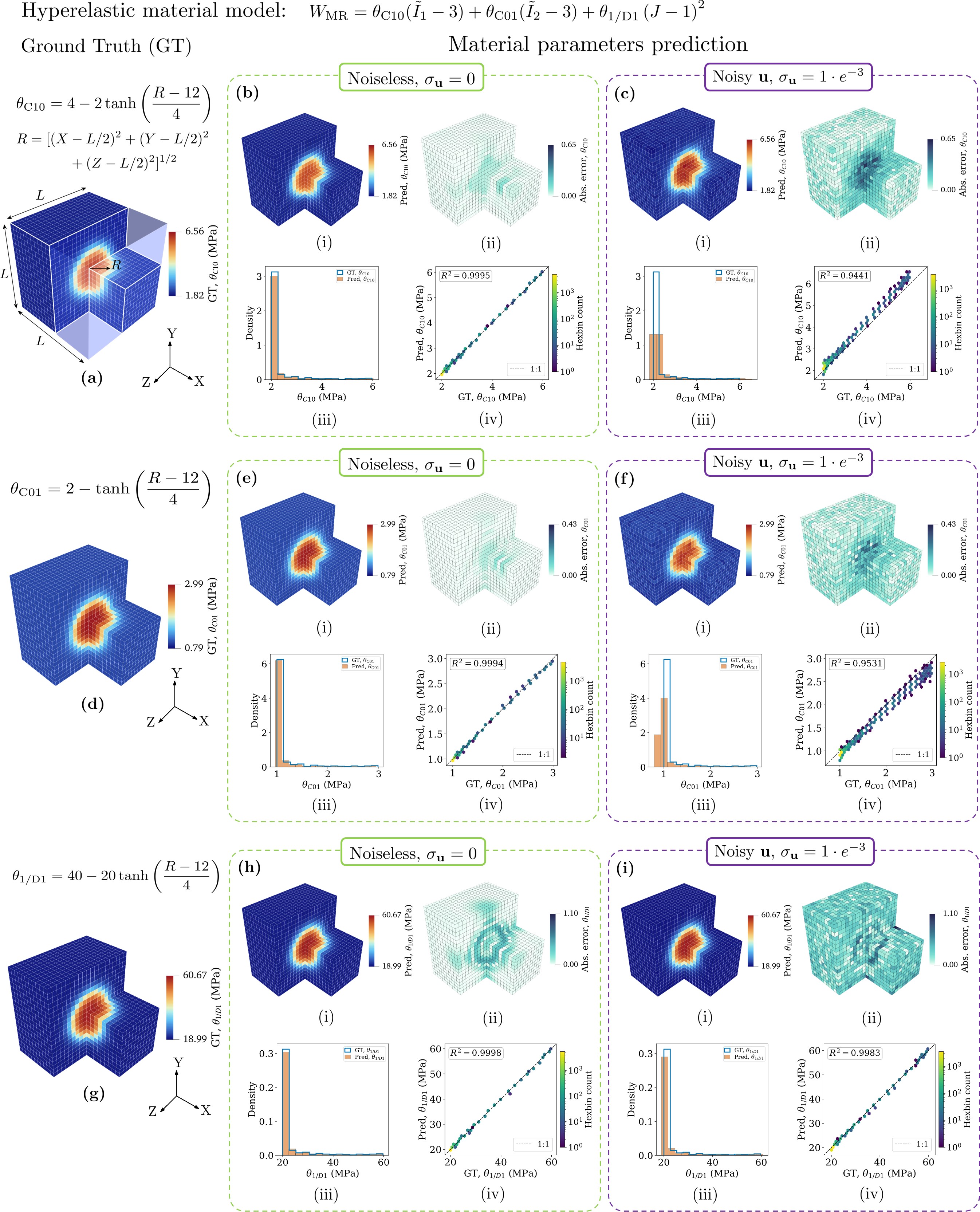}
        \caption{\textbf{Graded spherical inclusion for Mooney-Rivlin material.}  The ground truth (GT) spatial distributions of \(\boldsymbol{\theta}_{\mathrm{C10}}\), \(\boldsymbol{\theta}_{\mathrm{C01}}\) and \(\boldsymbol{\theta}_{\mathrm{1/D1}}\) are visualized in \textbf{a}, \textbf{d} and \textbf{g}, respectively. Figure panels \textbf{b}, \textbf{e} and \textbf{h} correspond to the inverse identification results for noiseless displacement data, showing sub-figures (i) the predicted (Pred.) material parameter field, (ii) the absolute error, (iii) comparison of the GT and predicted parameter histogram distributions, and (iv) hexbin parity plots between the ground truth and predicted values. Figure panels \textbf{c}, \textbf{f} and \textbf{i} highlight the corresponding results for noise-perturbed displacement data, presented using the same quantities. The dashed diagonal line in the parity plots denotes the \(1{:}1\) reference, and the corresponding \(R^2\) values quantify agreement between the predicted and ground truth parameter fields.}
        \label{fig_3D_tanhsph_MR}
    \end{figure}

Next, we extend the analysis to a smoothly graded spherical inclusion embedded within a surrounding matrix, representing a localized three-dimensional material heterogeneity with a gradual transition. Such smooth transitions are particularly relevant to biological tissues in which material properties do not necessarily change abruptly across a distinct boundary. In contrast to the previous Neo-Hookean example, this case is modeled using the Mooney-Rivlin \citep{Mooney1940,Rivlin1948} constitutive law to assess the framework's ability to recover multiple spatially varying constitutive parameter fields simultaneously. The domain is meshed using 8,000 C3D8 elements and subjected to the mixed loading conditions summarized in Table~\ref{table_loading_conditions}. The use of multiple loading modes provides more varied deformation information, which helps the optimizer distinguish the contributions of different deviatoric invariants and improves the inverse method for constitutive identification \citep{Ta2026}. The strain-energy density function in Eq.~\eqref{eq_MR} contains three constitutive terms, \((\tilde{I}_{1}-3)\), \((\tilde{I}_{2}-3)\), and \(\left(J-1\right)^{2}\), with the corresponding element-wise parameters \(\boldsymbol{\theta}_{\mathrm{C10}}\), \(\boldsymbol{\theta}_{\mathrm{C01}}\), and \(\boldsymbol{\theta}_{\mathrm{1/D1}}\), respectively. The parameters are varying simultaneously according to the spatial gradation prescribed using the hyperbolic tangent function defined in Eq.~\eqref{eq_3D_tanhsph_MR}. Their ground-truth distributions are visualized using three octant cut-sections in Fig.~\ref{fig_3D_tanhsph_MR}a, \ref{fig_3D_tanhsph_MR}d, and \ref{fig_3D_tanhsph_MR}g. \par
    
For the noiseless displacement-field case, Fig.~\ref{fig_3D_tanhsph_MR}b, \ref{fig_3D_tanhsph_MR}e, and \ref{fig_3D_tanhsph_MR}h panels summarize the recovered fields and corresponding error estimates for \(\boldsymbol{\theta}_{\mathrm{C10}}\), \(\boldsymbol{\theta}_{\mathrm{C01}}\), and \(\boldsymbol{\theta}_{\mathrm{1/D1}}\), respectively. The corresponding \(R^2\) values of 0.9995, 0.9994, and 0.9998 indicate that all three parameter fields are simultaneously recovered with high accuracy for an optimization time of 29.57 minutes (60,000 epochs). Under the noised-perturbed condition \((\sigma_{\boldsymbol{u}}=1\times10^{-3})\), the corresponding predictions are shown in Fig.~\ref{fig_3D_tanhsph_MR}c, \ref{fig_3D_tanhsph_MR}f, and \ref{fig_3D_tanhsph_MR}i panels. As expected, the absolute errors increase relative to the noiseless case, particularly for \(\boldsymbol{\theta}_{\mathrm{C10}}\) and \(\boldsymbol{\theta}_{\mathrm{C01}}\); nevertheless, the recovered spatial distributions remain in close agreement with the ground truth, with \(R^2\) values of 0.9441, 0.9531, and 0.9983, respectively. The optimization for the noised displacement-field case took 29.49 minutes (60,000 epochs). These results demonstrate that the framework can simultaneously recover multiple spatially varying Mooney-Rivlin parameter fields in three dimensions with acceptable accuracy in the presence of heteroscedastic Gaussian displacement noise. \par
    
\subsection{Anisotropic graded spherical inclusion with spatially varying material parameters}
\label{case_3D_tanhsph_NH_HGO}

\begin{figure}[H]
        \centering
        \includegraphics[width=0.97\linewidth]{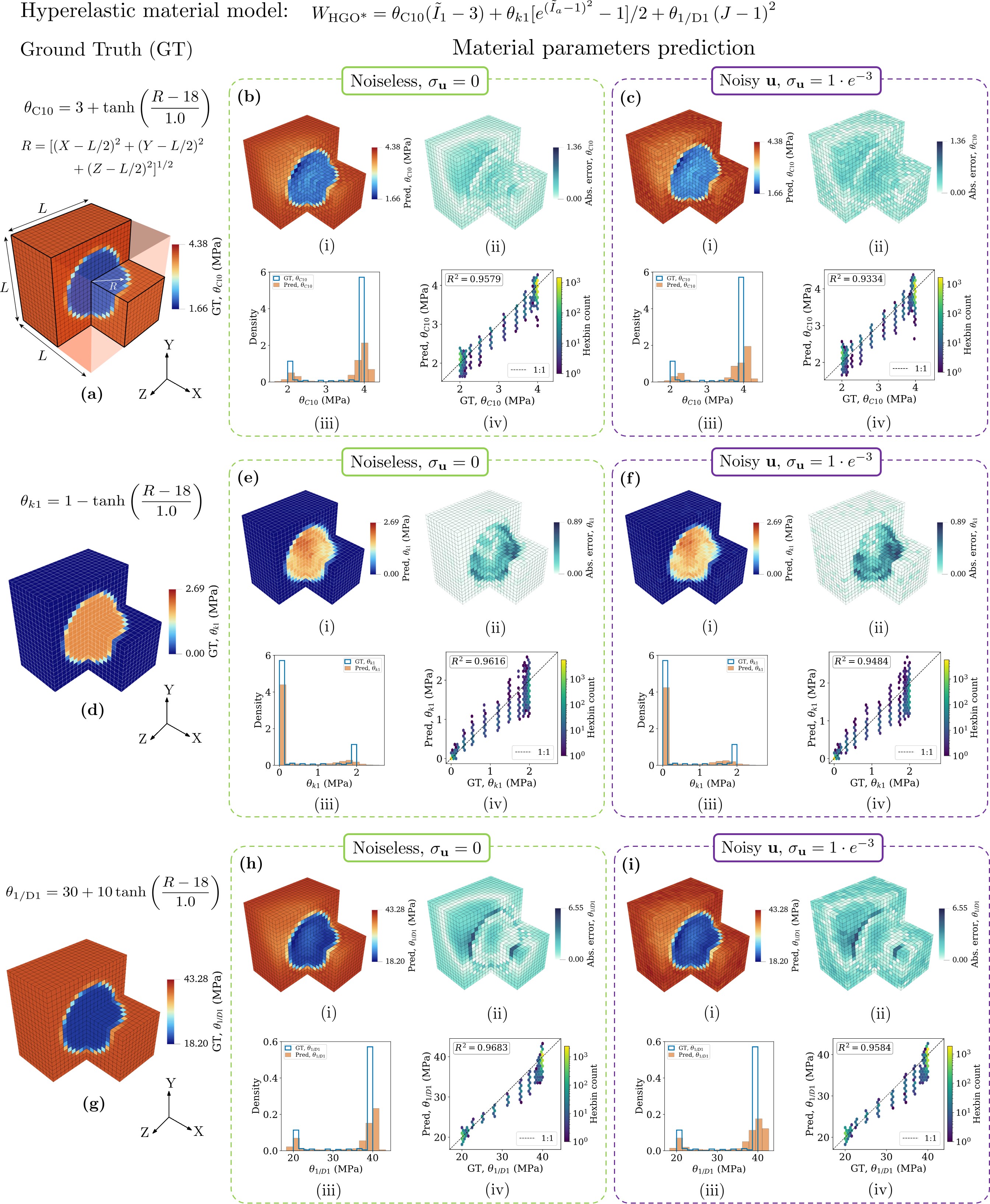}
        \caption{\textbf{Recovering anisotropic graded spherical inclusion.}  The ground truth (GT) spatial distributions of \(\boldsymbol{\theta}_{\mathrm{C10}}\), \(\boldsymbol{\theta}_{k1}\) and \(\boldsymbol{\theta}_{\mathrm{1/D1}}\) are visualized in \textbf{a}, \textbf{d} and \textbf{g}, respectively. Figure panels \textbf{b}, \textbf{e} and \textbf{h} correspond to the inverse identification results for noiseless displacement data, showing sub-figures (i) the predicted (Pred.) material parameter field, (ii) the absolute error, (iii) comparison of the GT and predicted parameter histogram distributions, and (iv) hexbin parity plots between the ground truth and predicted values. Figure panels \textbf{c}, \textbf{f} and \textbf{i} highlight the corresponding results for noise-perturbed displacement data, presented using the same quantities. The dashed diagonal line in the parity plots denotes the \(1{:}1\) reference, and the corresponding \(R^2\) values quantify agreement between the predicted and ground truth parameter fields.}
        \label{fig_3D_tanhsph_NH_HGO}
    \end{figure}

We further investigate the performance of the proposed framework for anisotropic materials using a cubic domain with a graded spherical anisotropic inclusion embedded in an isotropic matrix. To represent this material distribution in ABAQUS, we developed a UMAT subroutine (supplementary section S4) based on a modified Holzapfel-Gasser-Ogden formulation (HGO*) \citep{Holzapfel2000}. The modified HGO* strain-energy function in Eq.~\eqref{eq_HGO} consists of an isotropic isochoric term \((\tilde{I}_{1}-3)\) describing the matrix response, an anisotropic term \([e^{(\tilde{I}_{a}-1)^2}-1]\) associated with the prescribed fiber direction, and a volumetric penalty term \(\left(J-1\right)^{2}\) accounting for compressibility. The fiber direction is prescribed to lie along the body diagonal of the cube defined by the unit vector \(\boldsymbol{a}=(1,1,1)/\sqrt{3}\). The original volumetric strain-energy contribution in the HGO model \citep{Holzapfel2000}, \(\left[(J^2-1)/2-\ln J\right]\), is replaced by the quadratic volumetric penalty \((J-1)^2\) to maintain consistency with the compressible Neo-Hookean formulation (Eq.~\eqref{eq_NH2}) adopted in the present work. The cubic domain is discretized using 8,000 C3D8 elements and subjected to the mixed loading conditions listed in Table~\ref{table_loading_conditions}, with large stretch ratios used to ensure significant nonlinear anisotropic response. The corresponding element-wise constitutive parameters are \(\boldsymbol{\theta}_{\mathrm{C10}}\), \(\boldsymbol{\theta}_{\mathrm{k1}}\), and \(\boldsymbol{\theta}_{\mathrm{1/D1}}\) as defined in Eq.~\eqref{eq_HGO}. Using the hyperbolic tangent variation defined in Eq.~\eqref{eq_3D_tanhsph_HGO}, \(\boldsymbol{\theta}_{\mathrm{k1}}\) is assigned a non-zero value within the spherical inclusion and approaches zero in the interface and is exactly zero in the surrounding matrix, thereby restricting the anisotropic contribution to the inclusion while incorporating the isotropic Neo-Hookean response in the surrounding matrix. This spatial variation is shown in the ground-truth field in Fig.~\ref{fig_3D_tanhsph_NH_HGO}d. \par

Fig.~\ref{fig_3D_tanhsph_NH_HGO}a shows the ground-truth variation of \(\boldsymbol{\theta}_{\mathrm{C10}}\), while the corresponding predictions for the noiseless and noisy cases are presented in Fig.~\ref{fig_3D_tanhsph_NH_HGO}b(i)-(iv) and Fig.~\ref{fig_3D_tanhsph_NH_HGO}c(i)-(iv), respectively. The recovered field follows the prescribed graded spherical distribution, with \(R^2\) values of 0.9579 and 0.9334 for the noiseless and noisy cases, respectively. The recovery of the anisotropic parameter field-- \(\boldsymbol{\theta}_{\mathrm{k1}}\) is summarized in Fig.~\ref{fig_3D_tanhsph_NH_HGO}e and \ref{fig_3D_tanhsph_NH_HGO}f and compared against the ground truth in Fig.~\ref{fig_3D_tanhsph_NH_HGO}d. The corresponding hexbin parity plots give \(R^2\) values of 0.9616 and 0.9484 for the noiseless and noisy cases, respectively. The volumetric parameter field-- \(\boldsymbol{\theta}_{\mathrm{1/D1}}\) is likewise recovered with \(R^2\) values of 0.9683 and 0.9584 for the noiseless and noisy cases, respectively, as shown in Fig.~\ref{fig_3D_tanhsph_NH_HGO}h and \ref{fig_3D_tanhsph_NH_HGO}i. The optimization times are 73.70 minutes (noiseless case) and 72.89 minutes (noisy case) for 150,000 epochs, indicating that the presence of displacement noise does not noticeably increase the computational cost. \par

Recovering the anisotropic material field is more challenging compared to the isotropic case studies. This increased complexity is reflected in the comparatively reduced \(R^2\) values and longer optimization time observed for the anisotropic case. Nevertheless, the spatial distributions of all three parameters are recovered with acceptable accuracy in both the noiseless and noisy cases, demonstrating the applicability of our framework to three-dimensional anisotropic, graded hyperelastic materials. Further, our framework accurately identifies $\theta_{\text{k1}}$ to be zero for elements that are supposed to have isotropic material behavior. \par
    
\subsection{Local 3D elastography using displacement measurements only}
\label{case_3D_tanhsph_MR_local}

\begin{figure}[H]
        \centering
        \includegraphics[width=\linewidth]{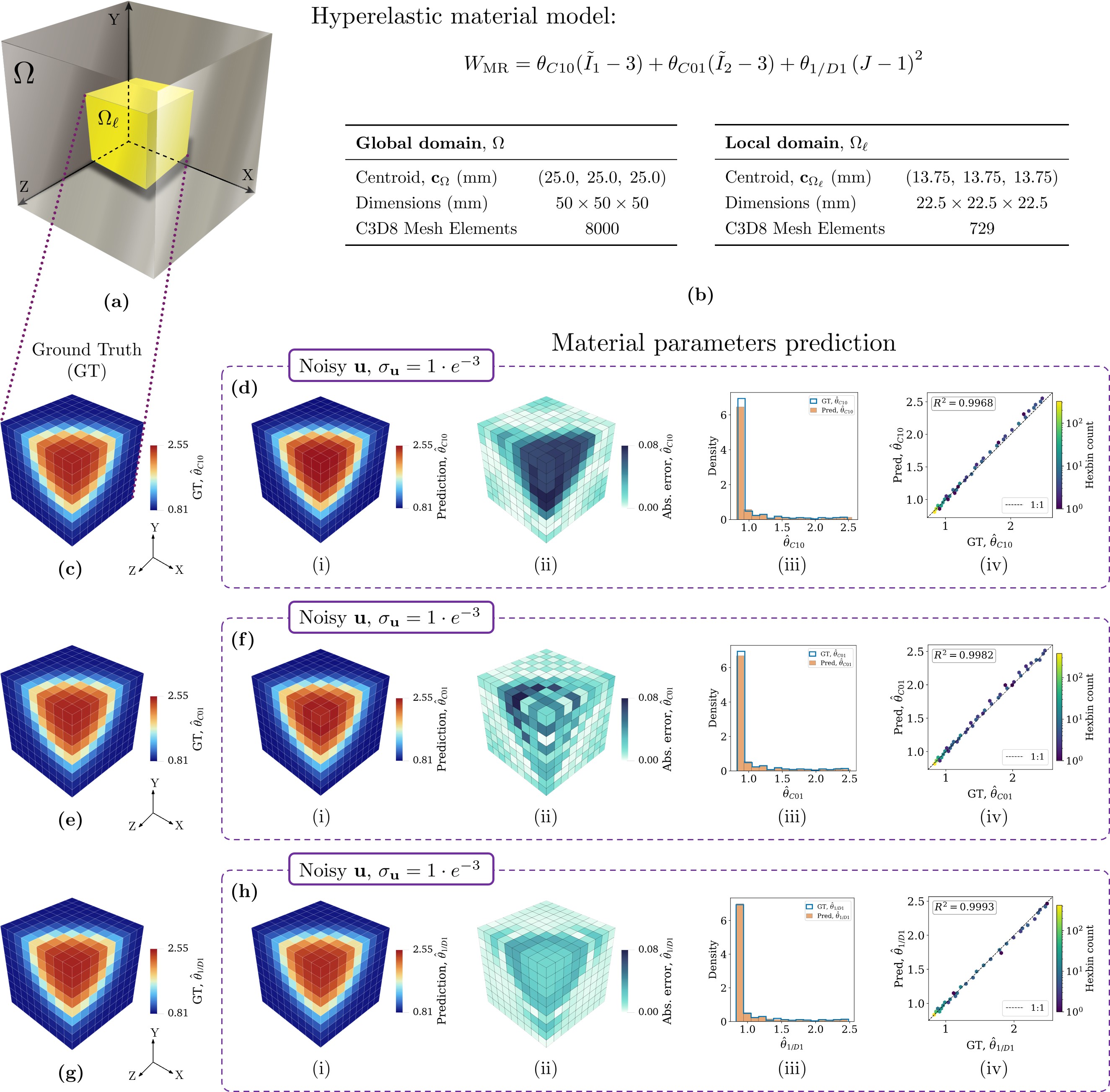}
        \caption{\textbf{Local 3D elastography for graded spherical inclusion for Mooney-Rivlin material.} \textbf{a} highlights the size, shape, and location of the local domain \(\Omega_{\ell}\) within the cubical specimen denoted by \(\Omega\). \textbf{b} tabulates the centroid, dimensions and mesh information of local domain \(\Omega_{\ell}\) with respect to the global domain \(\Omega\). The ground truth (GT) spatial distributions of normalized parameter fields \(\boldsymbol{\hat{\theta}}_{\mathrm{C10}}\), \(\boldsymbol{\hat{\theta}}_{\mathrm{C01}}\), and \(\boldsymbol{\hat{\theta}}_{\mathrm{1/D1}}\) are visualized in \textbf{c}, \textbf{e} and \textbf{g}, respectively. Figure panels \textbf{d}, \textbf{f} and \textbf{h} correspond to the inverse identification results for noisy displacement data, showing sub-figures (i) the predicted (Pred.) material parameter field, (ii) the absolute error, (iii) comparison of the GT and predicted parameter histogram distributions, and (iv) hexbin parity plots between the GT and predicted values. The dashed diagonal line in the parity plots denotes the \(1{:}1\) reference, and the corresponding \(R^2\) values quantify agreement between the predicted and true parameter fields.}
        \label{fig_3D_tanhsph_MR_local}
    \end{figure}

In this case study, we examine whether the proposed framework can recover the relative constitutive parameter fields within a selected region of the specimen using only nodal displacement measurements, without requiring boundary-force information. We refer to this as \textit{local elastography}, where the material-property contrast is identified up to a global multiplicative constant. As discussed in Eqs.~\eqref{eq_theta_elastofactor} and \eqref{eq_elastofactor}, absolute parameter values can be obtained when the corresponding force information is available to determine the scaling factor \(\eta\). For the present analysis, we consider the smoothly graded spherical inclusion governed by the Mooney-Rivlin constitutive law, previously introduced in case study \ref{case_3D_tanhsph_MR}. However, unlike the earlier example, the inverse identification is now restricted to a smaller local domain \(\Omega_{\ell}\) within the global domain \(\Omega\), following the force-free identification procedure illustrated in Fig.~\ref{fig_graphical_abstract}a-i. Fig.~\ref{fig_3D_tanhsph_MR_local}a shows the global cubic domain \(\Omega\) of dimensions 50~mm \(\times\) 50~mm \(\times\) 50~mm, centred at \(\mathbf{c}_{\Omega} (\mathrm{mm})=(25.0,25.0,25.0)\). The global domain is discretized using 8,000 C3D8 elements, and the nodal displacement fields are generated under the loading conditions listed in Table~\ref{table_loading_conditions}. From this domain \(\Omega\), we select a smaller cubic region \(\Omega_{\ell}\), highlighted in yellow in Fig.~\ref{fig_3D_tanhsph_MR_local}a, with dimensions 22.5~mm \(\times\) 22.5~mm \(\times\) 22.5~mm and centroid \(\mathbf{c}_{\Omega_{\ell}} (\mathrm{mm})=(13.75,13.75,13.75)\), measured with respect to the global coordinate system. Although the local domain can be chosen with different geometries, a cubic region is used here to provide a convenient visualization of the recovered parameter fields. The selected region \(\Omega_{\ell}\), contains 729 C3D8 elements, and the corresponding nodal displacement data are extracted from the global dataset. Using these (noise-affected) displacement measurements, we construct the \(\boldsymbol{A}^{\mathrm{free}}\) matrix and minimize the objective function in Eq.~\eqref{eq_Lossfunction}. Since the prescribed element-wise constitutive parameter fields (\(\boldsymbol{\theta}_{\mathrm{C10}}\), \(\boldsymbol{\theta}_{\mathrm{C01}}\), \(\boldsymbol{\theta}_{\mathrm{1/D1}}\)) in Eqs.~\eqref{eq_MR} and \eqref{eq_3D_tanhsph_MR} represent absolute values, we normalize both the ground-truth and recovered parameter fields according to Eq.~\eqref{eq_theta_cap}. This allows us to compare the ground-truth and predicted relative parameter fields.
\begin{equation}
    \boldsymbol{\hat{\theta}}_{\mathrm{C10}}=\dfrac{\boldsymbol{\theta}_{\mathrm{C10}}}{(\theta_{\mathrm{C10}})_{\mathrm{mean}}} \ ; \quad \boldsymbol{\hat{\theta}}_{\mathrm{C01}}=\dfrac{\boldsymbol{\theta}_{\mathrm{C01}}}{(\theta_{\mathrm{C01}})_{\mathrm{mean}}} \ ; \quad \boldsymbol{\hat{\theta}}_{\mathrm{1/D1}}=\dfrac{\boldsymbol{\theta}_{\mathrm{1/D1}}}{(\theta_{\mathrm{1/D1}})_{\mathrm{mean}}}
    \label{eq_theta_cap}
\end{equation}

The optimization is performed for 60,000 epochs, requiring 8.28 minutes and returns the relative constitutive parameter field \(\boldsymbol{\theta}_{\mathrm{opt}}\), which is subsequently normalized following Eq.~\eqref{eq_theta_cap}. The recovered fields \(\boldsymbol{\hat{\theta}}_{\mathrm{C10}}\), \(\boldsymbol{\hat{\theta}}_{\mathrm{C01}}\), and \(\boldsymbol{\hat{\theta}}_{\mathrm{1/D1}}\) are shown in Fig.~\ref{fig_3D_tanhsph_MR_local}c, \ref{fig_3D_tanhsph_MR_local}e, and \ref{fig_3D_tanhsph_MR_local}g, respectively, alongside their corresponding normalized ground-truth distributions in Fig.~\ref{fig_3D_tanhsph_MR_local}d, \ref{fig_3D_tanhsph_MR_local}f, and \ref{fig_3D_tanhsph_MR_local}h. The histogram comparisons and hexbin parity plots show close agreement between the recovered and ground-truth distributions, with \(R^2\) values of 0.9968, 0.9982, and 0.9993, respectively. Notably, these results indicate better constitutive field recovery accuracy when compared to the original case study in \cref{case_3D_tanhsph_MR}. This is because, as encountered in typical DVC settings \citep{DallAra2022, Krijnen2026}, the spatial correlation induces noises significantly higher than \(\sigma_{\boldsymbol{u}}=1\times10^{-3}\) (supplementary section S2) at the boundaries. Thus, when considering sub-domains containing less severe noise-afflicted nodes we observe improved accuracy in the predicted constitutive parameter field. These results demonstrate that the spatial variation of multiple constitutive parameters can be recovered accurately from displacement data confined to a smaller region of interest, without requiring boundary-force measurements or inverse characterization of the entire specimen. \par
    
    \begin{figure}[H]
        \centering
        \includegraphics[width=\linewidth]{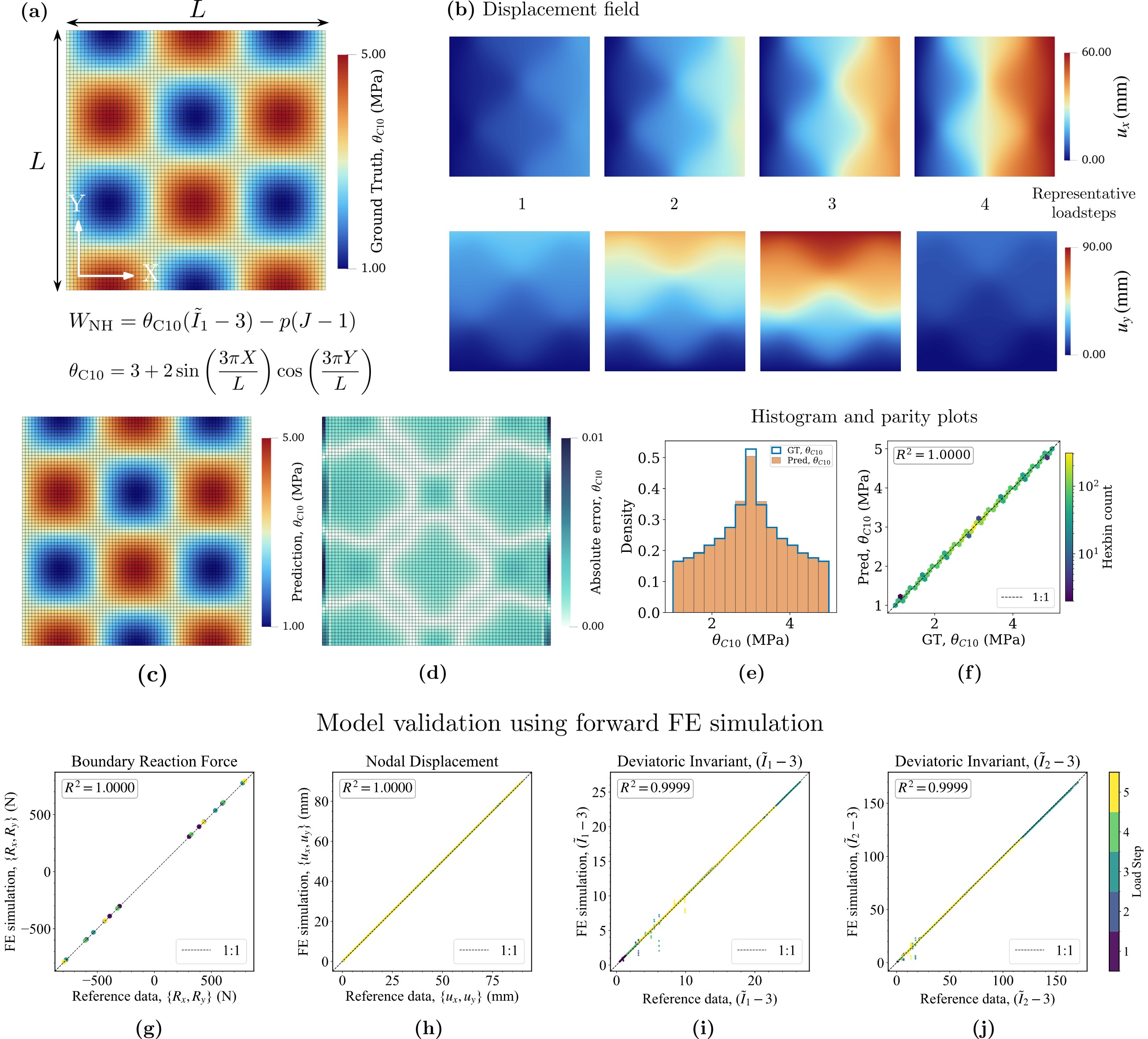}
       \caption{\textbf{Inverse identification and forward validation for the in-plane synthetic checkerboard case.} \textbf{a}, Ground-truth spatial distribution of the Neo-Hookean material parameter \(\boldsymbol{\theta}_{\mathrm{C10}}\). \textbf{b}, In-plane displacement fields across the four selected loading steps (for display), while a total of five loading steps were used for the inverse identification and subsequent forward validation. \textbf{c, d}, Predicted \(\boldsymbol{\theta}_{\mathrm{C10}}\) field and the corresponding absolute error, respectively. \textbf{e}, Comparison of the ground-truth (GT) and predicted parameter distributions. \textbf{f}, Hexbin parity plot between the ground-truth and predicted values. \textbf{g--j}, Forward finite-element (FE) validation comparing the reference and predicted responses in terms of boundary reaction force, nodal displacement, and the deviatoric invariants \((\tilde I_1-3)\) and \((\tilde I_2-3)\), respectively. Colors in the validation plots distinguish the all five load steps. The dashed diagonal lines denote the \(1{:}1\) reference, and the reported \(R^2\) values quantify agreement between the corresponding reference and predicted quantities.}
        \label{fig_2D_checker}
    \end{figure}
    
\subsection{Experimental validation using two-dimensional DIC surface displacement measurements}
\label{case_exp}

To validate the proposed method experimentally, we performed biaxial tension tests on polydimethylsiloxane (PDMS) samples and obtained the surface displacement fields using the two-dimensional DIC setup described in Methods \cref{methods_exp}. Since the DIC measurements provide only the in-plane displacement components-- \(u_x\) and \(u_y\), it is not possible to directly evaluate all components of the three-dimensional deformation gradient. Following the strategy adopted in Abbasi et al.~\citep{Abbasi2026}, we leverage material incompressibility assumption alongside a plane-stress condition and construct the deformation gradient given in Eq.~\eqref{eq_defGrad}. Before applying this formulation to experimental measurements, we verify it using a synthetic in-plane deformation problem. A thin square plate of dimensions \(50\times50\times1~\mathrm{mm}^3\), shown in Fig.~\ref{fig_2D_checker}a, is discretized as a one-element-thick domain using 5,041 C3D8 elements and subjected to non-equibiaxial tensile loading. The corresponding in-plane displacement fields for the prescribed load steps are shown in Fig.~\ref{fig_2D_checker}b. Under the incompressibility assumption, the Neo-Hookean constitutive model in Eq.~\eqref{eq_NH} is adopted, with the spatially varying parameter \(\boldsymbol{\theta}_{\mathrm{C10}}\) prescribed according to the checkerboard distribution in Eq.~\eqref{eq_2D_checker_NH}. For the 2D synthetic case study, minimization of the loss function in Eq.~\eqref{eq_Lossfunction} requires 4.64 minutes (60,000 epochs). The recovered and prescribed material fields are shown in Fig.~\ref{fig_2D_checker}c. The corresponding absolute error is presented in Fig.~\ref{fig_2D_checker}d, while the histogram and hexbin parity plots in Fig.~\ref{fig_2D_checker}e and \ref{fig_2D_checker}f show near-perfect agreement with an \(R^2\) value of 1.000. \par

For the experimental studies that follow, we adopt an additional forward finite element (FE) validation step to assess whether the assumed constitutive model, along with the predicted parameter field, provides a consistent representation of the measured response. This procedure is first verified using the 2D synthetic case. The element-wise predicted \(\boldsymbol{\theta}_{\mathrm{C10}}\) field is assigned directly to the finite element model through a UMAT subroutine, while the known nodal displacement of the boundary edge nodes (not internal nodes) is imposed directly as the Dirichlet boundary condition for the forward simulation in Abaqus/CAE 2025. The resulting model response is compared with the corresponding synthetic reference quantities in Fig.~\ref{fig_2D_checker}g--j. The boundary reaction forces obtained from the forward simulation closely reproduce the ground-truth values, with \(R^2=1.000\) as shown in Fig.~\ref{fig_2D_checker}g, while the nodal displacements are reproduced with the same level of agreement in Fig.~\ref{fig_2D_checker}h. To further compare the deformation state, the deviatoric invariants \((\tilde{I}_{1}-3)\) and \((\tilde{I}_{2}-3)\) computed from the forward FE solution are compared with those obtained directly from the ground-truth displacement field across all six load steps. The corresponding comparisons in Fig.~\ref{fig_2D_checker}i and \ref{fig_2D_checker}j yield \(R^2\) scores of 0.9999 for both invariants. This synthetic verification, therefore, establishes that the recovered element-wise constitutive parameters can be transferred directly to a forward FE model and reproduce the prescribed mechanical response. We use the same validation procedure for the experimental 2D-DIC cases, where the ground-truth constitutive parameter fields are not known \textit{a prioiri}. \par

We conduct three experimental studies on PDMS specimens with increasing levels of material heterogeneity. In each case, the surface displacement fields measured using 2D-DIC are used for inverse identification of the element-wise \(\boldsymbol{\theta}_{\mathrm{C10}}\) field, followed by a forward FE simulation using the recovered \(\boldsymbol{\theta}_{\mathrm{C10}}\) distribution for validation. The forward FE simulation validation metrics for the experimental case-studies are summarized in Table~\ref{table_exp_forward_validation}. The comparison of the simulated and experimentally measured reaction forces, nodal displacement fields and deviatoric invariants provide a direct assessment of whether the recovered element-wise material field, together with the assumed constitutive model, can reproduce the measured deformation within the specimen. Importantly, the boundary reaction forces are not prescribed during the forward FE simulation, and thus comparing the FE simulation's boundary reaction force output with the experimentally observed boundary reaction force provides an additional level of validation.\par

\begin{figure}[H]
        \centering
        \includegraphics[width=\linewidth]{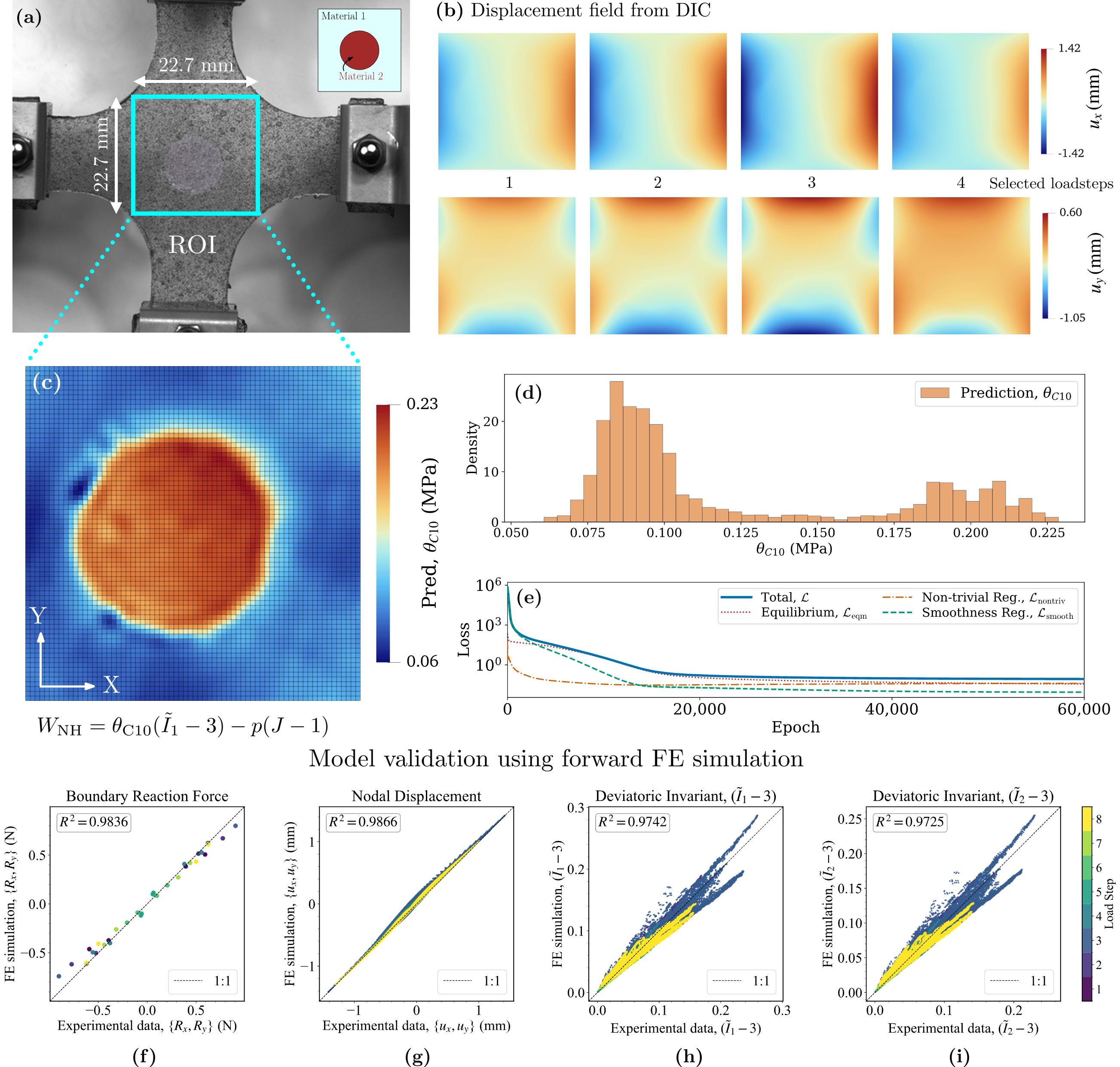}
        \caption{\textbf{Experimental inverse identification and validation of a circular inclusion specimen.} \textbf{a}, Experimental specimen and region of interest (ROI) used for the inverse analysis, with the inset indicating the nominal bi-material heterogeneous distribution. \textbf{b}, In-plane displacement fields obtained from DIC for four representative loading steps (for display). A total of eight loading steps were used for inverse identification and forward FE validation, with the complete dataset accessible as described in the Data availability statement. \textbf{c}, Predicted spatial distribution of the Neo-Hookean material parameter \(\boldsymbol{\theta}_{\mathrm{C10}}\). \textbf{d}, Histogram of the identified \(\theta_{\mathrm{C10}}\) values at each element in the analyzed domain. \textbf{e}, Convergence histories showing the total, equilibrium, regularization, and smoothness loss terms as functions of epoch on a logarithmic scale. \textbf{f--i}, Forward finite-element (FE) validation comparing the experimental and predicted responses in terms of boundary reaction force, nodal displacement, and the deviatoric invariants \((\tilde I_1-3)\) and \((\tilde I_2-3)\), respectively. Colors in the validation plots distinguish the all eight loading steps.}
        \label{fig_2D_circular_inclusion}
    \end{figure}

\begin{figure}[H]
        \centering
        \includegraphics[width=\linewidth]{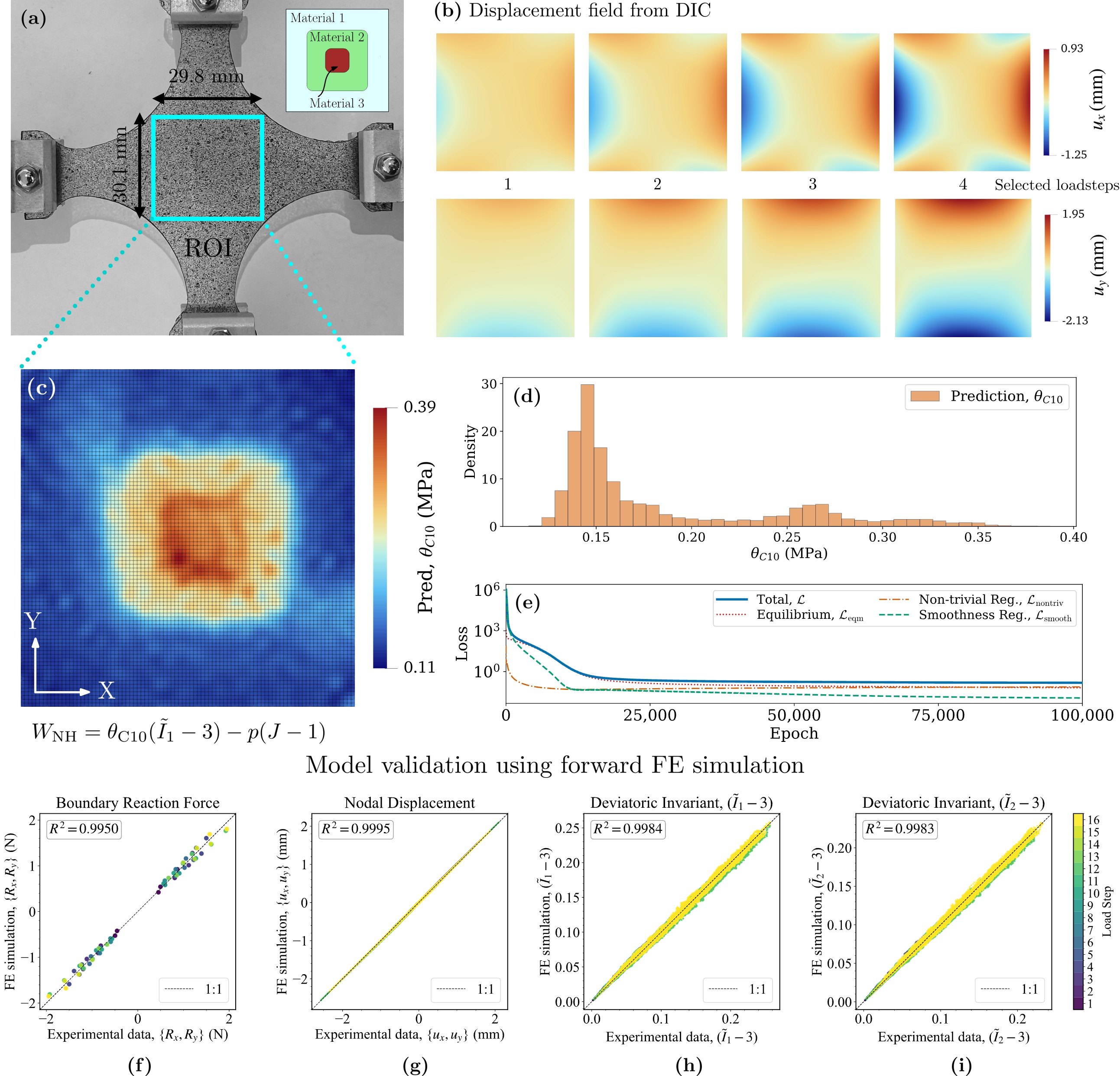}
        \caption{\textbf{Experimental inverse characterization of a three-region concentric-square material heterogeneity.} \textbf{a}, Experimental specimen and region of interest (ROI) used for the inverse analysis, with the inset indicating the nominal three-material concentric-square distribution. \textbf{b}, In-plane displacement fields obtained from DIC for four representative loading steps (for display). A total of sixteen loading steps were used for inverse identification and forward FE validation, with the complete dataset accessible as described in the Data availability statement. \textbf{c}, Predicted spatial distribution of the Neo-Hookean material parameter \(\boldsymbol{\theta}_{\mathrm{C10}}\). \textbf{d}, Histogram of the identified \(\theta_{\mathrm{C10}}\) values at each element in the analyzed domain. \textbf{e}, Convergence histories showing the total, equilibrium, regularization, and smoothness loss terms as functions of epoch on a logarithmic scale. \textbf{f--i}, Forward finite-element (FE) validation comparing the experimental and predicted responses in terms of boundary reaction force, nodal displacement, and the deviatoric invariants \((\tilde I_1-3)\) and \((\tilde I_2-3)\), respectively. Colors in the validation plots distinguish all the sixteen loading steps. The dashed diagonal lines denote the \(1{:}1\) reference, and the reported \(R^2\) values quantify agreement between the corresponding experimental and FE-predicted quantities.}
        \label{fig_2D_concentric_square}
    \end{figure}

\begin{enumerate}[label=\textbf{(\alph*)}]
    \item \textbf{Homogeneous specimen:} We first consider a homogeneous specimen made of PDMS containing a central circular hole of diameter \(\phi = 7.20\)~mm, with an ROI of 21.6~mm \(\times\) 23.8~mm, as shown in Fig.~\ref{fig_2D_homogeneous_hole}a. The specimen (further details discussed in section \ref{methods_exp}) is subjected to non-equibiaxial tensile loading, and the corresponding DIC-measured surface displacement fields, obtained using the VIC-2D commercial software \citep{vic2dmanual7} are shown in Fig.~\ref{fig_2D_homogeneous_hole}b. The selected domain is discretized using 8,559 C3D8 elements, and the optimization requires 10.27 minutes for 80,000 epochs with convergence shown in Fig.~\ref{fig_2D_homogeneous_hole}e. As expected for a homogeneous specimen, the recovered \(\boldsymbol{\theta}_{\mathrm{C10}}\) field remains concentrated within a relatively narrow range over most of the ROI, as seen from the spatial reconstruction in Fig.~\ref{fig_2D_homogeneous_hole}c and the corresponding histogram in Fig.~\ref{fig_2D_homogeneous_hole}d. A local deterioration in the recovered parameter is observed near the traction-free hole boundary due to DIC uncertainty close to the hole edge. The forward FE validation in Fig.~\ref{fig_2D_homogeneous_hole}f--i shows good agreement with the experimental response, with the reaction-force and nodal-displacement comparisons giving \(R^2\) values of 0.9698 and 0.9906, respectively. The corresponding comparisons of \((\tilde{I}_{1}-3)\) and \((\tilde{I}_{2}-3)\) give \(R^2\) values of 0.8819 and 0.8981.
    \begin{figure}[H]
        \centering
        \includegraphics[width=\linewidth]{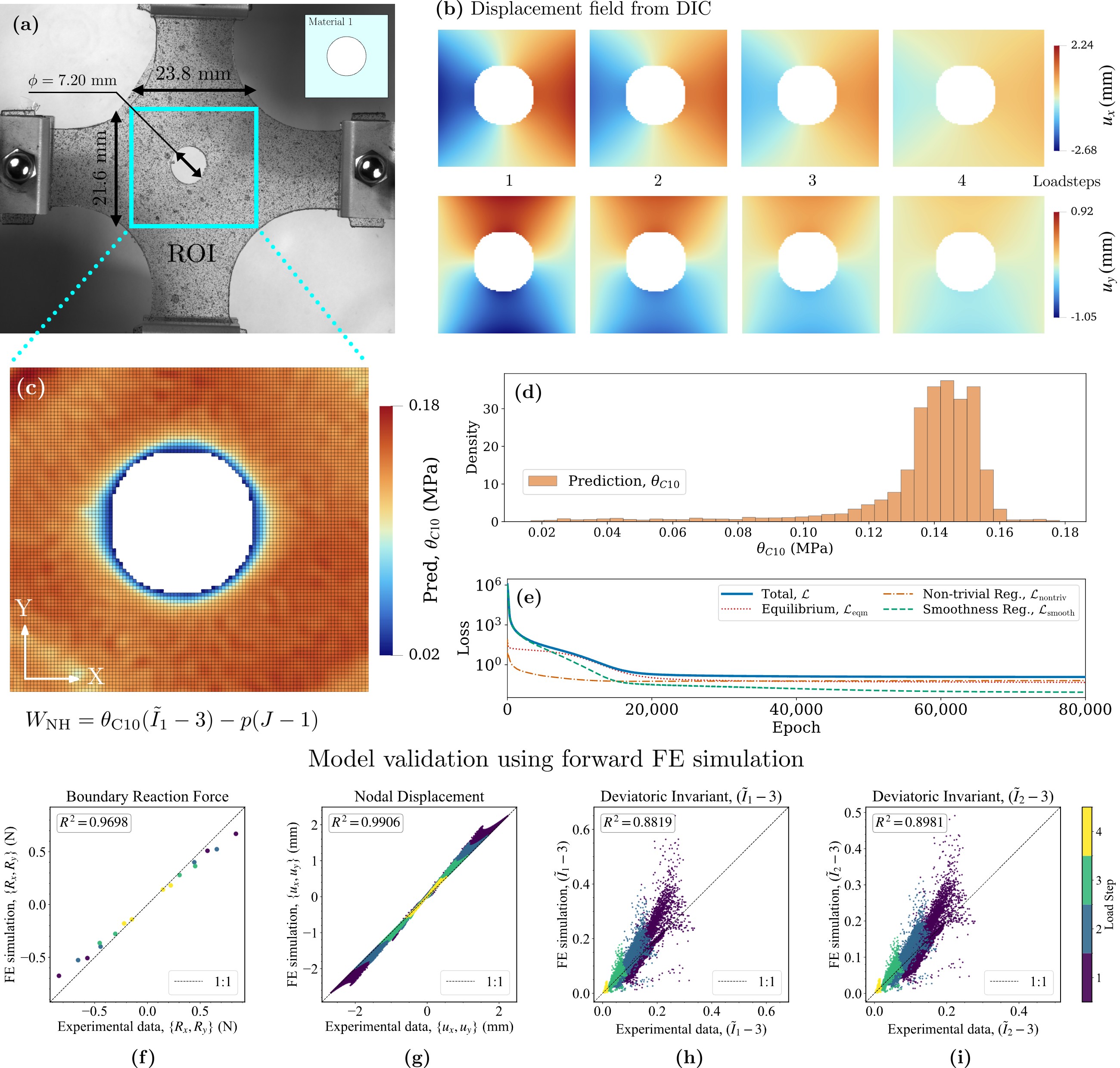}
        \caption{\textbf{Inverse characterization of the experimental homogeneous specimen.} \textbf{a}, Experimental specimen and region of interest (ROI) used for the inverse analysis, with the inset indicating the nominal homogeneous material distribution. \textbf{b}, In-plane displacement fields obtained from digital image correlation (DIC) for the four loading steps. \textbf{c}, Predicted spatial distribution of the Neo-Hookean material parameter \(\boldsymbol{\theta}_{\mathrm{C10}}\). \textbf{d}, Histogram of the identified \(\theta_{\mathrm{C10}}\) values at each element in the analyzed domain. \textbf{e}, Convergence histories showing the total, equilibrium, regularization, and smoothness loss terms as functions of epoch on a logarithmic scale. \textbf{f--i}, Forward finite-element (FE) validation comparing the experimental and predicted responses in terms of boundary reaction force, nodal displacement, and the deviatoric invariants \((\tilde I_1-3)\) and \((\tilde I_2-3)\), respectively. Colors in the validation plots distinguish the all four loading steps. The dashed diagonal lines denote the \(1{:}1\) reference, and the reported \(R^2\) values quantify agreement between the corresponding experimental and FE-predicted quantities.}
        \label{fig_2D_homogeneous_hole}
    \end{figure}    
    \newpage
    \item \textbf{Circular-inclusion specimen}: We next introduce a material contrast by embedding a stiffer circular inclusion within a  compliant matrix, both made of PDMS with different base-to-curing-agent ratios (refer \cref{methods_exp}). The selected ROI is 22.7~mm \(\times\) 22.7~mm, as shown in Fig.~\ref{fig_2D_circular_inclusion}a, and the measured \(u_x\) and \(u_y\) displacement fields obtained from VIC-2D are presented in Fig.~\ref{fig_2D_circular_inclusion}b. Using the incompressible Neo-Hookean (NH) model given by Eq.~\eqref{eq_NH}, the inverse problem is solved over a discretization of 4,900 C3D8 elements in 4.58 minutes for 60,000 epochs with convergence history depicted in Fig.~\ref{fig_2D_circular_inclusion}e. In contrast to the homogeneous case, the recovered \(\boldsymbol{\theta}_{\mathrm{C10}}\) field clearly separates the inclusion from the surrounding matrix, as shown in Fig.~\ref{fig_2D_circular_inclusion}c. This distinction is also reflected in the histogram in Fig.~\ref{fig_2D_circular_inclusion}d, where two dominant material ranges become visible. The recovered \(\boldsymbol{\theta}_{\mathrm{C10}}\) field is subsequently transferred to the forward FE model for validation as showcased in Fig.~\ref{fig_2D_circular_inclusion}f-i. The reaction force and nodal displacement are reproduced with \(R^2\) values of 0.9836 and 0.9866, while the corresponding values for \((\tilde{I}_{1}-3)\) and \((\tilde{I}_{2}-3)\) are 0.9742 and 0.9725, respectively, indicating close agreement between the reconstructed model response and the experimental measurements.
    
    \item \textbf{Concentric-square inclusion specimen:} Finally, we consider a more complex heterogeneous specimen containing three material regions arranged in a concentric-square configuration, with stiffness increasing from the surrounding matrix towards the central region. As before, this stiffness variation is achieved by varying the base-to-curing-agent ratios. The ROI measures 30.1~mm \(\times\) 29.8~mm and is shown in Fig.~\ref{fig_2D_concentric_square}a, with representative DIC displacement fields presented in Fig.~\ref{fig_2D_concentric_square}b. The selected domain is discretized with 8,010 C3D8 elements, and Fig.~\ref{fig_2D_concentric_square}e shows the convergence history where the optimization is performed for 100,000 epochs requiring 12.28 minutes. Fig.~\ref{fig_2D_concentric_square}c shows the performance of the proposed framework on the current specimen, which has increased material and geometric complexity. It can be observed that the recovered \(\boldsymbol{\theta}_{\mathrm{C10}}\) parameter field resolves into three material regions and preserves the expected increasing trend of stiffness towards the core. The corresponding histogram in Fig.~\ref{fig_2D_concentric_square}d further shows distinct ranges associated with the different material regions. The forward FE simulation results shown in Fig.~\ref{fig_2D_concentric_square}f-i provides the closest agreement among the three experimental cases, with \(R^2\) values of 0.9950, 0.9995, 0.9984, and 0.9983 for the reaction force, nodal displacement, \((\tilde{I}_{1}-3)\), and \((\tilde{I}_{2}-3)\), respectively. These results show that the proposed framework can recover spatial material contrast not only for a single inclusion but also for more complex multi-material distributions from experimentally measured surface displacement fields.
\end{enumerate}
In addition to comparing the FE simulation predictions against the experimental observations, we performed uniaxial tensile tests to calibrate the incompressible NH model for each specific base-to-curing-agent PDMS used above. The $C_{\mathrm{10}}$ parameters obtained from the uniaxial data-fit are compared to the $\boldsymbol{\theta}_{\mathrm{C10}}$ predictions from our method in supplementary section S3.5.
\begin{table}[t]
    \centering
    \renewcommand{\arraystretch}{1.20}
    \begin{tabular}{lcccc}
        \toprule
        \textbf{Experimental Specimen} &
        \(\boldsymbol{R^2_{\mathrm{F}}}\) &
        \(\boldsymbol{R^2_{\mathrm{u}}}\) &
        \(\boldsymbol{R^2_{(\tilde{I}_1-3)}}\) &
        \(\boldsymbol{R^2_{(\tilde{I}_2-3)}}\) \\
        \midrule
        Homogeneous with hole       & 0.9698 & 0.9906 & 0.8819 & 0.8981 \\
        Circular inclusion          & 0.9836 & 0.9866 & 0.9742 & 0.9725 \\
        Concentric-square inclusion & 0.9950 & 0.9995 & 0.9984 & 0.9983 \\
        \bottomrule
    \end{tabular}
    \caption{Forward FE simulation validation metrics for the experimental specimens. Here, \(R^2_{\mathrm{F}}\) denotes the coefficient of determination between the experimentally measured and forward FE-predicted boundary reaction forces, \(R^2_{\mathrm{u}}\) corresponds to the comparison of nodal displacement fields, and \(R^2_{(\tilde{I}_1-3)}\) and \(R^2_{(\tilde{I}_2-3)}\) denote the corresponding comparisons of the deviatoric invariants.}
    \label{table_exp_forward_validation}
\end{table}

\begin{table}[H]
    \centering
    \small
    \renewcommand{\arraystretch}{1.15}
    \begin{tabularx}{\linewidth}{@{}Xcccc@{}}
        \toprule
        \textbf{Test case description} &
        \textbf{Constitutive} &
        \textbf{C3D8} &
        \textbf{Total} &
        \textbf{GPU time (minutes)} \\

          &
        \textbf{model type} &
        \textbf{elements} &
        \textbf{epochs} &
        \textit{noiseless ; noisy case} \\
        \midrule

        \multicolumn{4}{@{}l}{\textbf{Three-dimensional numerical studies}} \\[1pt]

        -- Smooth checkerboard (Fig.~\ref{fig_3D_checker_NH})
            & NH2   & 8,000 & 60,000 & 16.81 ; 17.20 \\
        -- Graded spherical inclusion (Fig.~\ref{fig_3D_tanhsph_MR})
            & MR   & 8,000  & 60,000 & 29.57 ; 29.49 \\
        -- Anisotropic graded spherical inclusion (Fig.~\ref{fig_3D_tanhsph_NH_HGO})
            & HGO* & 8,000  & 150,000 & 73.70 ; 72.89  \\
        -- Linear face-diagonal variation (Fig.~S3)
            & NH2   & 8,000 & 60,000 & 16.93 ; 17.02 \\            
        -- Discrete cylindrical inclusion (Fig.~S4)
            & NH2   & 11,445 & 60,000 & 28.76 ; 29.65 \\
        -- Local elastography for graded region (Fig.~\ref{fig_3D_tanhsph_MR_local})
            & MR   & 729  & 60,000  & 8.28 \\
            
        \addlinespace
        \multicolumn{4}{@{}l}{\textbf{In-plane synthetic and experimental studies}} \\[1pt]
        -- Smooth checkerboard (Fig.~\ref{fig_2D_checker}) 
            & NH   & 5,041 & 60,000 & 4.64 \\
        -- Homogeneous specimen with hole (Fig.~\ref{fig_2D_homogeneous_hole})
            & NH   & 8,559 & 80,000 & 10.27 \\
        -- Circular-inclusion specimen (Fig.~\ref{fig_2D_circular_inclusion})
            & NH   & 4,900 & 60,000 & 4.58 \\
        -- Concentric-square inclusion specimen (Fig.~\ref{fig_2D_concentric_square})
            & NH   & 8,010 & 100,000 & 12.28 \\
        \bottomrule
    \end{tabularx}
    \caption{Summary of the GPU optimization time for the numerical and experimental test cases. NH2/NH, MR, and HGO* denote the Neo-Hookean, Mooney-Rivlin, and modified Holzapfel-Gasser-Ogden models, respectively.}
    \label{table_computational_summary}
\end{table}

\section{Conclusion}
\label{sec_conclusion}
In this work, we presented an inverse identification framework for predicting constitutive parameters of heterogeneous/graded hyperelastic materials from full-field displacement measurements. A key contribution of our proposed method is assigning independent constitutive parameters to individual finite elements in the geometry and allowing their values to be identified through an optimization framework that combines equilibrium enforcement with positivity constraints, regularization, and smoothing. The proposed method does not require any prior segmentation into distinct material regions and provides an interpretable element-wise prediction of constitutive parameters. A key outcome is the explicit description of the recovered material field that can be readily incorporated into conventional forward finite-element (FE) simulations. Another important aspect of the framework is its ability to recover the relative material property contrast without the of measured boundary reaction forces. By enforcing equilibrium on the (internal) free degrees of freedom, the spatial variation and relative contrast of the constitutive parameters can be identified from displacement measurements alone, while the absolute scale remains undetermined. When boundary reaction forces are available, this absolute scaling multiplier can be computed through an additional force-balance condition. This feature is useful in contexts such as in vivo MRI/X-ray based computed tomography and subsequent DVC, where boundary reaction forces are often inaccessible.\par

We examined the framework using three-dimensional numerical examples covering different constitutive models and spatial patterns of heterogeneity. The compressible Neo-Hookean and Mooney-Rivlin cases demonstrate the simultaneous recovery of spatially varying isochoric and volumetric parameters, while the case study on the modified Holzapfel-Gasser-Ogden model validates the formulation for anisotropic hyperelastic materials with a prescribed fiber direction. The additional examples reported in the supplementary section S3, involving linear property gradients and a sharp interface with a cylindrical inclusion, further show that the same formulation can identify both smooth and discontinuous material distributions with acceptable accuracies. For these studies, we considered a combination of biaxial loading in the \(x\)-\(y\), \(x\)-\(z\), and \(y\)-\(z\) planes, together with uniaxial loading along each principal direction, providing rich deformation-state data for inverse identification. Across the investigated cases, the predicted parameter fields show good agreement with their ground truth distributions, even when the displacement measurements were perturbed by noise.\par

We conducted experimental studies to further assess the proposed framework using specimens with homogeneous and heterogeneous material distributions, with in-plane displacement measurements obtained through 2D-DIC. The experiments included a homogeneous specimen containing a central hole, a specimen with a circular inclusion, and a three-material concentric-square configuration. The identified parameter fields distinguish the different material regions without prescribing their interfaces during optimization. We subsequently evaluated the predicted material response by assigning the recovered element-wise parameters to forward FE models, with the measured boundary displacements imposed as Dirichlet conditions. A comparison study of forward FE results with experimental measurements shows close agreement in the boundary reaction forces, nodal displacements, and the element-wise deformation invariants $\left(\tilde{I}_1-3\right)$ and $\left(\tilde{I}_2-3\right)$ for all experimental studies. We also compare our material property predictions with incompressible NH parameters calibrated from uniaxial stretch tests on individual PDMS samples with identical base-to-curing-agent ratios (supplementary section S3.5).\par

The computational performance of the proposed framework for the numerical and experimental studies is summarized in Table~\ref{table_computational_summary}. All optimizations are performed on an NVIDIA RTX 5000 Ada Generation GPU with 32 GB of dedicated memory. The three-dimensional numerical cases involving approximately 8,000--11,445 elements required about 17--74 minutes, while the in-plane synthetic and experimental cases were completed within approximately 5--12 minutes. The anisotropic HGO* example required a longer optimization time, with 150,000 epochs compared with 60,000 epochs for most other numerical cases. Although these results demonstrate the computational feasibility of element-wise parameter identification on a single GPU, the optimization cost depends on the mesh resolution, constitutive formulation, and number of training epochs. The present work also has certain limitations. The constitutive model (and fiber orientation) is prescribed \textit{a priori}, and the recoverable spatial detail depends on the discretization and the quality of the measured displacement field. Furthermore, the constitutive parameters are identified at each element in a given discretization of the geometry, thereby restricting forward FE simulations only to the discretization on which the inverse problem (parameter identification) was solved. Though not attempted by us in this work, this shortcoming can be addressed by applying piecewise cubic or Radial Basis Function spline interpolation to the predicted constitutive parameter field to allow for forward FE simulations with other discretizations of the same geometry. In this work, we limit experimental validation to 2D contexts using surface DIC data. In future work, we will focus on extending the experimental validation to volumetric displacement measurements using DVC. Another promising direction would be to extend our framework to characterize polycrystalline solids at the micrometer length scale. At this length scale, polycrystals are both-- heterogeneous, anisotropic linear-elastic solids (under small deformations) and heterogeneous elastoplastic materials (under larger deformations). Our existing approach would need to be modified to characterize heterogeneous elastoplastic materials. This is because the inverse problem for elastoplastic solids is non-linear \citep{Flaschel2022} and therefore the parameter-field solution obtained using internal equilibrium would not be a linear multiple of the actual parameter field.\par

\section*{Code and data availability}

    The codes and data generated during the current study are available from the corresponding author upon reasonable request.
    
\section*{Acknowledgments}
    A.J. would like to acknowledge funding by Anusandhan National Research Foundation (ANRF) through Grant No. ANRF/ECRG/2024/005303/ENS. D.D. and A.J. wish to acknowledge Indian Institute of Science's startup and capital equipment (2024) grant.
    
\section*{Author contributions}
    A.J., D.D. and K.L.C. conceptualized the study. K.L.C., V.H., P.P. and A.J. developed and implemented the computational algorithms. K.L.C., P.P. and G.N.S. conducted the experiments and processed the experimental data.  K.L.C. performed the numerical simulations and extracted the computational results, while K.L.C. and V.H. analyzed the results and prepared the figures and illustrations. D.D. and A.J. supervised the study, and acquired funding. V.H. and P.P. contributed equally to this work. All authors participated in the writing of the manuscript.
\section*{Competing interests}
    The authors declare no competing interests.

% \appendix

\bibliographystyle{elsarticle-num-names} 
\bibliography{references}

@article{Koorman2022,
  title = {Spatial collagen stiffening promotes collective breast cancer cell invasion by reinforcing extracellular matrix alignment},
  volume = {41},
  issn = {1476-5594},
  doi = {10.1038/s41388-022-02258-1},
  number = {17},
  journal = {Oncogene},
  publisher = {Springer Science and Business Media LLC},
  author = {Koorman, Thijs and Jansen, Karin A. and Khalil, Antoine and Haughton, Peter D. and Visser, Daan and Rätze, Max A. K. and Haakma, Wisse E. and Sakalauskaitè, Gabrielè and van Diest, Paul J. and de Rooij, Johan and Derksen, Patrick W. B.},
  year = {2022},
  month = Mar,
  pages = {2458–2469},
}

@article{Micalet2021,
  title = {3D In Vitro Models for Investigating the Role of Stiffness in Cancer Invasion},
  volume = {9},
  issn = {2373-9878},
  doi = {10.1021/acsbiomaterials.0c01530},
  number = {7},
  journal = {ACS Biomaterials Science \& Engineering},
  publisher = {American Chemical Society (ACS)},
  author = {Micalet, Auxtine and Moeendarbary, Emad and Cheema, Umber},
  year = {2021},
  month = Jun,
  pages = {3729–3741},
}

@article{Skerl2016,
  title = {Anisotropy of Solid Breast Lesions in 2D Shear Wave Elastography is an Indicator of Malignancy},
  volume = {23},
  issn = {1076-6332},
  doi = {10.1016/j.acra.2015.09.016},
  number = {1},
  journal = {Academic Radiology},
  publisher = {Elsevier BV},
  author = {Skerl, Katrin and Vinnicombe, Sarah and Thomson, Kim and McLean, Denis and Giannotti, Elisabetta and Evans, Andrew},
  year = {2016},
  month = Jan,
  pages = {53–61},
}

@article{Nia2020,
  title = {Physical traits of cancer},
  volume = {370},
  issn = {1095-9203},
  doi = {10.1126/science.aaz0868},
  number = {6516},
  journal = {Science},
  publisher = {American Association for the Advancement of Science (AAAS)},
  author = {Nia, Hadi T. and Munn, Lance L. and Jain, Rakesh K.},
  year = {2020},
  month = Oct,
}

@article{SydneyGladman2016,
  title = {Biomimetic 4D printing},
  volume = {15},
  issn = {1476-4660},
  doi = {10.1038/nmat4544},
  number = {4},
  journal = {Nature Materials},
  publisher = {Springer Science and Business Media LLC},
  author = {Sydney Gladman, A. and Matsumoto, Elisabetta A. and Nuzzo, Ralph G. and Mahadevan, L. and Lewis, Jennifer A.},
  year = {2016},
  month = Jan,
  pages = {413–418},
}

@article{Connolly2015,
  title = {Mechanical Programming of Soft Actuators by Varying Fiber Angle},
  volume = {2},
  issn = {2169-5180},
  doi = {10.1089/soro.2015.0001},
  number = {1},
  journal = {Soft Robotics},
  publisher = {SAGE Publications},
  author = {Connolly, Fionnuala and Polygerinos, Panagiotis and Walsh, Conor J. and Bertoldi, Katia},
  year = {2015},
  month = Mar,
  pages = {26–32},
}

@article{Brooks2019,
  title = {High-fidelity aerostructural optimization of tow-steered composite wings},
  volume = {88},
  issn = {0889-9746},
  doi = {10.1016/j.jfluidstructs.2019.04.005},
  journal = {Journal of Fluids and Structures},
  publisher = {Elsevier BV},
  author = {Brooks, Timothy R. and Martins, Joaquim R.R.A. and Kennedy, Graeme J.},
  year = {2019},
  month = Jul,
  pages = {122–147},
}

@article{Holzapfel2019,
  title = {On fibre dispersion modelling of soft biological tissues: a review},
  volume = {475},
  issn = {1471-2946},
  doi = {10.1098/rspa.2018.0736},
  number = {2224},
  journal = {Proceedings of the Royal Society A: Mathematical, Physical and Engineering Sciences},
  publisher = {The Royal Society},
  author = {Holzapfel, Gerhard A. and Ogden, Ray W. and Sherifova, Selda},
  year = {2019},
  month = Apr,
  pages = {20180736},
}

@article{Bayar2026,
  title = {Additive Manufacturing of Functionally Graded Lattice Structures: Process–Structure–Property Relationships, Design Strategies, and Future Perspectives},
  volume = {10},
  issn = {2504-4494},
  doi = {10.3390/jmmp10080267},
  number = {8},
  journal = {Journal of Manufacturing and Materials Processing},
  publisher = {MDPI AG},
  author = {Bayar, Abdulcelil and Altun, Fatih and Sarac, Ahmet Enes and Ateiwi, Saleh and Alasahin, Seyda Naz and Asmatulu, Eylem},
  year = {2026},
  month = Jul,
  pages = {267},
}

@article{Li2020,
  title = {A Review on Functionally Graded Materials and Structures via Additive Manufacturing: From Multi‐Scale Design to Versatile Functional Properties},
  volume = {5},
  issn = {2365-709X},
  doi = {10.1002/admt.201900981},
  number = {6},
  journal = {Advanced Materials Technologies},
  publisher = {Wiley},
  author = {Li, Yan and Feng, Zuying and Hao, Liang and Huang, Lijing and Xin, Chenxing and Wang, Yushen and Bilotti, Emiliano and Essa, Khamis and Zhang, Han and Li, Zheng and Yan, Feifei and Peijs, Ton},
  year = {2020},
  month = Apr,
}

@article{Birman2007,
  title = {Modeling and Analysis of Functionally Graded Materials and Structures},
  volume = {60},
  issn = {2379-0407},
  doi = {10.1115/1.2777164},
  number = {5},
  journal = {Applied Mechanics Reviews},
  publisher = {ASME International},
  author = {Birman, Victor and Byrd, Larry W.},
  year = {2007},
  month = Sep,
  pages = {195–216},
}

@article{Genovese2021,
  title = {Multi-view Digital Image Correlation Systems for In Vitro Testing of Arteries from Mice to Humans},
  volume = {61},
  issn = {1741-2765},
  doi = {10.1007/s11340-021-00746-1},
  number = {9},
  journal = {Experimental Mechanics},
  publisher = {Springer Science and Business Media LLC},
  author = {Genovese, K. and Badel, P. and Cavinato, C. and Pierrat, B. and Bersi, M. R. and Avril, S. and Humphrey, J. D.},
  year = {2021},
  month = Jul,
  pages = {1455–1472},
}

@article{apunaman2026,
  title = {A robotic framework for high-throughput and multi-view 3D digital image correlation (3D-DIC): Increasing measurement volume and versatility for deformation analysis},
  volume = {99},
  issn = {0736-5845},
  doi = {10.1016/j.rcim.2025.103187},
  journal = {Robotics and Computer-Integrated Manufacturing},
  publisher = {Elsevier BV},
  author={{\c{C}}apunaman, {\"O}zg{\"u}{\c{c}} Bertu{\u{g}} and Mohseni, Alale and Dombrovskij, Dennis and Yin, Kaiyang and G{\"u}rsoy, Benay and Mylo, Max David},
  year = {2026},
  month = Jun,
  pages = {103187},
}

@article{Chen2022a,
  title = {Mirror-assisted multi-view digital image correlation: Principles, applications and implementations},
  volume = {149},
  issn = {0143-8166},
  doi = {10.1016/j.optlaseng.2021.106786},
  journal = {Optics and Lasers in Engineering},
  publisher = {Elsevier BV},
  author = {Chen, Bin and Pan, Bing},
  year = {2022},
  month = Feb,
  pages = {106786},
}

@article{Wang2024,
  title = {3D observations provide striking findings in rubber elasticity},
  volume = {121},
  issn = {1091-6490},
  doi = {10.1073/pnas.2404205121},
  number = {24},
  journal = {Proceedings of the National Academy of Sciences},
  publisher = {National Academy of Sciences},
  author = {Wang, Zifan and Das, Shuvrangsu and Joshi, Akshay and Shaikeea, Angkur J. D. and Deshpande, Vikram S.},
  year = {2024},
  month = Jun,
}

@article{Estrada2020,
  title = {MR-u: Material Characterization Using 3D Displacement-Encoded Magnetic Resonance and the Virtual Fields Method},
  volume = {60},
  issn = {1741-2765},
  doi = {10.1007/s11340-020-00595-4},
  number = {7},
  journal = {Experimental Mechanics},
  publisher = {Springer Science and Business Media LLC},
  author = {Estrada, J.B. and Luetkemeyer, C.M. and Scheven, U.M. and Arruda, E.M.},
  year = {2020},
  month = Jun,
  pages = {907–924},
}

@article{Kim2026,
  title = {Morphology-, noise-, and resolution-robust ultrasound elasticity imaging with Fourier neural operator},
  volume = {283},
  issn = {0169-2607},
  doi = {10.1016/j.cmpb.2026.109432},
  journal = {Computer Methods and Programs in Biomedicine},
  publisher = {Elsevier BV},
  author = {Kim, Heekyu and Lee, Hugon and Park, Minwoo and Ryu, Seunghwa},
  year = {2026},
  month = Aug,
  pages = {109432},
}

@book{Marwala2010,
  title = {Finite-element-model Updating Using Computional Intelligence Techniques},
  isbn = {9781849963237},
  doi = {10.1007/978-1-84996-323-7},
  publisher = {Springer London},
  author = {Marwala, Tshilidzi},
  year = {2010},
}

@article{Ikehata1990,
  title = {Inversion Formulas for the Linearized Problem for an Inverse Boundary Value Problem in Elastic Prospection},
  volume = {50},
  issn = {1095-712X},
  doi = {10.1137/0150097},
  number = {6},
  journal = {SIAM Journal on Applied Mathematics},
  publisher = {Society for Industrial & Applied Mathematics (SIAM)},
  author = {Ikehata, Masaru},
  year = {1990},
  month = Dec,
  pages = {1635–1644},
}

@article{Avril2008,
  title = {Overview of Identification Methods of Mechanical Parameters Based on Full-field Measurements},
  volume = {48},
  issn = {1741-2765},
  doi = {10.1007/s11340-008-9148-y},
  number = {4},
  journal = {Experimental Mechanics},
  publisher = {Springer Science and Business Media LLC},
  author = {Avril, Stéphane and Bonnet, Marc and Bretelle, Anne-Sophie and Grédiac, Michel and Hild, François and Ienny, Patrick and Latourte, Félix and Lemosse, Didier and Pagano, Stéphane and Pagnacco, Emmanuel and Pierron, Fabrice},
  year = {2008},
  month = Jul,
  pages = {381–402},
}

@article{Kavanagh1971,
  title = {Finite element applications in the characterization of elastic solids},
  volume = {7},
  issn = {0020-7683},
  doi = {10.1016/0020-7683(71)90015-1},
  number = {1},
  journal = {International Journal of Solids and Structures},
  publisher = {Elsevier BV},
  author = {Kavanagh, Kenneth T. and Clough, Ray W.},
  year = {1971},
  month = Jan,
  pages = {11–23},
}

@article{Pottier2011,
  title = {Contribution of heterogeneous strain field measurements and boundary conditions modelling in inverse identification of material parameters},
  volume = {30},
  issn = {0997-7538},
  doi = {10.1016/j.euromechsol.2010.10.001},
  number = {3},
  journal = {European Journal of Mechanics - A/Solids},
  publisher = {Elsevier BV},
  author = {Pottier, T. and Toussaint, F. and Vacher, P.},
  year = {2011},
  month = May,
  pages = {373–382},
}

@article{Toussaint2006,
  title = {The virtual fields method with piecewise virtual fields},
  volume = {48},
  issn = {0020-7403},
  doi = {10.1016/j.ijmecsci.2005.10.002},
  number = {3},
  journal = {International Journal of Mechanical Sciences},
  publisher = {Elsevier BV},
  author = {Toussaint, Evelyne and Grédiac, Michel and Pierron, Fabrice},
  year = {2006},
  month = Mar,
  pages = {256–264},
}

@article{Grdiac2006,
  title = {The Virtual Fields Method for Extracting Constitutive Parameters From Full-Field Measurements: a Review},
  volume = {42},
  issn = {1475-1305},
  doi = {10.1111/j.1475-1305.2006.00283.x},
  number = {4},
  journal = {Strain},
  publisher = {Wiley},
  author = {Grédiac, M. and Pierron, F. and Avril, S. and Toussaint, E.},
  year = {2006},
  month = Nov,
  pages = {233–253},
}

@article{Florentin2010,
  title = {Identification of the parameters of an elastic material model using the constitutive equation gap method},
  volume = {46},
  issn = {1432-0924},
  doi = {10.1007/s00466-010-0496-y},
  number = {4},
  journal = {Computational Mechanics},
  publisher = {Springer Science and Business Media LLC},
  author = {Florentin, Eric and Lubineau, Gilles},
  year = {2010},
  month = Apr,
  pages = {521–531},
}

@article{Claire2004,
  title = {A finite element formulation to identify damage fields: the equilibrium gap method},
  volume = {61},
  issn = {1097-0207},
  doi = {10.1002/nme.1057},
  number = {2},
  journal = {International Journal for Numerical Methods in Engineering},
  publisher = {Wiley},
  author = {Claire, D. and Hild, F. and Roux, S.},
  year = {2004},
  month = Jul,
  pages = {189–208},
}

@article{Claire2007,
  title = {Identification of a Damage Law by Using Full-field Displacement Measurements},
  volume = {16},
  issn = {1530-7921},
  doi = {10.1177/1056789506064940},
  number = {2},
  journal = {International Journal of Damage Mechanics},
  publisher = {SAGE Publications},
  author = {Claire, Damien and Hild, François and Roux, Stéphane},
  year = {2007},
  month = Apr,
  pages = {179–197},
}

@article{Geymonat2002,
  title = {Identification of elastic parameters by displacement field measurement},
  volume = {330},
  issn = {1873-7234},
  doi = {10.1016/s1631-0721(02)01476-6},
  number = {6},
  journal = {Comptes Rendus. Mécanique},
  publisher = {MathDoc/Centre Mersenne},
  author = {Geymonat, Giuseppe and Hild, François and Pagano, Stéphane},
  year = {2002},
  month = Jan,
  pages = {403–408},
}

@article{Geymonat2003,
  title = {Identification of Mechanical Properties by Displacement Field Measurement: A Variational Approach},
  volume = {38},
  issn = {1572-9648},
  doi = {10.1023/a:1024766911435},
  number = {5},
  journal = {Meccanica},
  publisher = {Springer Science and Business Media LLC},
  author = {Geymonat, Giuseppe and Pagano, Stéphane},
  year = {2003},
  month = Oct,
  pages = {535–545},
}

@article{Avril2008a,
  title = {3D Heterogeneous Stiffness Reconstruction Using MRI and the Virtual Fields Method},
  volume = {48},
  issn = {1741-2765},
  doi = {10.1007/s11340-008-9128-2},
  number = {4},
  journal = {Experimental Mechanics},
  publisher = {Springer Science and Business Media LLC},
  author = {Avril, S. and Huntley, J. M. and Pierron, F. and Steele, D. D.},
  year = {2008},
  month = Mar,
  pages = {479–494},
}

@article{Chen2021,
  title = {Learning hidden elasticity with deep neural networks},
  volume = {118},
  issn = {1091-6490},
  doi = {10.1073/pnas.2102721118},
  number = {31},
  journal = {Proceedings of the National Academy of Sciences},
  publisher = {National Academy of Sciences},
  author = {Chen, Chun-Teh and Gu, Grace X.},
  year = {2021},
  month = Jul,
}

@article{Chen2023,
  title = {Physics‐Informed Deep‐Learning For Elasticity: Forward, Inverse, and Mixed Problems},
  volume = {10},
  issn = {2198-3844},
  doi = {10.1002/advs.202300439},
  number = {18},
  journal = {Advanced Science},
  publisher = {Wiley},
  author = {Chen, Chun‐Teh and Gu, Grace X.},
  year = {2023},
  month = Apr,
}

@article{Kamali2023,
  title = {Elasticity imaging using physics-informed neural networks: Spatial discovery of elastic modulus and Poisson’s ratio},
  volume = {155},
  issn = {1742-7061},
  doi = {10.1016/j.actbio.2022.11.024},
  journal = {Acta Biomaterialia},
  publisher = {Elsevier BV},
  author = {Kamali, Ali and Sarabian, Mohammad and Laksari, Kaveh},
  year = {2023},
  month = Jan,
  pages = {400–409},
}

@article{Kamali2024,
  title = {Physics-informed UNets for discovering hidden elasticity in heterogeneous materials},
  volume = {150},
  issn = {1751-6161},
  doi = {10.1016/j.jmbbm.2023.106228},
  journal = {Journal of the Mechanical Behavior of Biomedical Materials},
  publisher = {Elsevier BV},
  author = {Kamali, Ali and Laksari, Kaveh},
  year = {2024},
  month = Feb,
  pages = {106228},
}

@article{Srikitrungruang2025,
  title = {Robust Physics‐Informed Neural Network Approach for Estimating Heterogeneous Elastic Properties from Noisy Displacement Data},
  volume = {12},
  issn = {2198-3844},
  doi = {10.1002/advs.202508445},
  number = {48},
  journal = {Advanced Science},
  publisher = {Wiley},
  author = {Srikitrungruang, Tatthapong and Aghaee Dabaghan Fard, Sina and Lemon, Matthew and Lee, Jaesung and Zhou, Yuxiao},
  year = {2025},
  month = Nov,
}

@misc{Srikitrungruang2026,
  doi = {10.48550/ARXIV.2607.14563},
  author = {Srikitrungruang, Tatthapong and Lee, Jaesung},
  title = {Probabilistic Physics-Informed Neural Networks for Estimating Heterogeneous Elastic Properties from Low-Resolution and Noisy Displacement Data},
  publisher = {arXiv},
  year = {2026},
  copyright = {Creative Commons Attribution Non Commercial No Derivatives 4.0 International},
}

@article{Zhu2026,
  title = {Robust Inverse Identification of Heterogeneous Hyperelastic Materials Under Large Deformation via Residual‐Adaptive Physics‐Informed Neural Networks},
  volume = {127},
  issn = {1097-0207},
  doi = {10.1002/nme.70427},
  number = {17},
  journal = {International Journal for Numerical Methods in Engineering},
  publisher = {Wiley},
  author = {Zhu, Jing‐Ang and Lu, Wenjing and Li, Han and Liu, Zishun},
  year = {2026},
  month = Sep,
}

@article{Shi2025,
  title = {Deep learning without stress data on the discovery of multi-regional hyperelastic properties},
  volume = {76},
  issn = {1432-0924},
  doi = {10.1007/s00466-024-02591-0},
  number = {1},
  journal = {Computational Mechanics},
  publisher = {Springer Science and Business Media LLC},
  author = {Shi, Ruike and Yang, Haitian and Chen, Jianxu and Hackl, Klaus and Avril, Stéphane and He, Yiqian},
  year = {2025},
  month = Jan,
  pages = {117–146},
}

@article{Thakolkaran2022,
  title = {NN-EUCLID: Deep-learning hyperelasticity without stress data},
  volume = {169},
  issn = {0022-5096},
  doi = {10.1016/j.jmps.2022.105076},
  journal = {Journal of the Mechanics and Physics of Solids},
  publisher = {Elsevier BV},
  author = {Thakolkaran, Prakash and Joshi, Akshay and Zheng, Yiwen and Flaschel, Moritz and De Lorenzis, Laura and Kumar, Siddhant},
  year = {2022},
  month = Dec,
  pages = {105076},
}

@article{Chaurasiya2026,
  title = {Hetero-EUCLID: Interpretable model discovery for heterogeneous hyperelastic materials using stress-unsupervised learning},
  volume = {452},
  issn = {0045-7825},
  doi = {10.1016/j.cma.2026.118729},
  journal = {Computer Methods in Applied Mechanics and Engineering},
  publisher = {Elsevier BV},
  author = {Chaurasiya, Kanhaiya Lal and Dutta, Saurav and Kumar, Siddhant and Joshi, Akshay},
  year = {2026},
  month = Apr,
  pages = {118729},
}

@article{Joshi2022,
  title = {Bayesian-EUCLID: Discovering hyperelastic material laws with uncertainties},
  volume = {398},
  issn = {0045-7825},
  doi = {10.1016/j.cma.2022.115225},
  journal = {Computer Methods in Applied Mechanics and Engineering},
  publisher = {Elsevier BV},
  author = {Joshi, Akshay and Thakolkaran, Prakash and Zheng, Yiwen and Escande, Maxime and Flaschel, Moritz and De Lorenzis, Laura and Kumar, Siddhant},
  year = {2022},
  month = Aug,
  pages = {115225},
}

@article{Ta2026,
  title = {Fully data-driven inverse characterization of heterogeneous materials with hyper-network neural ODEs},
  volume = {12},
  issn = {2057-3960},
  doi = {10.1038/s41524-026-02027-8},
  number = {1},
  journal = {npj Computational Materials},
  publisher = {Springer Science and Business Media LLC},
  author = {Taç, Vahidullah and Amiri-Hezaveh, Amirhossein and Bechtel, Grace N. and Loftin, Titus and Rausch, Manuel K. and Sahli Costabal, Francisco and Tepole, Adrian Buganza},
  year = {2026},
  month = Mar,
}

@article{Abbasi2026,
  title = {Discovery of Hyperelastic Constitutive Laws from Experimental Data with EUCLID},
  volume = {66},
  issn = {1741-2765},
  doi = {10.1007/s11340-026-01290-6},
  number = {5},
  journal = {Experimental Mechanics},
  publisher = {Springer Science and Business Media LLC},
  author = {Abbasi, A. and Ricci, M. and Carrara, P. and Flaschel, M. and Kumar, S. and Marfia, S. and De Lorenzis, L.},
  year = {2026},
  month = Apr,
  pages = {877–908},
}

@article{Tac2022,
  title = {Data-driven tissue mechanics with polyconvex neural ordinary differential equations},
  volume = {398},
  issn = {0045-7825},
  doi = {10.1016/j.cma.2022.115248},
  journal = {Computer Methods in Applied Mechanics and Engineering},
  publisher = {Elsevier BV},
  author = {Tac, Vahidullah and Sahli Costabal, Francisco and Tepole, Adrian B.},
  year = {2022},
  month = Aug,
  pages = {115248},
}

@article{Bay1999,
  title = {Digital volume correlation: Three-dimensional strain mapping using X-ray tomography},
  volume = {39},
  issn = {1741-2765},
  doi = {10.1007/bf02323555},
  number = {3},
  journal = {Experimental Mechanics},
  publisher = {Springer Science and Business Media LLC},
  author = {Bay, B. K. and Smith, T. S. and Fyhrie, D. P. and Saad, M.},
  year = {1999},
  month = Sep,
  pages = {217–226},
}

@article{Grega2025,
  title = {High-speed X-ray tomography for 4D imaging},
  volume = {122},
  issn = {1091-6490},
  doi = {10.1073/pnas.2521089122},
  number = {51},
  journal = {Proceedings of the National Academy of Sciences},
  publisher = {National Academy of Sciences},
  author = {Grega, Ivan and Whitney, William and Deshpande, Vikram Sudhir},
  year = {2025},
  month = Dec,
}

@misc{Li2020a,
  doi = {10.48550/ARXIV.2010.08895},
  author = {Li, Zongyi and Kovachki, Nikola and Azizzadenesheli, Kamyar and Liu, Burigede and Bhattacharya, Kaushik and Stuart, Andrew and Anandkumar, Anima},
  title = {Fourier Neural Operator for Parametric Partial Differential Equations},
  publisher = {arXiv},
  year = {2020},
  copyright = {arXiv.org perpetual, non-exclusive license},
}

@inproceedings{Mallampati2021,
  title = {Measuring Tissue Elastic Properties Using Physics Based Neural Networks},
  doi = {10.1109/laus53676.2021.9639231},
  booktitle = {2021 IEEE UFFC Latin America Ultrasonics Symposium (LAUS)},
  publisher = {IEEE},
  author = {Mallampati, Aishwarya and Almekkawy, Mohamed},
  year = {2021},
  month = Oct,
  pages = {1–4},
}

@misc{Zhang2020,
  doi = {10.48550/ARXIV.2009.04525},
  author = {Zhang, Enrui and Yin, Minglang and Karniadakis, George Em},
  title = {Physics-Informed Neural Networks for Nonhomogeneous Material Identification in Elasticity Imaging},
  publisher = {arXiv},
  year = {2020},
  copyright = {arXiv.org perpetual, non-exclusive license},
}

@article{Fuhg2022,
  title = {On physics-informed data-driven isotropic and anisotropic constitutive models through probabilistic machine learning and space-filling sampling},
  volume = {394},
  issn = {0045-7825},
  doi = {10.1016/j.cma.2022.114915},
  journal = {Computer Methods in Applied Mechanics and Engineering},
  publisher = {Elsevier BV},
  author = {Fuhg, Jan N. and Bouklas, Nikolaos},
  year = {2022},
  month = May,
  pages = {114915},
}

@article{Linden2023,
  title = {Neural networks meet hyperelasticity: A guide to enforcing physics},
  volume = {179},
  issn = {0022-5096},
  doi = {10.1016/j.jmps.2023.105363},
  journal = {Journal of the Mechanics and Physics of Solids},
  publisher = {Elsevier BV},
  author = {Linden, Lennart and Klein, Dominik K. and Kalina, Karl A. and Brummund, Jörg and Weeger, Oliver and Kästner, Markus},
  year = {2023},
  month = Oct,
  pages = {105363},
}

@article{Ta2023,
  title = {Benchmarking physics-informed frameworks for data-driven hyperelasticity},
  volume = {73},
  issn = {1432-0924},
  doi = {10.1007/s00466-023-02355-2},
  number = {1},
  journal = {Computational Mechanics},
  publisher = {Springer Science and Business Media LLC},
  author = {Taç, Vahidullah and Linka, Kevin and Sahli-Costabal, Francisco and Kuhl, Ellen and Tepole, Adrian Buganza},
  year = {2023},
  month = Jun,
  pages = {49–65},
}

@article{Haustrate2025,
  title = {Identifying heterogeneous elastic parameters from full-field measurements with reduced bases},
  volume = {305},
  issn = {0020-7403},
  doi = {10.1016/j.ijmecsci.2025.110682},
  journal = {International Journal of Mechanical Sciences},
  publisher = {Elsevier BV},
  author = {Haustrate, Rémi and Hild, François and Fau, Amélie},
  year = {2025},
  month = Nov,
  pages = {110682},
}

@misc{Kingma2014,
  doi = {10.48550/ARXIV.1412.6980},
  author = {Kingma, Diederik P. and Ba, Jimmy},
  title = {Adam: A Method for Stochastic Optimization},
  publisher = {arXiv},
  year = {2014},
  copyright = {arXiv.org perpetual, non-exclusive license},
}

@article{Treloar1943,
  title = {The elasticity of a network of long-chain molecules—II},
  volume = {39},
  issn = {0014-7672},
  doi = {10.1039/tf9433900241},
  number = {0},
  journal = {Trans. Faraday Soc.},
  publisher = {Royal Society of Chemistry (RSC)},
  author = {Treloar, L. R. G.},
  year = {1943},
  pages = {241–246},
}

@article{Mooney1940,
  title = {A Theory of Large Elastic Deformation},
  volume = {11},
  issn = {1089-7550},
  doi = {10.1063/1.1712836},
  number = {9},
  journal = {Journal of Applied Physics},
  publisher = {AIP Publishing},
  author = {Mooney, M.},
  year = {1940},
  month = Sep,
  pages = {582–592},
}

@article{Rivlin1948,
  title = {Large elastic deformations of isotropic materials IV. further developments of the general theory},
  volume = {241},
  issn = {2054-0272},
  doi = {10.1098/rsta.1948.0024},
  number = {835},
  journal = {Philosophical Transactions of the Royal Society of London. Series A, Mathematical and Physical Sciences},
  publisher = {The Royal Society},
  author = {Rivlin, R. S.},
  year = {1948},
  month = Oct,
  pages = {379–397},
}

@article{Holzapfel2000,
  author = {Holzapfel, Gerhard A. and Gasser, Thomas C. and Ogden, Ray W.},
  title = {A New Constitutive Framework for Arterial Wall Mechanics and a Comparative Study of Material Models},
  journal = {Journal of Elasticity},
  year = {2000},
  volume = {61},
  issn = {0374-3535},
  doi = {10.1023/a:1010835316564},
  number = {1/3},
  publisher = {Springer Science and Business Media LLC},
  pages = {1–48},
}

@article{Hartmann2003,
  title = {Polyconvexity of generalized polynomial-type hyperelastic strain energy functions for near-incompressibility},
  volume = {40},
  issn = {0020-7683},
  doi = {10.1016/s0020-7683(03)00086-6},
  number = {11},
  journal = {International Journal of Solids and Structures},
  publisher = {Elsevier BV},
  author = {Hartmann, Stefan and Neff, Patrizio},
  year = {2003},
  month = Jun,
  pages = {2767–2791},
}

@article{Roux2020,
  title = {Optimal procedure for the identification of constitutive parameters from experimentally measured displacement fields},
  volume = {184},
  issn = {0020-7683},
  doi = {10.1016/j.ijsolstr.2018.11.008},
  journal = {International Journal of Solids and Structures},
  publisher = {Elsevier BV},
  author = {Roux, Stéphane and Hild, François},
  year = {2020},
  month = Feb,
  pages = {14–23},
}

@article{Krijnen2026,
  title = {Unsupervised full-field Bayesian inference of orthotropic hyperelasticity from a single biaxial test: a myocardial case study},
  volume = {459},
  issn = {0045-7825},
  doi = {10.1016/j.cma.2026.119034},
  journal = {Computer Methods in Applied Mechanics and Engineering},
  publisher = {Elsevier BV},
  author = {Krijnen, Rogier P. and Joshi, Akshay and Kumar, Siddhant and Peirlinck, Mathias},
  year = {2026},
  month = Sep,
  pages = {119034},
}

@article{Promma2009,
  title = {Application of the virtual fields method to mechanical characterization of elastomeric materials},
  volume = {46},
  issn = {0020-7683},
  doi = {10.1016/j.ijsolstr.2008.09.025},
  number = {3-4},
  journal = {International Journal of Solids and Structures},
  publisher = {Elsevier BV},
  author = {Promma, N. and Raka, B. and Grédiac, M. and Toussaint, E. and Le Cam, J.-B. and Balandraud, X. and Hild, F.},
  year = {2009},
  month = Feb,
  pages = {698–715},
}

@article{sheets2012standard,
  author = {Sheets, Standard Vulcanized and from Products, Purposes},
  title = {Standard Test Methods for Vulcanized Rubber and Thermoplastic Elastomers—},
  journal = {Annu. B. ASTM Stand},
  year = {2012},
  volume = {1},
  pages = {1--14},
  doi = {10.1520/D0412-16R21},
}

@article{Flaschel2021,
  title = {Unsupervised discovery of interpretable hyperelastic constitutive laws},
  volume = {381},
  issn = {0045-7825},
  doi = {10.1016/j.cma.2021.113852},
  journal = {Computer Methods in Applied Mechanics and Engineering},
  publisher = {Elsevier BV},
  author = {Flaschel, Moritz and Kumar, Siddhant and De Lorenzis, Laura},
  year = {2021},
  month = Aug,
  pages = {113852},
}

@article{Tan2017,
  title = {Gradient-Based Optimization for Poroelastic and Viscoelastic MR Elastography},
  volume = {36},
  issn = {1558-254X},
  doi = {10.1109/tmi.2016.2604568},
  number = {1},
  journal = {IEEE Transactions on Medical Imaging},
  publisher = {Institute of Electrical and Electronics Engineers (IEEE)},
  author = {Tan, Likun and McGarry, Matthew D. J. and Van Houten, Elijah E. W. and Ji, Ming and Solamen, Ligin and Weaver, John B. and Paulsen, Keith D.},
  year = {2017},
  month = Jan,
  pages = {236–250},
}

@article{Marino2023,
  title = {Automated identification of linear viscoelastic constitutive laws with EUCLID},
  volume = {181},
  issn = {0167-6636},
  doi = {10.1016/j.mechmat.2023.104643},
  journal = {Mechanics of Materials},
  publisher = {Elsevier BV},
  author = {Marino, Enzo and Flaschel, Moritz and Kumar, Siddhant and De Lorenzis, Laura},
  year = {2023},
  month = Jun,
  pages = {104643},
}

@article{Flaschel2022,
  title = {Discovering plasticity models without stress data},
  volume = {8},
  issn = {2057-3960},
  doi = {10.1038/s41524-022-00752-4},
  number = {1},
  journal = {npj Computational Materials},
  publisher = {Springer Science and Business Media LLC},
  author = {Flaschel, Moritz and Kumar, Siddhant and De Lorenzis, Laura},
  year = {2022},
  month = Apr,
}

@article{Tripura2023,
  title = {A wavelet neural operator based elastography for localization and quantification of tumors},
  volume = {232},
  issn = {0169-2607},
  doi = {10.1016/j.cmpb.2023.107436},
  journal = {Computer Methods and Programs in Biomedicine},
  publisher = {Elsevier BV},
  author = {Tripura, Tapas and Awasthi, Abhilash and Roy, Sitikantha and Chakraborty, Souvik},
  year = {2023},
  month = Apr,
  pages = {107436},
}

@article{Liu2025,
  title = {Resolving 3D heterogeneous mechanical property identification via dropout-regularized physics-informed deep learning with experimental validation},
  volume = {47},
  issn = {2352-4928},
  doi = {10.1016/j.mtcomm.2025.113225},
  journal = {Materials Today Communications},
  publisher = {Elsevier BV},
  author = {Liu, Yizhe},
  year = {2025},
  month = Jul,
  pages = {113225},
}

@manual{vic2dmanual7,
  title        = {Vic-2D Reference Manual (Version 7)},
  author       = {{Correlated Solutions, Inc.}},
  organization = {Correlated Solutions, Inc.},
  address      = {Irmo, SC, USA},
  year         = {2025},
  note         = {Software version 7.2.68, build 4509}
}

@article{DallAra2022,
  title = {Digital volume correlation for the characterization of musculoskeletal tissues: Current challenges and future developments},
  volume = {10},
  issn = {2296-4185},
  doi = {10.3389/fbioe.2022.1010056},
  journal = {Frontiers in Bioengineering and Biotechnology},
  publisher = {Frontiers Media SA},
  author = {Dall’Ara, Enrico and Tozzi, Gianluca},
  year = {2022},
  month = Oct,
}

@article{Isihara1951,
  title = {Statistical Theory of Rubber-Like Elasticity. IV. (Two-Dimensional Stretching)},
  volume = {19},
  issn = {1089-7690},
  doi = {10.1063/1.1748111},
  number = {12},
  journal = {The Journal of Chemical Physics},
  publisher = {AIP Publishing},
  author = {Isihara, Akira and Hashitsume, Natsuki and Tatibana, Masao},
  year = {1951},
  month = Dec,
  pages = {1508–1512},
}

\end{document}

% ---- S-prefixed numbering ----
\renewcommand{\thesection}{S\arabic{section}}
\renewcommand{\thesubsection}{S\arabic{section}.\arabic{subsection}}
\renewcommand{\thefigure}{S\arabic{figure}}
\let\oldthetable\thetable
\renewcommand{\thetable}{S\oldthetable}
\renewcommand{\theequation}{S\arabic{equation}}
\raggedbottom

\begin{frontmatter}



\title{\textbf{Supplementary Information -- }Interpretable identification of constitutive parameter field for three-dimensional heterogeneous and graded hyperelastic materials from full-field displacements}


\author[iisc_label]{Kanhaiya Lal Chaurasiya}            
\author[iisc_label]{Vaskar Halder\fnref{equal}}  
\author[iisc_label]{Prajyot Patil\fnref{equal}} 
\author[iisc_label]{Gaurav Navnath Shinde}              
\author[iisc_label]{Debashish Das}              
\author[iisc_label]{Akshay Joshi\corref{cor1}}  
\cortext[cor1]{Corresponding author}
\ead{akshayjoshi@iisc.ac.in}

\fntext[equal]{These authors contributed equally.}

\affiliation[iisc_label]{organization={Department of Mechanical Engineering, Indian Institute of Science},%Department and Organization 
            city={Bengaluru},
            country={India}}

\begin{abstract}
The inverse problem of identifying spatially varying material properties of heterogeneous solids using experimentally available displacement fields and boundary reaction forces alone remains challenging. Being able to characterize heterogeneous and graded solids would be advantageous in the design and manufacturing of functionally graded materials and composites. Additionally, an efficient technique for identifying heterogeneous material properties from in vivo measurements would be useful in biomedical contexts to characterize heterogeneous biological tissues. Although several techniques developed over the past two decades have addressed inverse identification for homogeneous materials, including linear elastic, hyperelastic, viscoelastic, and elastoplastic solids, comparatively fewer studies have considered heterogeneous and graded materials. Furthermore, several recent approaches to characterizing heterogeneous and graded materials employ parameter-heavy, deep learning-based frameworks trained to enforce force equilibrium at specified collocation points within the material. In this work, we propose an interpretable and time-efficient framework that predicts the three-dimensional relative spatial variation of a constitutive parameter field for multiple parameters of a heterogeneous hyperelastic material, using only displacement-field data. To achieve this, we subdivide the geometry into finite elements and formulate an objective function that depends on the constitutive parameters at each finite element, rather than on the parameters of a neural network that outputs the constitutive parameters. The objective function enforces force equilibrium at the integration point of each element, along with additional regularization terms to ensure realistic and physically consistent constitutive parameters. In this work, we validate our framework on 3D synthetic and 2D experimental benchmarks involving heterogeneous, graded, and anisotropic hyperelastic materials. By leveraging the linear nature of the inverse problem for hyperelastic materials and directly optimizing for the constitutive parameters at the finite-element level, we achieve acceptable prediction accuracy with significantly lower computational time compared to recently reported deep-learning-based frameworks.  
\end{abstract}

%% Keywords
\begin{keyword}
Heterogeneity \sep hyperelastic material \sep constrained optimization \sep elastography  \sep digital image correlation
\end{keyword}

\end{frontmatter}


\tableofcontents

\section{The objective as a posterior distribution on \texorpdfstring{$\bm{\theta}$}{theta}, given \texorpdfstring{$\bm{G}$}{G}}

\begin{figure}[H]
        \centering
        \includegraphics[width=0.75\linewidth]{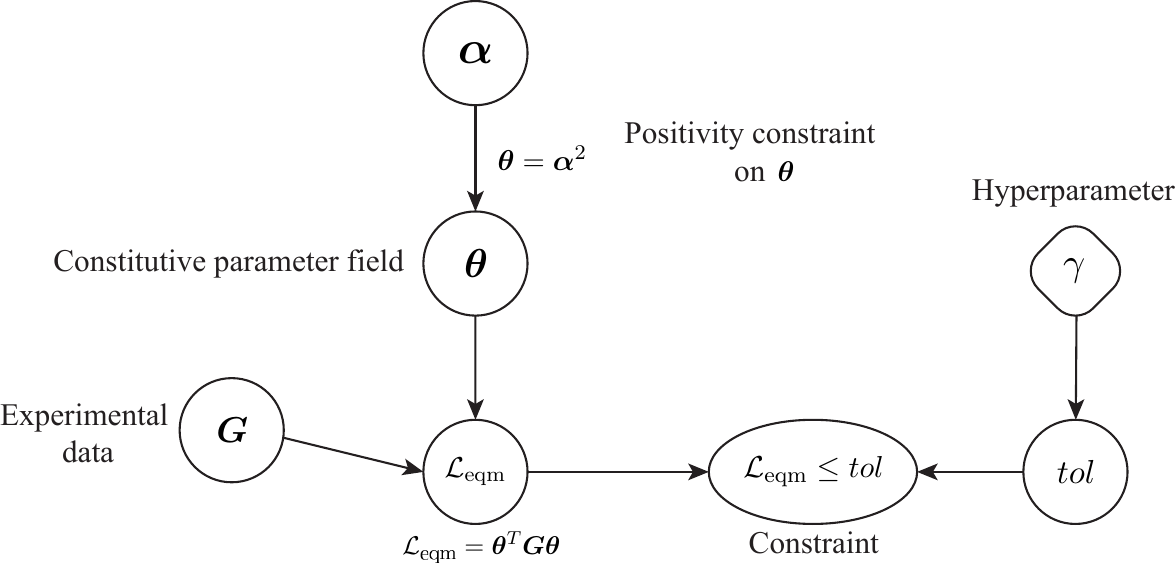}
        \caption{\textbf{Dependency graph for the Bayesian formulation.} The unconstrained variables \(\boldsymbol{\alpha}\) define the constitutive parameter field through \(\boldsymbol{\theta}=\boldsymbol{\alpha}^2\), ensuring positivity. The equilibrium loss \(\mathcal{L}_{\mathrm{eqm}}=\boldsymbol{\theta}^{T}\boldsymbol{G}\boldsymbol{\theta}\) depends on the parameter field and the data-derived matrix \(\boldsymbol{G}\). The tolerance \(tol\), governed by the hyperparameter \(\gamma\), imposes the condition \(\mathcal{L}_{\mathrm{eqm}}\leq tol\), which enters the Bayesian inference of the constitutive parameters. Arrows indicate the dependencies between the quantities.}

        \label{fig_dependency}
    \end{figure}

With reference to the Eq. (19) in the main text, we obtain that the correct $\boldsymbol{\theta}$ minimizes $\boldsymbol{\theta}^{T}\boldsymbol{G}\boldsymbol{\theta}$ needs to be minimized, subject to three physical constraints. If we treat $\boldsymbol{\theta}$ as a random variable, we can interpret the physical constraints as priors on the random variable. We introduce the dependency graph and the various random variables in Fig.~\ref{fig_dependency}. The problem can thus be recast as finding the Maximum \textit{a posteriori} estimate (MAP) for the posterior on $\boldsymbol{\theta}$, given the experimental displacement field data across multiple loading steps, encapsulated in the $\boldsymbol{G}$ matrix, i.e.,
\begin{equation}
    \boldsymbol{\theta}_{\text{opt}} = \underset{\boldsymbol{\theta}}{\text{argmin}}~ -\ln \mathcal{P}\left(\boldsymbol{\theta}|\boldsymbol{G}\right)
\end{equation}

To achieve this, we make use of the dependencies shown in Fig.~\ref{fig_dependency}. Note that due to the positive semi-definite nature of $\boldsymbol{G}$, $\mathcal{L}_{\text{eqm}}$ is always non-negative.

We define the necessary probability density functions for the various random variables as follows
\begin{align}
    &\mathcal{P}\left(\boldsymbol{\theta}\right) \propto \exp\left(-\sum_{j=1}^{n_e}\frac{\beta}{\theta_1^{(j)}}\right) \exp\left(-\frac{1}{2}\sum_{i=1}^{\mathit{n_f}}\sum_{p=1}^{n_e}\sum_{q=1}^{n_e}\theta_i^{(p)}\mathcal{S}^{-1}_{pq}\theta_i^{(q)}\right)\mathds{1}_{\boldsymbol{\theta}^{+}} \quad \text{(Prior on $\boldsymbol{\theta}$)}\label{eq:theta_prior}
    \\
    &\text{Where, \quad}
    \mathds{1}_{\boldsymbol{\theta}^{+}} = \begin{cases}
        1 & \text{if } \boldsymbol{\theta}_i > 0 \text{ $\forall$ } i\\
        0 & \text{if } \text{ $\exists$ }i: \boldsymbol{\theta}_i \leq 0
    \end{cases} \text{\quad defines a positive truncation to the probability density.}
\end{align}
The prior on the random variable $tol$ defined as
\begin{equation}
    \mathcal{P}\left(tol\right) \propto \exp\left(-\gamma ~ tol\right) \label{eq:tolprior}
\end{equation}
Next, we define the conditional distribution on $\mathcal{L}_{\text{eqm}}$
\begin{equation}
    \mathcal{P}\left(\mathcal{L}_{\text{eqm}}\right | \boldsymbol{G}, \boldsymbol{\theta})  = \delta\left(\mathcal{L}_{\text{eqm}} - \boldsymbol{\theta}^{T}\boldsymbol{G}\boldsymbol{\theta}\right)
    \label{eqn:LgivenGtheta}
\end{equation}
where $\delta$ indicates the Dirac delta function.
The conditional relation \cref{eqn:LgivenGtheta}, can be used to obtain a cumulative density estimate for the ``tolerance'' constraint $\mathcal{L}_{\text{eqm}}\leq tol$, given $\boldsymbol{G}$ and $\boldsymbol{\theta}$ as follows
\begin{align}
    \mathcal{P}\left(\mathcal{L}_{\text{eqm}}\leq tol~|\boldsymbol{G}, \boldsymbol{\theta}\right) &=  \int_{0}^{tol} \mathcal{P}\left(\mathcal{L}_{\text{eqm}}\right | \boldsymbol{G}, \boldsymbol{\theta})~d(\mathcal{L}_{\text{eqm}}) \\
    & = H \left(tol-\boldsymbol{\theta}^{T}\boldsymbol{G}\boldsymbol{\theta}\right) \text{\quad where, $H(x)$ denotes the Heaviside step function}
    \label{eqn:CDF}
\end{align}

Referring to the dependency graph in Fig.~\ref{fig_dependency},
\begin{equation}
    \mathcal{P}\left(\boldsymbol{\theta}\right|\boldsymbol{G}) \propto \int_{tol=-\infty}^{+\infty} \mathcal{P}\left(\boldsymbol{\theta}~|\boldsymbol{G},\mathcal{L}_{\text{eqm}} \leq tol\right)~\mathcal{P}(tol) ~d(tol)
    \label{eqn:theInteq}
\end{equation}

Using Bayes' theorem,
\begin{equation}    \mathcal{P}\left(\boldsymbol{\theta}|\boldsymbol{G},\mathcal{L}_{\text{eqm}}\leq tol\right) \propto \mathcal{P}\left(\mathcal{L}_{\text{eqm}}\leq tol~| \boldsymbol{G}, \boldsymbol{\theta}\right) ~\mathcal{P}(\boldsymbol{\theta})
\label{eq:BayesTh}
\end{equation}

Thus, using \cref{eq:theta_prior,eq:tolprior,eqn:CDF,eq:BayesTh}, \cref{eqn:theInteq} can be simplified as follows
\begin{align}
    \mathcal{P}\left(\boldsymbol{\theta}\right|\boldsymbol{G}) &\propto
    \int_{tol=-\infty}^{+\infty} H \left(tol-\boldsymbol{\theta}^{T}\boldsymbol{G}\boldsymbol{\theta}\right) \mathcal{P}\left(\boldsymbol{\theta}\right) \exp(-\gamma ~ tol)~d(tol) \\
    & \propto \int_{tol=\boldsymbol{\theta}^{T}\boldsymbol{G}\boldsymbol{\theta}}^{+\infty} \mathcal{P}\left(\boldsymbol{\theta}\right) \exp(-\gamma ~ tol)~d(tol) \\
    & \propto
    \exp\left(-\gamma \text{ }\boldsymbol{\theta}^{T}\boldsymbol{G}\boldsymbol{\theta}\right) \mathcal{P}\left(\boldsymbol{\theta}\right) \\
    & \propto \exp\left(-\gamma \text{ }\boldsymbol{\theta}^{T}\boldsymbol{G}\boldsymbol{\theta}\right) \exp\left(-\sum_{j=1}^{n_e}\frac{\beta}{\theta_1^{(j)}}\right) \exp\left(-\frac{1}{2}\sum_{i=1}^{\mathit{n_f}}\sum_{p=1}^{n_e}\sum_{q=1}^{n_e}\theta_i^{(p)}\mathcal{S}^{-1}_{pq}\theta_i^{(q)}\right)\mathds{1}_{\boldsymbol{\theta}^{+}}
    \label{eq:penulatimate}
\end{align}

Thus,
\begin{equation}
    -\frac{1}{\gamma}\ln \mathcal{P}\left(\boldsymbol{\theta}|\boldsymbol{G}\right) = \boldsymbol{\theta}^{T}\boldsymbol{G}\boldsymbol{\theta} + \frac{\beta}{\gamma} \sum_{j=1}^{n_e}\frac{1}{\theta_1^{(j)}} + \frac{1}{2\gamma} \sum_{i=1}^{\mathit{n_f}}\sum_{p=1}^{n_e}\sum_{q=1}^{n_e}\theta_i^{(p)}\mathcal{S}^{-1}_{pq}\theta_i^{(q)} - \frac{1}{\gamma}\ln \mathds{1}_{\boldsymbol{\theta}^{+}} + \text{const.}
    \label{eqn:finalTheta}
\end{equation}

As discussed in the main text, by recasting the optimization objective in terms of $\boldsymbol{\alpha}^2$ instead of $\boldsymbol{\theta}$, we get $\mathds{1}_{\boldsymbol{\theta}^{+}} = 1$ for all $\boldsymbol{\alpha}\in \mathbb{R}^{n_e \mathit{n_f} \times 1}$.
After ignoring the scaling and additive constant terms, we thus recover the objective function shown in the main text:
\begin{equation}
    \mathcal{L}(\boldsymbol{\alpha}) =
        \left(\boldsymbol{\alpha}^2\right)^{T} \boldsymbol{G} \boldsymbol{\alpha}^2
        +
        \lambda_{r} \sum\limits_{j=1}^{\mathit{n_e}} \dfrac{1}{\left(\alpha_1^{(j)}\right)^2}
        +
        \lambda_{s}\sum\limits_{j=1}^{\mathit{n_f}}\sum\limits_{p=1}^{\mathit{n_e}}\sum\limits_{q=1}^{\mathit{n_e}} \left(\alpha^{(p)}_{j}\right)^2
        \mathcal{S}^{-1}_{pq} \left(\alpha^{(q)}_{j}\right)^2
    \label{eq_Lossfunction_alpha}
\end{equation}
where \(\lambda_{r}=\dfrac{\beta}{\gamma}\) and \(\lambda_{s}=\dfrac{1}{2\gamma}\)

\section{Spatially varying heteroscedastic Gaussian displacement noise}
    \begin{figure}[H]
        \centering
        \includegraphics[width=\linewidth]{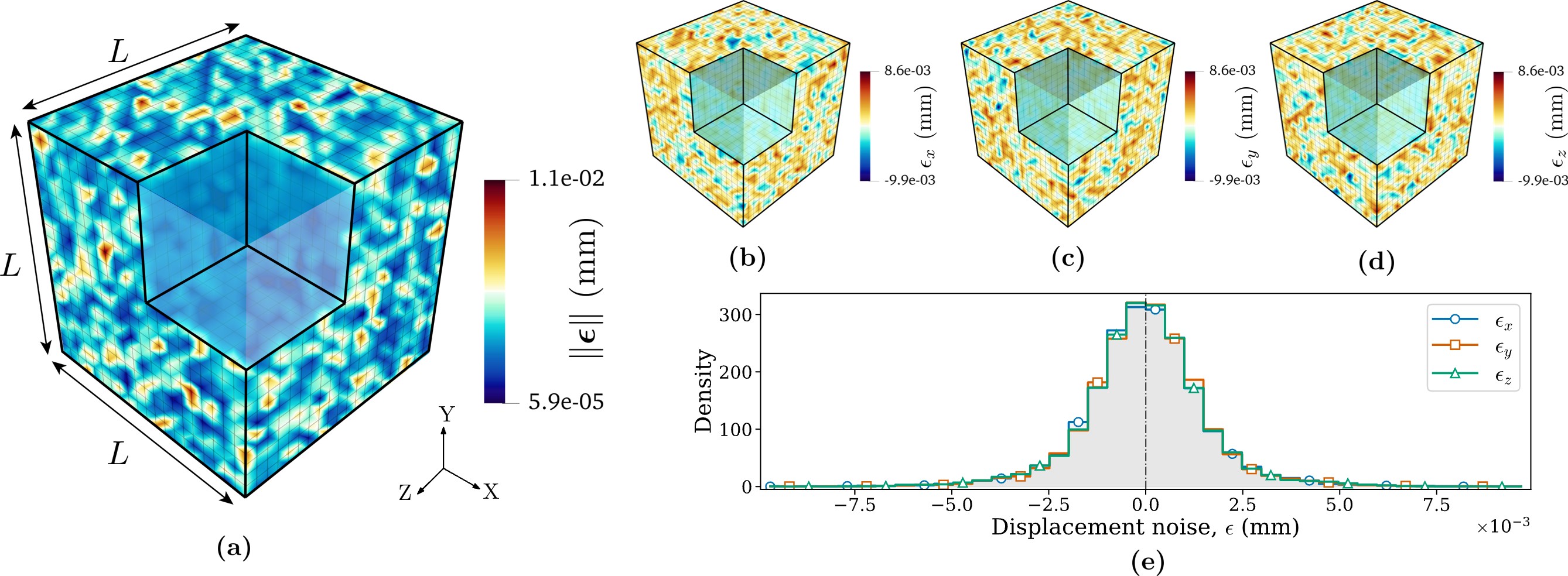}
        \caption{\textbf{Spatially varying heteroscedastic Gaussian displacement noise.} \textbf{a}, Spatial distribution of the displacement-noise magnitude, \(\lVert\boldsymbol{\epsilon}\rVert\), showing the increased noise amplitude near the specimen boundaries. \textbf{b-d}, Corresponding noise fields for the individual displacement components \(\epsilon_x\), \(\epsilon_y\), and \(\epsilon_z\), respectively. \textbf{e}, Probability-density distributions of the three displacement-noise components.}
        \label{fig_spatial_noise}
    \end{figure}
To examine the robustness of the proposed method to measurement uncertainty, the displacement fields obtained from the FE simulations are perturbed using a spatially varying heteroscedastic Gaussian noise model, in which the displacement-noise variance varies spatially and increases near the specimen boundaries \citep{DallAra2022}. To achieve this, we follow procedure outlined in Krijnen et al. \citep{Krijnen2026}. For a node `\(a\)' located at \(\boldsymbol{X}^{a}=(x^{a},y^{a},z^{a})\), we first compute its shortest distance from the external boundary of the cubical domain as
\begin{equation}
    d^{a}    =    \min  \left(
    \{x^{a}-x_{\min}\},
    \{x_{\max}-x^{a}\},
    \{y^{a}-y_{\min}\},
    \{y_{\max}-y^{a}\},
    \{z^{a}-z_{\min}\},
    \{z_{\max}-z^{a} \}   \right)
    \label{eq_noise_boundary_distance}
\end{equation}
The local noise standard deviation is then prescribed as
\begin{equation}
    \sigma^{a}    =
    \sigma_{\boldsymbol{u}}
    \sqrt{ 1 +  \alpha \exp\left(
    -\frac{(d^{a})^{2}}{2\lambda_{\boldsymbol{u}}^{2}}
    \right) }
    \label{eq_spatial_noise_std}
\end{equation}

where \(\sigma_{\boldsymbol{u}}\) defines the baseline displacement-noise level, \(\alpha\) controls the increase in noise near the external boundary, and \(\lambda_{\boldsymbol{u}}\) governs the distance over which this boundary effect decays. In the present study, \(\sigma_{\boldsymbol{u}} = 1 \times 10^{-3}\), \(\alpha=5.0\) and \(\lambda_{\boldsymbol{u}}=1.0\) are used. The displacement-noise vector at each node is independently sampled as
\begin{equation}
    \boldsymbol{\epsilon}^{a}
    \sim    \mathcal{N}
    \left( \boldsymbol{0},  (\sigma^{a})^{2}\boldsymbol{I}
    \right),    \qquad    \boldsymbol{u}_{\mathrm{noisy}}^{a}
    =   \boldsymbol{u}^{a}    +   \boldsymbol{\epsilon}^{a}
    \label{eq_noisy_displacement}
\end{equation}
This construction introduces a non-uniform noise level across the specimen, with larger fluctuations near the external surfaces and a lower baseline level within the interior.  Fig.~\ref{fig_spatial_noise} shows the three-dimensional distributions of the imposed noise and the corresponding component-wise (\(\epsilon_x\), \(\epsilon_y\), and \(\epsilon_z\)) histograms.

\section{Supplementary results}
\label{sec_supplementary}

\subsection{Smooth monotonic grading}

\begin{figure}[H]
        \centering
        \includegraphics[width=\linewidth]{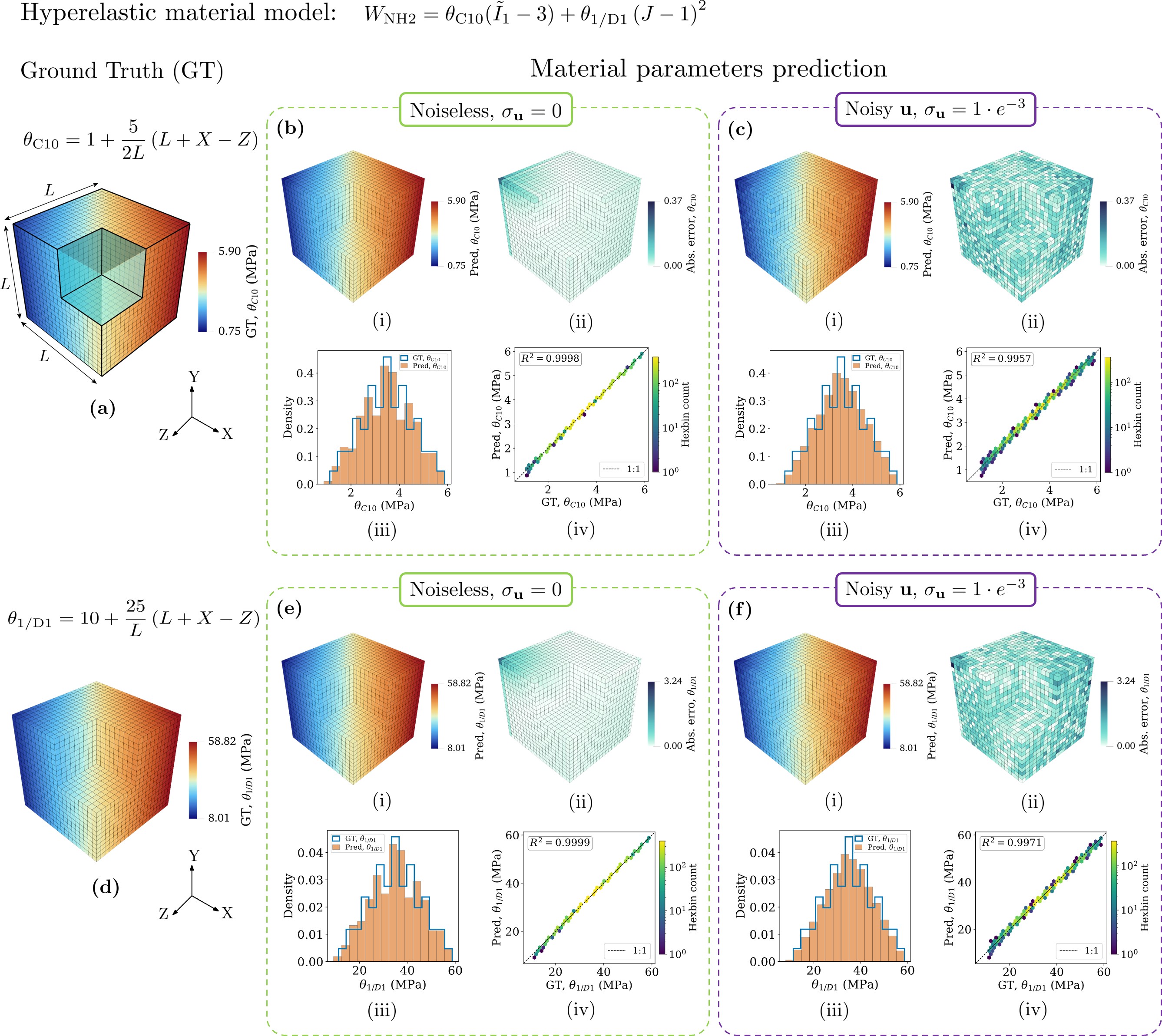}
        \caption{\textbf{Smooth monotonic linear gradation for Neo-Hookean material.}  The ground truth (GT) spatial distributions of \(\boldsymbol{\theta}_{\mathrm{C10}}\) and \(\boldsymbol{\theta}_{\mathrm{1/D1}}\) are visualized in \textbf{a} and \textbf{d}, respectively. Figure panels \textbf{b} and \textbf{e} correspond to the inverse identification results for noiseless displacement data,  showing sub-figures (i) the predicted (Pred.) material parameter field, (ii) the absolute error, (iii) comparison of the GT and predicted parameter histogram distributions, and (iv) hexbin parity plots between the ground truth and predicted values. Figure panels \textbf{c} and \textbf{f} highlight the corresponding results for noise-perturbed displacement data, presented using the same quantities. The dashed diagonal line in the parity plots denotes the \(1{:}1\) reference, and the corresponding \(R^2\) values quantify agreement between the predicted and ground truth parameter fields.}
        \label{fig_3D_xzdiag_NH}
    \end{figure}

In this test case, we consider a specimen with a monotonic linear variation of material properties along the X-Z face diagonal and evaluate the performance of the proposed inverse identification framework. The compressible Neo-Hookean (NH2) model given in Eq.~\eqref{eq_NH2} is adopted, consisting of two independent constitutive terms, \((\tilde{I}_{1}-3)\) and \(\left(J-1\right)^{2}\), with the corresponding parameter fields \(\boldsymbol{\theta}_{\mathrm{C10}}\) and \(\boldsymbol{\theta}_{\mathrm{1/D1}}\). The prescribed spatial variations of these parameter fields are given in Eq.~\eqref{eq_3D_xzdiag_NH}, with their ground-truth distributions visualized using octant cut-sections in Fig.~\ref{fig_3D_xzdiag_NH}a and \ref{fig_3D_xzdiag_NH}d. The cubic domain is discretized using 8,000 C3D8 elements and subjected to the loading conditions listed in Table~3. The optimization is performed for 60,000 epochs using both noiseless and noise-perturbed displacement data with convergence histories shown in Fig.~\ref{fig_3D_xzdiag_NH}g and \ref{fig_3D_xzdiag_NH}h, requiring 16.93 min and 17.02 min, respectively. For the noiseless case, the recovered \(\boldsymbol{\theta}_{\mathrm{C10}}\) and \(\boldsymbol{\theta}_{\mathrm{1/D1}}\) fields shown in Fig.~\ref{fig_3D_xzdiag_NH}b and \ref{fig_3D_xzdiag_NH}e closely follow the prescribed linear variation, with \(R^2\) values of 0.9998 and 0.9999, respectively. Under the noise-perturbed condition, the corresponding results in Fig.~\ref{fig_3D_xzdiag_NH}c and \ref{fig_3D_xzdiag_NH}f retain good agreement with the ground truth, with the histogram and hexbin parity plots giving \(R^2\) values of 0.9957 and 0.9971 for \(\boldsymbol{\theta}_{\mathrm{C10}}\) and \(\boldsymbol{\theta}_{\mathrm{1/D1}}\), respectively. \\ \par

\noindent\textbf{ Neo-Hookean model (NH2)}: \textit{Face diagonal linear property gradient}
\begin{equation}
     W_{\text{NH2}} = \theta_{\mathrm{C10}}(\tilde{I}_{1}-3)
                   + \theta_{1/ \mathrm{D1}} \left(J-1\right)^{2}
\label{eq_NH2}
\end{equation}
where,
\begin{equation}
\begin{split}
    \theta_{\mathrm{C10}}(X,Y,Z)  &= 1+\frac{5}{2L}\left(L+X-Z\right) \\
    \theta_{1/ \mathrm{D1}}(X,Y,Z) &= 10+\frac{25}{L}\left(L+X-Z\right)
\end{split}
\label{eq_3D_xzdiag_NH}
\end{equation}

\subsection{Sharp discontinuous heterogeneity}
\begin{figure}[H]
        \centering
        \includegraphics[width=\linewidth]{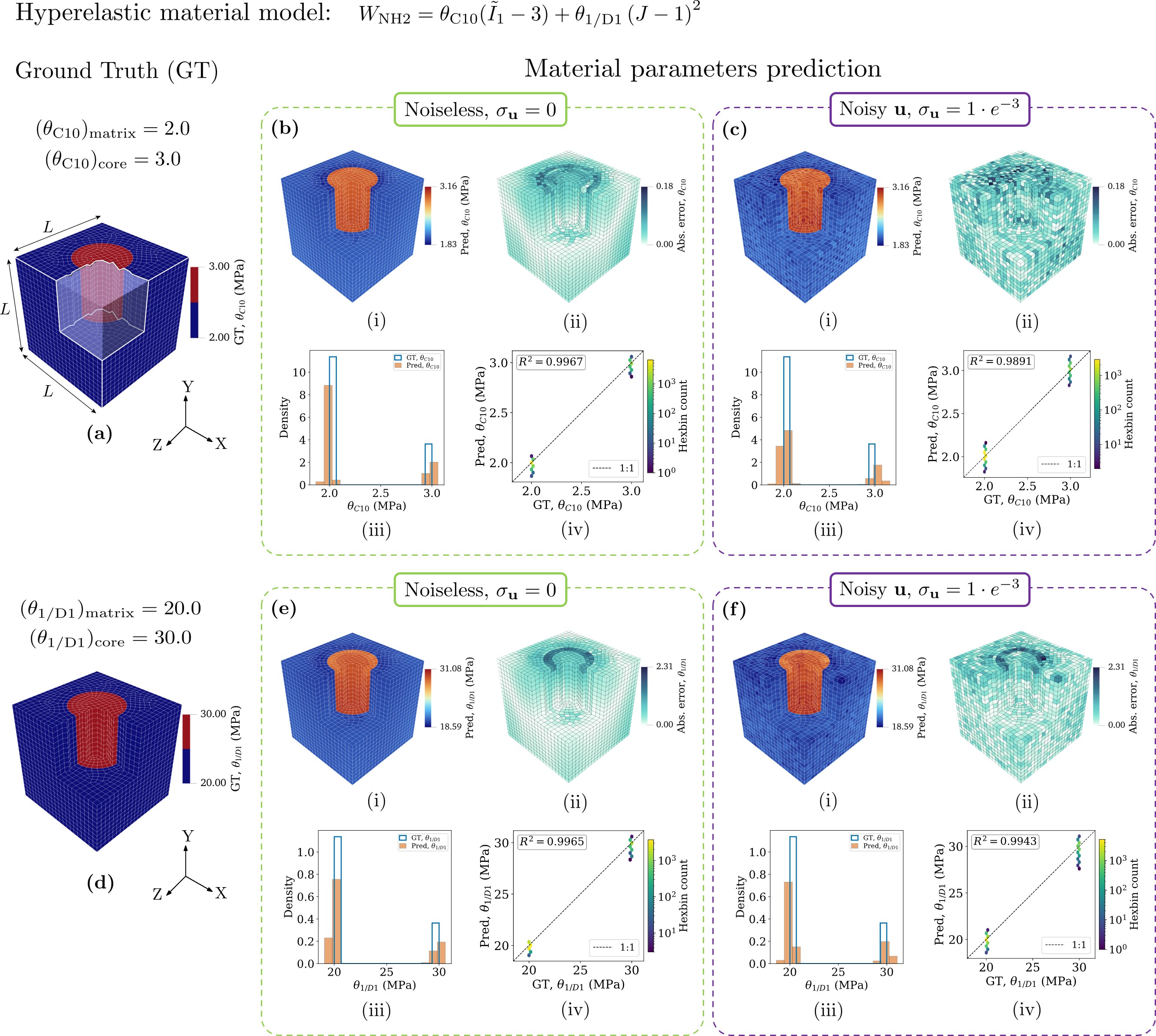}
        \caption{\textbf{Sharp discontinuous heterogeneity for Neo-Hookean material.}  The ground truth (GT) spatial distributions of \(\boldsymbol{\theta}_{\mathrm{C10}}\) and \(\boldsymbol{\theta}_{\mathrm{1/D1}}\) are visualized in \textbf{a} and \textbf{d}, respectively. Figure panels \textbf{b} and \textbf{e} correspond to the inverse identification results for noiseless displacement data,  showing sub-figures (i) the predicted (Pred.) material parameter field, (ii) the absolute error, (iii) comparison of the GT and predicted parameter histogram distributions, and (iv) hexbin parity plots between the ground truth and predicted values. Figure panels \textbf{c} and \textbf{f} highlight the corresponding results for noise-perturbed displacement data, presented using the same quantities. The dashed diagonal line in the parity plots denotes the \(1{:}1\) reference, and the corresponding \(R^2\) values quantify agreement between the predicted and ground truth parameter fields.}
        \label{fig_3D_cylincube_NH}
    \end{figure}

The preceding test cases considered in the present work cover smoothly varying constitutive fields. We next examine the performance of the proposed method on a material distribution with a sharp interface at the inter-segment boundary. A compressible Neo-Hookean formulation, given in Eq.~\eqref{eq_NH2}, is adopted for a bi-material specimen consisting of a stiffer cylindrical core embedded within a softer surrounding matrix, as shown in Fig.~\ref{fig_3D_cylincube_NH}a and \ref{fig_3D_cylincube_NH}d. The material parameters are prescribed as \((\theta_{\mathrm{C10}})_{\mathrm{core}}=3.0\)~MPa and \((\theta_{1/\mathrm{D1}})_{\mathrm{core}}=30.0\)~MPa within the cylindrical inclusion, and \((\theta_{\mathrm{C10}})_{\mathrm{matrix}}=2.0\)~MPa and \((\theta_{1/\mathrm{D1}})_{\mathrm{matrix}}=20.0\)~MPa in the surrounding matrix. The domain is discretized using 11,445 C3D8 elements and subjected to the mixed loading conditions listed in Table~3. For the noiseless case, the recovered parameter fields and corresponding absolute errors are presented in Fig.~\ref{fig_3D_cylincube_NH}b and \ref{fig_3D_cylincube_NH}e. Both \(\boldsymbol{\theta}_{\mathrm{C10}}\) and \(\boldsymbol{\theta}_{\mathrm{1/D1}}\) recover the cylindrical interface at the material boundary, giving \(R^2\) values of 0.9967 and 0.9965, respectively. Under the noise-perturbed condition, the corresponding results in Fig.~\ref{fig_3D_cylincube_NH}c and \ref{fig_3D_cylincube_NH}f continue to distinguish the two material regions clearly, as also reflected by the two distinct peaks in the histogram distributions, with \(R^2\) values of 0.9861 and 0.9943 for \(\boldsymbol{\theta}_{\mathrm{C10}}\) and \(\boldsymbol{\theta}_{\mathrm{1/D1}}\), respectively. The optimization is performed for 60,000 epochs, requiring 28.76 min for the noiseless case and 29.65 min for the noise-perturbed case as shown in Fig.~\ref{fig_3D_cylincube_NH}g and \ref{fig_3D_cylincube_NH}h, respectively. These results show that the framework can recover sharp constitutive transitions in addition to the smoothly varying material fields considered earlier. \par        

\subsection{Ablation study of the loss function}
    \begin{figure}[H]
        \centering
        \includegraphics[width=\linewidth]{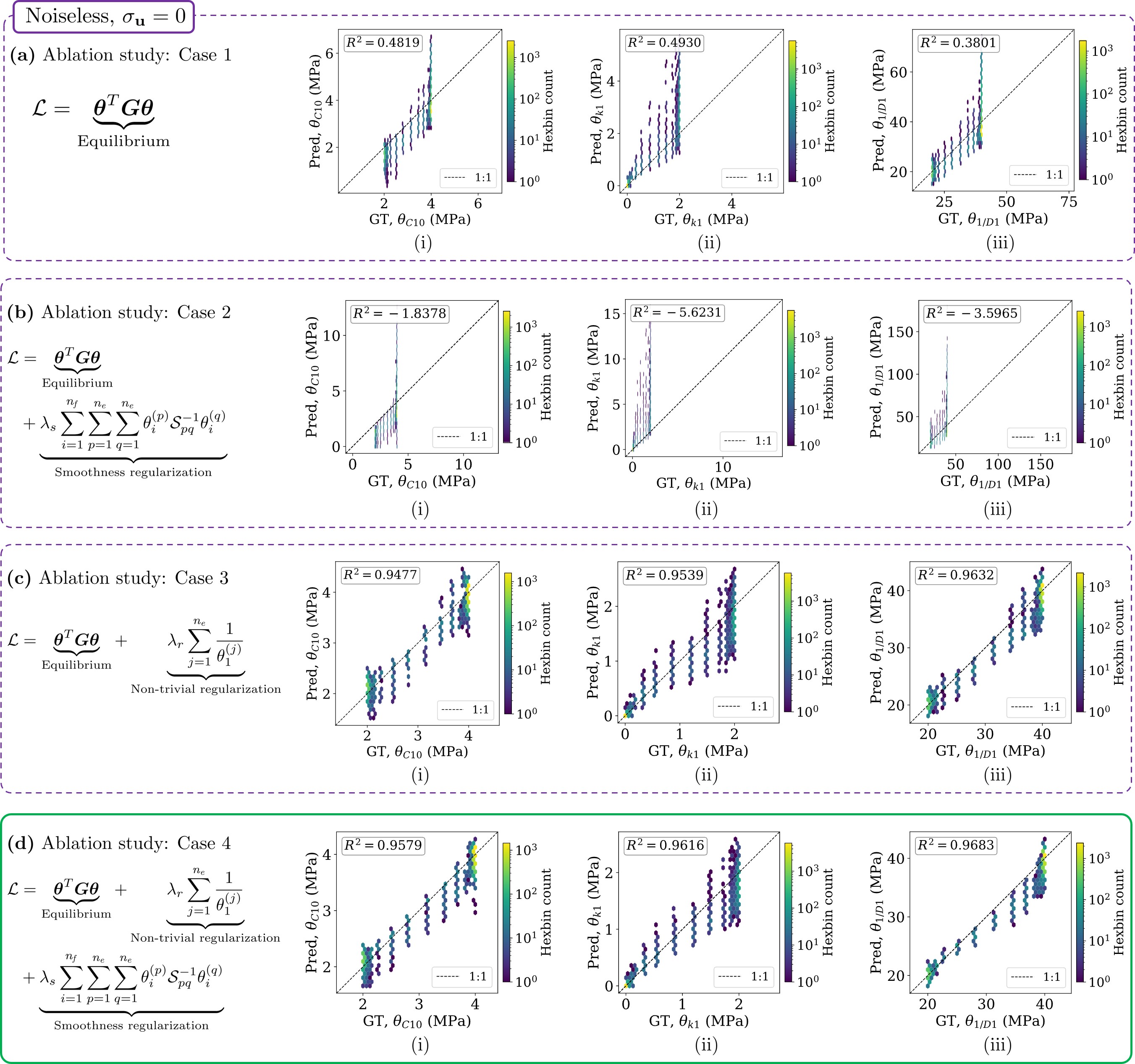}
        \caption{\textbf{Ablation study of the loss-function terms for the anisotropic HGO* material model.} The influence of different loss-function terms on the inverse identification of constitutive parameters is examined using noiseless displacement measurements (\(\sigma_u=0\)). \textbf{a}, Case 1: equilibrium loss alone. \textbf{b}, Case 2: equilibrium loss combined with smoothness regularization. \textbf{c}, Case 3: equilibrium loss combined with non-triviality regularization. \textbf{d}, Case 4: complete loss function incorporating equilibrium, non-triviality, and smoothness regularization. For each case, hexbin parity plots compare the predicted (Pred.) and ground-truth (GT) material parameter fields: (i) \(\boldsymbol{\theta}_{\mathrm{C10}}\), (ii) \(\boldsymbol{\theta}_{\mathrm{k1}}\), and (iii) \(\boldsymbol{\theta}_{\mathrm{1/D1}}\). The dashed diagonal lines denote the \(1{:}1\) reference, while the reported \(R^2\) values quantify agreement between the predicted and ground-truth parameter fields. The color scale indicates the hexbin counts.}
        \label{fig_ablation_study}
    \end{figure}

\subsection{Convergence behavior of the loss function}
    \begin{figure}[H]
        \centering
        \includegraphics[width=\linewidth]{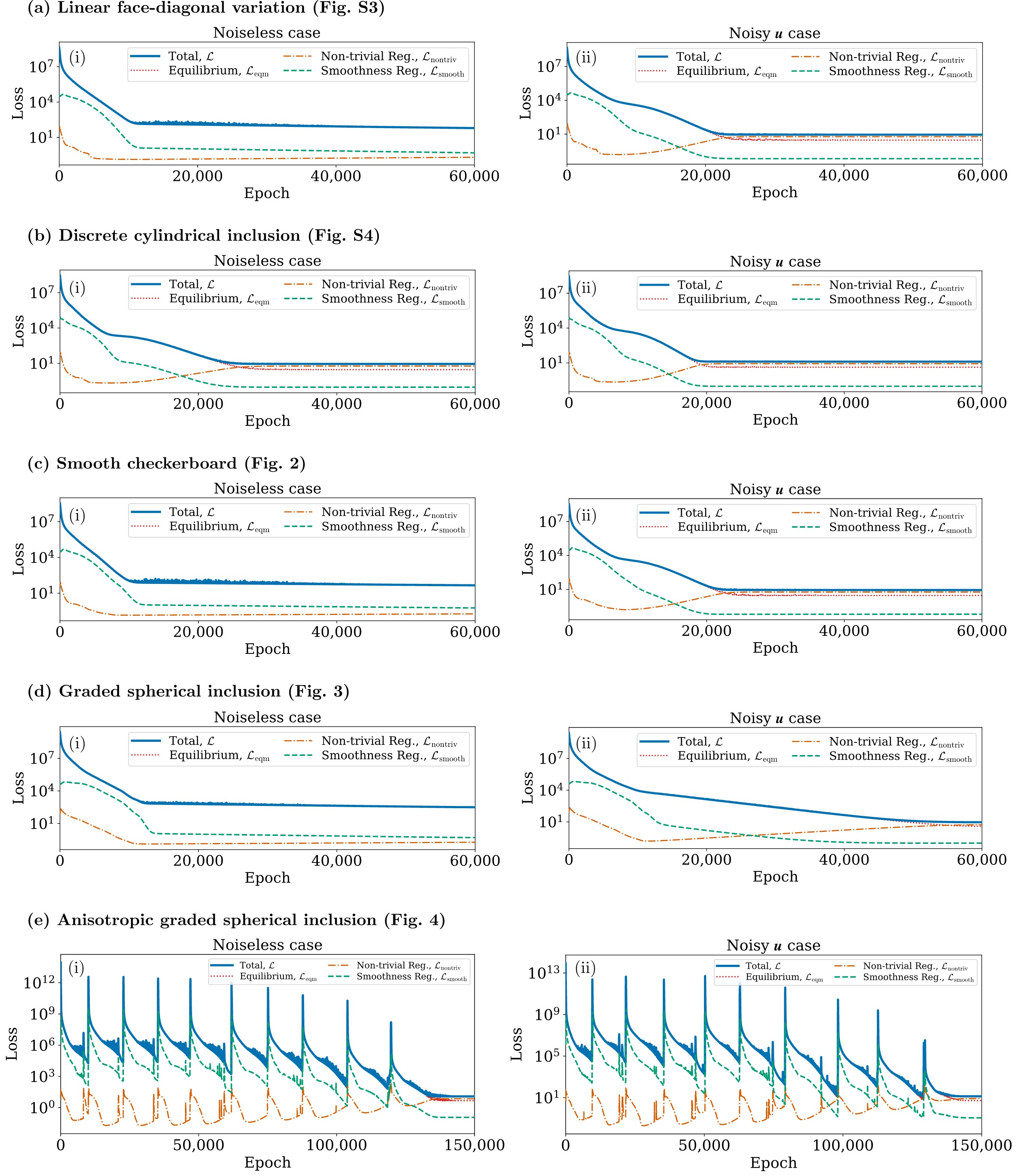}
        \caption{\textbf{Convergence histories of the loss function for the three-dimensional numerical test cases.} \textbf{a--e}, Loss evolution for the linear face-diagonal variation, discrete cylindrical inclusion, smooth checkerboard, graded spherical inclusion, and anisotropic graded spherical inclusion cases, respectively. Subpanels (i) and (ii) correspond to the noiseless and noise-perturbed displacement cases. Each plot presents the total loss along with its equilibrium, non-triviality regularization, and smoothness regularization loss terms as functions of epoch on a logarithmic scale.}
        \label{fig_loss_vs_epochs_3D}
    \end{figure}

\subsection{Comparison with independent uniaxial characterization}

    \begin{figure}[H]
        \centering
        \includegraphics[width=\linewidth]{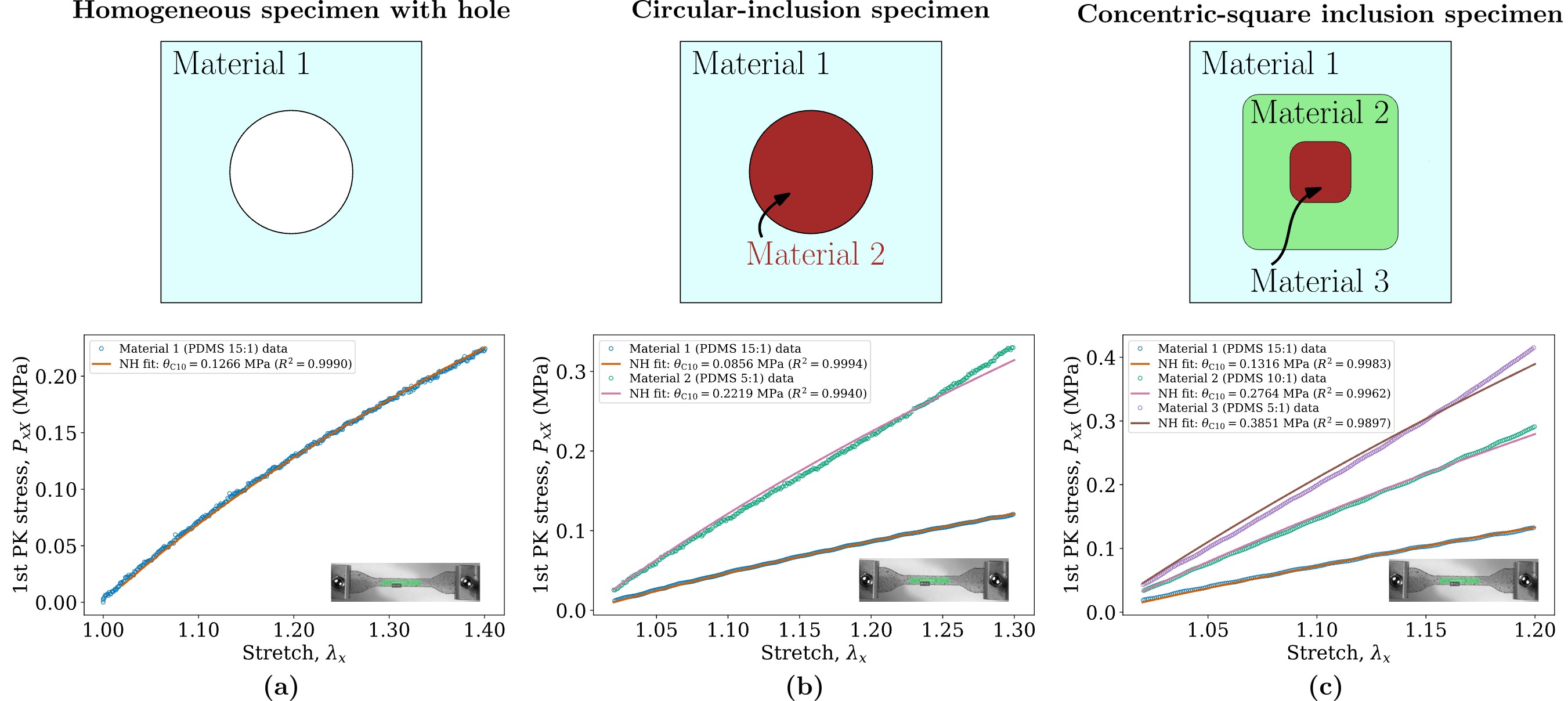}
        \caption{\textbf{Uniaxial test characterization and Neo-Hookean fitting of the PDMS constituent materials.} The first Piola-Kirchhoff stress--stretch responses obtained from uniaxial tensile tests and corresponding incompressible Neo-Hookean fits for the material formulations used in the experimental specimens. \textbf{a}, PDMS 15:1 matrix material used for the homogeneous specimen. \textbf{b}, PDMS 15:1 matrix and 5:1 circular-inclusion materials used for the circular-inclusion specimen. \textbf{c}, PDMS 15:1 matrix together with the 10:1 and 5:1 inclusion materials used for the concentric-square specimen. The fitted \(\theta_{\mathrm{C10}}\) values and corresponding \(R^2\) coefficients are indicated in each panel.}
        \label{fig_uniaxial_tests}
    \end{figure}
    
\begin{table}[htbp]
\centering
\renewcommand{\arraystretch}{1.25}
\setlength{\tabcolsep}{4pt}
% \small

\begin{tabularx}{\textwidth}{
    >{\raggedright\arraybackslash}p{4.5cm}
    >{\raggedright\arraybackslash}p{3.6cm}
    >{\centering\arraybackslash}p{3.8cm}
    >{\centering\arraybackslash}p{2.9cm}
}
\toprule
\textbf{Specimen case} &
\textbf{Material region} \textit{Refer} Fig.~\ref{fig_uniaxial_tests} &
\textbf{Proposed framework, \(\boldsymbol{\theta}_{\mathrm{C10}}\) (MPa)} &
\textbf{Uniaxial test fit, \(\boldsymbol{\theta}_{\mathrm{C10}}\) (MPa)} \\
\midrule

Homogeneous specimen
& Material 1
& 0.1432 $\pm$ 0.0087
& 0.1266 \\

\midrule

Circular-inclusion specimen
& Material 1
& 0.0911 $\pm$ 0.0118
& 0.0856 \\

& Material 2
& 0.1918 $\pm$ 0.0218 
& 0.2219 \\

\midrule

Concentric-square inclusion 
& Material 1
& 0.1453 $\pm$ 0.0086 
& 0.1316 \\

& Material 2
& 0.2591 $\pm$ 0.0181
& 0.2764 \\

& Material 3
& 0.3212 $\pm$ 0.0216
& 0.3851 \\

\bottomrule
\end{tabularx}
\caption{\textbf{Comparison of the predicted Neo-Hookean material parameter with independent uniaxial-test fits.}
The predicted value (mean \(\pm\) std.) of the spatially identified \(\theta_{\mathrm{C10}}\) field is compared with the corresponding \(\theta_{\mathrm{C10}}\) obtained by fitting the incompressible Neo-Hookean model to uniaxial experimental data for each material region. We fit Gaussian Mixture Models to the histograms shown in Fig.~9d, 10d and 11d in the main text to obtain the mean and standard deviation for the predicted \(\theta_{\mathrm{C10}}\).}
\label{table_uniaxial_comparison}
\end{table}

\subsection{Sensitivity analysis of regularization length scale on parameter recovery}
    \begin{figure}[H]
        \centering
        \includegraphics[width=0.97\linewidth]{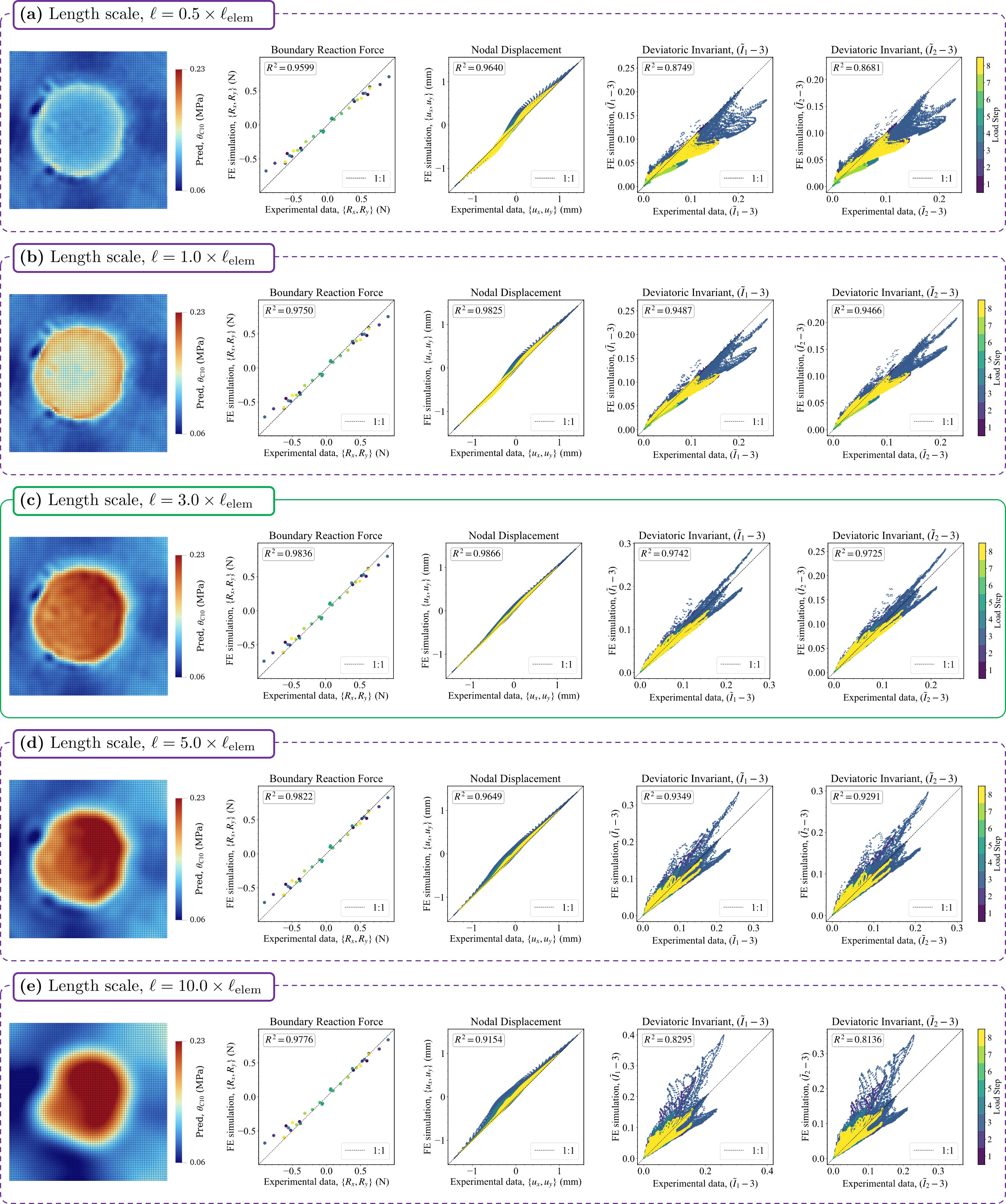}
        \caption{\textbf{Effect of the normalized smoothness regularization length scale on inverse identification of the circular-inclusion specimen.} \textbf{a--e}, Results for normalized length scales \(\ell/\ell_{\mathrm{elem}}=0.5, 1, 3, 5,\) and \(10\), respectively, where \(\ell_{\mathrm{elem}}\) is the characteristic element length. Each case presents the recovered \(\boldsymbol{\theta}_{\mathrm{C10}}\) field and forward FE validation of boundary reaction forces, nodal displacements, and deviatoric invariants \((\tilde{I}_1-3)\) and \((\tilde{I}_2-3)\).}
        \label{fig_length_scale_study}
    \end{figure}

    \begin{figure}[H]
        \centering
        \includegraphics[width=\linewidth]{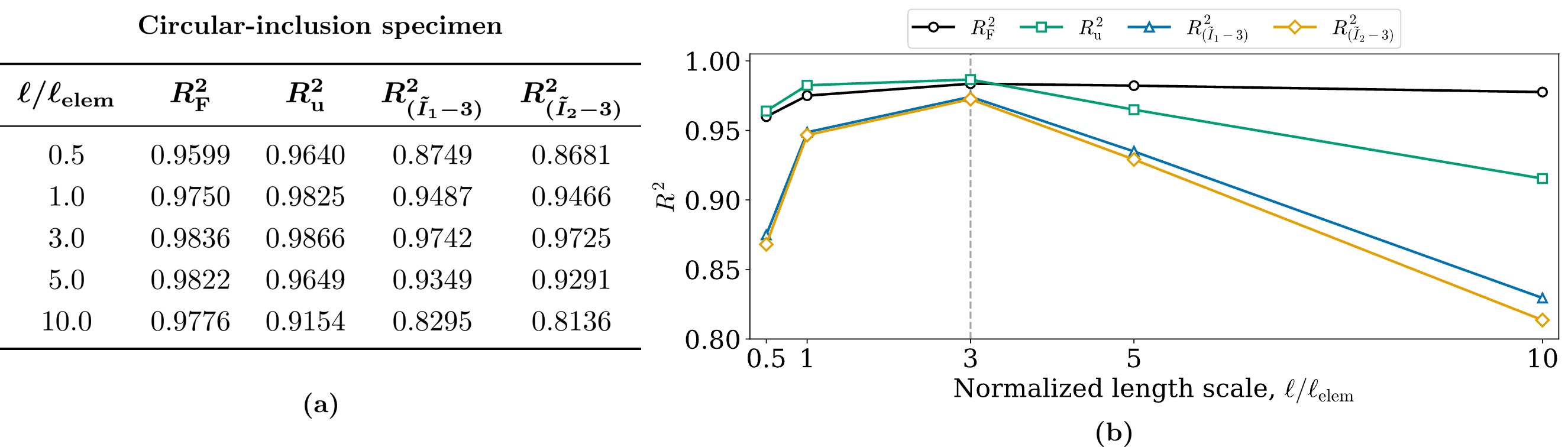}
        \caption{\textbf{Sensitivity of forward FE validation to the smoothness regularization length scale for the circular-inclusion specimen.} \textbf{a}, Comparison of the coefficients of determination (\(R^2\)) for boundary reaction forces, nodal displacements, and deviatoric invariants \((\tilde{I}_1-3)\) and \((\tilde{I}_2-3)\) at different normalized length scales \(\ell/\ell_{\mathrm{elem}}\). \textbf{b}, Corresponding variation of the four \(R^2\) metrics with normalized length scale. The vertical dashed line indicates \(\ell/\ell_{\mathrm{elem}}=3.0\), which yields the highest agreement across all four validation metrics among the investigated cases.}
        \label{fig_length_scale_study_table_plot}
    \end{figure}

\section{Strain-energy functions, stresses and tangent moduli}
\label{sec_umat}

This section summarizes the strain-energy density \(W\), the Cauchy stress \(\bm{\sigma}\), and the tangent modulus \(C_{ijkl}\) (\texttt{DDSDDE}) used in the Abaqus UMAT implementation of the Isihara and Holzapfel-Gasser-Ogden (HGO) constitutive models. The Neo-Hookean and Mooney-Rivlin models are recovered as special cases of the Isihara formulation by setting the corresponding material coefficients to zero, as applicable. For each model, the second Piola-Kirchhoff stress and Cauchy stress are obtained from \(\bm{S}=2\,\partial W/\partial\bm{C}\) and \(\bm{\sigma}=J^{-1}\bm{F}\bm{S}\bm{F}^{\mathrm T}\), respectively. The tangent modulus is formulated in accordance with the finite-strain UMAT implementation using the Jaumann rate of the Kirchhoff stress, \(\bm{\tau}=J\bm{\sigma}\).
\begin{equation}
    \overset{\nabla}{\tau}_{ij} = J\,C_{ijkl}\,D_{kl}
    \label{eq_jaumann_tangent}
\end{equation}
where \(\bm{D}\) denotes the rate-of-deformation tensor.

The following kinematic quantities are used throughout:
\begin{equation}
    \bm{b}=\bm{F}\bm{F}^{\mathrm T},\qquad
    J=\det\bm{F},\qquad
    \bar{\bm{b}}=J^{-2/3}\bm{b},\qquad
    \tilde{I}_1=J^{-2/3}I_1,\qquad
    \tilde{I}_2=J^{-4/3}I_2,
\end{equation}
with $I_1=\operatorname{tr}\bm{C}=\operatorname{tr}\bm{b}=b_{mm}$ and
$I_2=\tfrac12\left[(\operatorname{tr}\bm{C})^2-\operatorname{tr}(\bm{C}^2)\right]
=\tfrac12\left(b_{mm}b_{nn}-b_{mn}b_{nm}\right)$.

% ---------------------------------------------------------------------------
\subsection{Isihara model}
\label{sec_umat_Isihara}

\noindent\textbf{Strain-energy density} \citep{Isihara1951}
\begin{equation}
    W_{\mathrm{ISH}}
    = \theta_{\mathrm{C10}}\left(\tilde{I}_1-3\right)
    + \theta_{\mathrm{C01}}\left(\tilde{I}_2-3\right)
    + \theta_{\mathrm{C20}}\left(\tilde{I}_1-3\right)^2
    + \theta_{1/ \mathrm{D1}}\left(J-1\right)^2 
    \label{eq_W_Isihara}
\end{equation}

\noindent \textbf{Cauchy stress}
\begin{equation}
\begin{split}
    \bm{\sigma}_{\mathrm{ISH}}
    ={}& \frac{2\theta_{\mathrm{C10}}}{J^{5/3}}
         \left[\bm{b}-\frac{\operatorname{tr}(\bm{b})}{3}\bm{I}\right]
       + \frac{2\theta_{\mathrm{C01}}}{J^{7/3}}
         \left[\bm{b}\operatorname{tr}(\bm{b})-\bm{b}^2
         -\frac13\left(\left(\operatorname{tr}\bm{b}\right)^2
         -\operatorname{tr}(\bm{b}^2)\right)\bm{I}\right] \\
     & + \frac{4\theta_{\mathrm{C20}}}{J^{5/3}}
         \left(\frac{\operatorname{tr}(\bm{b})}{J^{2/3}}-3\right)
         \left[\bm{b}-\frac{\operatorname{tr}(\bm{b})}{3}\bm{I}\right]
       + 2\theta_{1/ \mathrm{D1}}\left(J-1\right)\bm{I}
\end{split}
\label{eq_sigma_Isihara}
\end{equation}

\noindent \textbf{Tangent modulus}\\
The tangent is additive over the four terms of Eq.~\eqref{eq_W_Isihara},
\begin{equation}
    C^{\mathrm{ISH}}_{ijkl}
    = C^{(\theta_{\mathrm{C10}})}_{ijkl}+C^{(\theta_{\mathrm{C01}})}_{ijkl}
    + C^{(\theta_{\mathrm{C20}})}_{ijkl}+C^{(\theta_{1/ \mathrm{D1}})}_{ijkl},
\end{equation}
with
\begin{equation}
    C^{(\theta_{\mathrm{C10}})}_{ijkl}
    = \frac{2\theta_{\mathrm{C10}}}{J}\left(
      \delta_{il}\bar{b}_{kj}+\bar{b}_{ik}\delta_{jl}
      -\frac23\bar{b}_{ij}\delta_{kl}
      -\frac23\delta_{ij}\bar{b}_{kl}
      +\frac29\bar{b}_{mm}\delta_{ij}\delta_{kl}\right)
    \label{eq_C_theta1}
\end{equation}
\begin{equation}
\begin{split}
    C^{(\theta_{\mathrm{C01}})}_{ijkl}
    = \frac{\theta_{\mathrm{C01}}}{J^{7/3}}\Big[\,
    & -\frac83 b_{mm}b_{ij}\delta_{kl}
      +2b_{mm}\left(b_{ik}\delta_{jl}+\delta_{il}b_{kj}\right)
      +4b_{ij}b_{kl}
      -4b_{ik}b_{lj}
      +\frac83 b_{im}b_{mj}\delta_{kl} \\
    & -2\left(b_{im}b_{mk}\delta_{jl}+\delta_{il}b_{km}b_{mj}\right)
      -\frac83 b_{mm}b_{kl}\delta_{ij}
      +\frac83 b_{lm}b_{mk}\delta_{ij} \\
    & +\frac89\left(b_{mm}b_{nn}-b_{mn}b_{nm}\right)\delta_{ij}\delta_{kl}
      \Big]
\end{split}
\label{eq_C_theta2}
\end{equation}
\begin{equation}
\begin{split}
    C^{(\theta_{\mathrm{C20}})}_{ijkl}
    = \theta_{\mathrm{C20}}\Bigg\{
    & -\frac{8}{3J^{5/3}}\left[
        \frac{b_{mm}b_{ij}\delta_{kl}}{J^{2/3}}
        -\frac{b_{mm}b_{nn}\delta_{ij}\delta_{kl}}{3J^{2/3}}
        -3b_{ij}\delta_{kl}
        +b_{mm}\delta_{ij}\delta_{kl}\right] \\
    & +\frac{4}{J^{5/3}}\left[
        \frac{b_{mm}}{J^{2/3}}\left(b_{ik}\delta_{jl}+\delta_{il}b_{kj}\right)
        -\frac{2\,b_{mm}b_{kl}\delta_{ij}}{3J^{2/3}}
        -3\left(b_{ik}\delta_{jl}+\delta_{il}b_{kj}\right)
        +2b_{kl}\delta_{ij}\right] \\
    & +\frac{8}{J^{7/3}}\left[
        b_{ij}b_{kl}
        -\frac13 b_{mm}b_{ij}\delta_{kl}
        -\frac13 b_{mm}b_{kl}\delta_{ij}
        +\frac19 b_{mm}b_{nn}\delta_{ij}\delta_{kl}\right]
    \Bigg\}
\end{split}
\label{eq_C_theta3}
\end{equation}
\begin{equation}
    C^{(\theta_{1/ \mathrm{D1}})}_{ijkl} = 2\theta_{1/ \mathrm{D1}}\left(2J-1\right)\delta_{ij}\delta_{kl}
    \label{eq_C_theta19}
\end{equation}

% ---------------------------------------------------------------------------
\subsection{Holzapfel-Gasser-Ogden (HGO) model}
\label{sec_umat_hgo}

For the anisotropic model, $\bm{a}$ denotes the unit fiber direction in the
reference configuration, and
\begin{equation}
    \bm{m}=\bm{F}\bm{a},\qquad
    I_a=\bm{a}\cdot\bm{C}\bm{a}=\bm{m}\cdot\bm{m},\qquad
    \tilde{I}_a=J^{-2/3}I_a
\end{equation}

\noindent \textbf{Strain-energy density} \citep{Holzapfel2000}
\begin{equation}
    W_{\mathrm{HGO}}
    = \theta_{\mathrm{C10}}\left(\tilde{I}_1-3\right)
    + \theta_{1/ \mathrm{D1}}\left[\frac{J^2-1}{2}-\ln J\right]
    + \frac{k_1}{2k_2}\left[\exp\!\left\{k_2\left(\tilde{I}_a-1\right)^2\right\}-1\right]
    \label{eq_W_hgo}
\end{equation}

\noindent \textbf{Cauchy stress}\\
Using the shorthand \( \mathcal{E}=\exp\!\left\{k_2\left(J^{-2/3}I_a-1\right)^2\right\} \) the Cauchy stress is
\begin{equation}
\begin{split}
    \bm{\sigma}_{\mathrm{HGO}}
    ={}& \frac{2\theta_{\mathrm{C10}}}{J^{5/3}}
         \left[\bm{b}-\frac{\operatorname{tr}(\bm{b})}{3}\bm{I}\right]
       + \theta_{1/ \mathrm{D1}}\left(J-\frac1J\right)\bm{I} \\
     & + \frac{k_1}{2}\,\mathcal{E}
         \left(I_aJ^{-7/3}-J^{-5/3}\right)
         \left[4\,\bm{m}\otimes\bm{m}-\frac43 I_a\bm{I}\right]
\end{split}
\label{eq_sigma_hgo}
\end{equation}

\noindent \textbf{Tangent modulus}\\
Defining the second-order tensor
\begin{equation}
    A_{ij}=(I_aJ^{-4/3}-J^{-2/3})
    \left[-\frac43 I_a\delta_{ij}+4m_im_j\right]
\end{equation}
the tangent modulus is
\begin{equation}
\begin{split}
    C^{\mathrm{HGO}}_{ijkl}
    ={}& C^{(\theta_{\mathrm{C10}})}_{ijkl}
       + 2\theta_{1/ \mathrm{D1}}J\,\delta_{ij}\delta_{kl}
       + k_1k_2\,\mathcal{E}\left(\tilde{I}_a-1\right)J^{-5/3}\,
         A_{ij}\left(2m_km_l-\frac23 I_a\delta_{kl}\right) \\
     & + \frac{k_1}{2}\,\mathcal{E}\,\Bigg[
         -\frac43\delta_{ij}\left(
            \frac{4I_a}{J^{7/3}}m_km_l
           -\frac{4I_a^2}{3J^{7/3}}\delta_{kl}
           -\frac{2}{J^{5/3}}m_km_l
           +\frac{2I_a}{3J^{5/3}}\delta_{kl}\right) \\
     & \hspace{5.5em}
         +4m_im_j\left(
            \frac{2}{J^{7/3}}m_km_l
           -\frac{4I_a}{3J^{7/3}}\delta_{kl}
           +\frac{2}{3J^{5/3}}\delta_{kl}\right)\Bigg] \\
     & + 2k_1\,\mathcal{E}\left(I_aJ^{-7/3}-J^{-5/3}\right)
         \left(\delta_{il}m_km_j+m_im_k\delta_{jl}\right)
\end{split}
\label{eq_C_hgo}
\end{equation}
where $C^{(\theta_{\mathrm{C10}})}_{ijkl}$ is the isotropic term given in
Eq.~\eqref{eq_C_theta1}.

The tangent moduli in Eqs.~\eqref{eq_C_theta1}-\eqref{eq_C_theta19} and
\eqref{eq_C_hgo} are written in the form obtained directly from the
derivation. Terms arising from products such as $\bm{D}\bm{b}$, $\bm{b}\bm{D}$,
$\bm{b}\bm{D}\bm{b}$ and $\bm{D}(\bm{m}\otimes\bm{m})$ give the correct stress
rate when contracted with the symmetric tensor $D_{kl}$, but are not
individually minor-symmetric. The tangent passed to Abaqus (DDSDDE) is
therefore the minor-symmetric part
\begin{equation}
    C^{\mathrm{sym}}_{ijkl}
    = \frac14\left(C_{ijkl}+C_{jikl}+C_{ijlk}+C_{jilk}\right),
    \label{eq_C_sym}
\end{equation}
which leaves $C_{ijkl}D_{kl}$ unchanged for symmetric $\bm{D}$. For the
hyperelastic models considered here, $C^{\mathrm{sym}}_{ijkl}$ also has the
major symmetry $C^{\mathrm{sym}}_{ijkl}=C^{\mathrm{sym}}_{klij}$, so no
further symmetrization is required.

\bibliographystyle{elsarticle-num-names} 
\bibliography{references}